\documentclass[journal]{IEEEtran}

\ifCLASSINFOpdf
\else
\fi

\usepackage{times}
\usepackage{epsfig}
\usepackage{graphicx}
\usepackage{amsmath}
\usepackage{amssymb}
\usepackage{mathrsfs}
\usepackage{multirow}
\usepackage{blkarray}
\usepackage{mathtools}
\usepackage{float}
\usepackage{array}
\usepackage{epstopdf}
\usepackage{subcaption}
\usepackage[T1]{fontenc}
\usepackage{algorithm}
\usepackage{algpseudocode}
\usepackage{subfig}
\usepackage{color}

\newcommand{\argmin}{\arg\!\min}

\begin{document}

\title{Screen Content Image Segmentation Using Robust Regression and Sparse Decomposition}

\author{Shervin~Minaee,~\IEEEmembership{Student Member,~IEEE,}
        and~Yao~Wang,~\IEEEmembership{Fellow,~IEEE}
}

\maketitle

\begin{abstract}
This paper considers how to separate text and/or graphics from smooth background in screen content and mixed document images and proposes two approaches to perform this segmentation task.
The proposed methods make use of the fact that the background in each block is usually smoothly varying and can be modeled well by a linear combination of a few smoothly varying basis functions, while the foreground text and graphics create sharp discontinuity. The algorithms separate the background and foreground pixels by trying to fit background pixel values in the block into a smooth function using two different schemes. One is based on robust regression, where the inlier pixels will be considered as background, while remaining outlier pixels will be considered foreground. 
The second approach uses a sparse decomposition framework where the background and foreground layers are modeled with a smooth and sparse components respectively.
These algorithms have been tested on images extracted from HEVC standard test sequences for screen content coding, and are shown to have superior performance over previous approaches. 
The proposed methods can be used in different applications such as text extraction, separate coding of background and foreground for compression of screen content, and medical image segmentation.
\end{abstract}


\IEEEpeerreviewmaketitle

\section{Introduction}
\IEEEPARstart{S}{creen} content images refer to images appearing on the display screens of electronic devices such as computers and smart phones \cite{SCC_tran}, \cite{SCC}.  These images have similar characteristics as mixed content documents (such as a magazine page). They often contain two layers, a pictorial smooth background and a foreground consisting of text and line graphics. The usual image compression algorithms such as JPEG2000 \cite{jpeg} and HEVC intra frame coding \cite{HEVC} may not result in a good compression rate for this kind of images because the foreground consists of sharp discontinuities. 
In these cases, segmenting the image into two layers and coding them separately may be more efficient.
The idea of segmenting an image for better compression was proposed for check image compression \cite{check}, in DjVu algorithm for scanned document compression \cite{djvu} and the mixed raster content representation \cite{raster}.
Foreground segmentation in such images is also a necessary preprocessing step for text extraction \cite{text}, which is essential for automatic character recognition and image understanding \cite{OCR}.

Screen content and mixed document images are hard to segment, because the foreground may be overlaid over a smoothly varying background that has a color range that overlaps with the color of the foreground. Also because of the use of sub-pixel rendering, the same text/line often has different colors. Even in the absence of sub-pixel rendering, pixels belonging to the same text/line often have somewhat different colors.

Different algorithms have been proposed in the past for foreground-background segmentation in mixed content document images and screen-content video frames such as hierarchical k-means clustering in DjVu \cite{djvu} and shape primitive extraction and coding (SPEC) \cite{SPEC}.
There is also some recent work for text extraction in screen content images called scale and orientation invariant text segmentation \cite{scale_inv}.

The hierarchical k-means clustering method proposed in DjVu applies the k-means clustering algorithm with k=2 on blocks in multi-resolution. It first applies the k-means clustering algorithm on large blocks to obtain foreground and background colors and then uses them as the initial foreground and background colors for the smaller blocks in the next stage. It also applies some post-processing at the end to refine the results. This algorithm has difficulty in segmenting regions where background and foreground color intensities overlap and it is hard to determine whether a pixel belongs to the background or foreground just based on its intensity value.

In the shape primitive extraction and coding (SPEC) method, which was developed for segmentation of screen content \cite{SPEC}, a two-step segmentation algorithm is proposed. In the first step the algorithm classifies each block of size $16 \times 16$ into either pictorial block or text/graphics based on the number of colors. If the number of colors is more than a threshold, 32, the block will be classified into pictorial block, otherwise to text/graphics.
In the second step, the algorithm refines the segmentation result of pictorial blocks, by extracting shape primitives (horizontal line, vertical line or a rectangle with the same color) and then comparing the size and color of the shape primitives with some threshold.
Because blocks containing smoothly varying background over a narrow range can also have a small color number,  it is hard to find a fixed color number threshold that can robustly separate pictorial blocks and text/graphics blocks. Furthermore, text and graphics  in screen content images typically have some variation in their colors, even in the absence of sub-pixel rendering. These challenges limit the effectiveness of SPEC.

In the scale and orientation invariant text segmentation, the authors proposed a two stage procedure where in the first step, a coarse segmentation layer for textural regions is extracted based on the intensity variation distribution of text characters. There could be a few pictorial regions in the segmentation output of the first stage. In the second stage, a textual connected component (TCC) based refinement is proposed to eliminate the survived pictorial regions. This approach is primarily based on morphological operations, and since it is designed based on a lot of assumption on the underlying text in the foreground, it cannot be used for segmentation of other foreground (such as line graphics patterns) in screen content.


The above problems with prior approaches motivate us to design a segmentation algorithm that does not rely solely on the pixel intensity but rather exploits the smoothness of the background region. In other words, instead of looking at the intensities of individual pixels and deciding whether each pixel should belong to background or foreground, we first look at the smoothness of a group of pixels and then decide whether each pixel should belong to background or foreground.
Towards this goal, in a preliminary work \cite{LAD}, we have developed a least absolute deviation (LAD) method, which fits a smooth model to an image block and classifies the pixels to either background or foreground based on the fitting error. It uses the $\ell_1$ norm on the fitting error to enforce the sparsity of the error term. Although this algorithm achieved significantly better segmentation than both the DjVu and SPEC, it suffers from several problems. One is the difficulty of determining the number of bases to be used in the smooth model. If we use very few bases we will not be able to represent a complicated background, and if we use too many bases we may end up with representing the foreground pixels with the smooth model and considering them as background. The second problem is that this prior method does not explicitly exploit the fact that foreground pixels are typically connected. Finally, minimizing the $\ell_1$ norm of the fitting error does not always give the sparsest fitting error.

In this work, we propose two segmentation algorithms, one relying on the robust regression technique \cite{robustregression} and the other one using sparse decomposition. These algorithms overcome some of the above problems, and to the best of our knowledge they have not been investigated previously.
In the robust regression based approach, we try to find  a smooth model that can fit the background pixels accurately. 
One important point is that, when the smooth model is fitted to the image, it should not be affected by the foreground pixels. This is accomplished through the RANSAC (Random sample consensus) method, which minimizes the number of pixels whose fitting error exceeds a certain threshold.

In the other approach, which uses sparse decomposition (SD), the background and foreground part of the image are modeled with a smooth  component and a sparse component, respectively. 
Since we do not know in advance how many basis functions to include for the background part, the method chooses from a large set of bases that we think are sufficient to represent the most "complex" background, while minimizing coefficient $\ell_1$ norm to avoid overfitting of the smooth model on the foreground pixels.
The total variation \cite{TV_osher} of the foreground component is also added in the cost function to promote the connectivity of the foreground pixels.
Both RANSAC and SD algorithms treat the smooth part of the image as background, and fit a smooth model to the image and treat outliers as foreground. 
But there are some differences between these two algorithms. In RANSAC, the model is fitted by directly minimizing the number of outliers ($\ell_0$ norm of outliers), whereas in the sparse decomposition case, the $\ell_1$ norm of foreground pixels are minimized. 
On the other hand, the SD algorithm specifically promotes the sparsity of the background coefficients and the connectivity of foreground pixels, whereas the RANSAC method does not enforce these properties. 

We note that none of the prior approaches for background/foreground segmentation explicitly make use of the fact that the background is typically smoothly varying, even though it may have a large dynamic range (e.g. a linear change from black to white). In fact any clustering-based segmentation method would fail for the case where the background color is smoothly changing  over a relatively large intensity range, and the background color could be similar to the foreground color in some regions. But using the proposed method we can easily segment this kind of images. An example of this case is shown in Figure 1 where we compare the segmentation result by our algorithm with the result of hierarchical k-means clustering used in DjVu \cite{djvu}. As it can be seen, in the clustering based result the background pixels are split into separate clusters.

\begin{figure}[1 h]
\begin{center}
\hspace{-0.15cm}
    \includegraphics [scale=0.14] {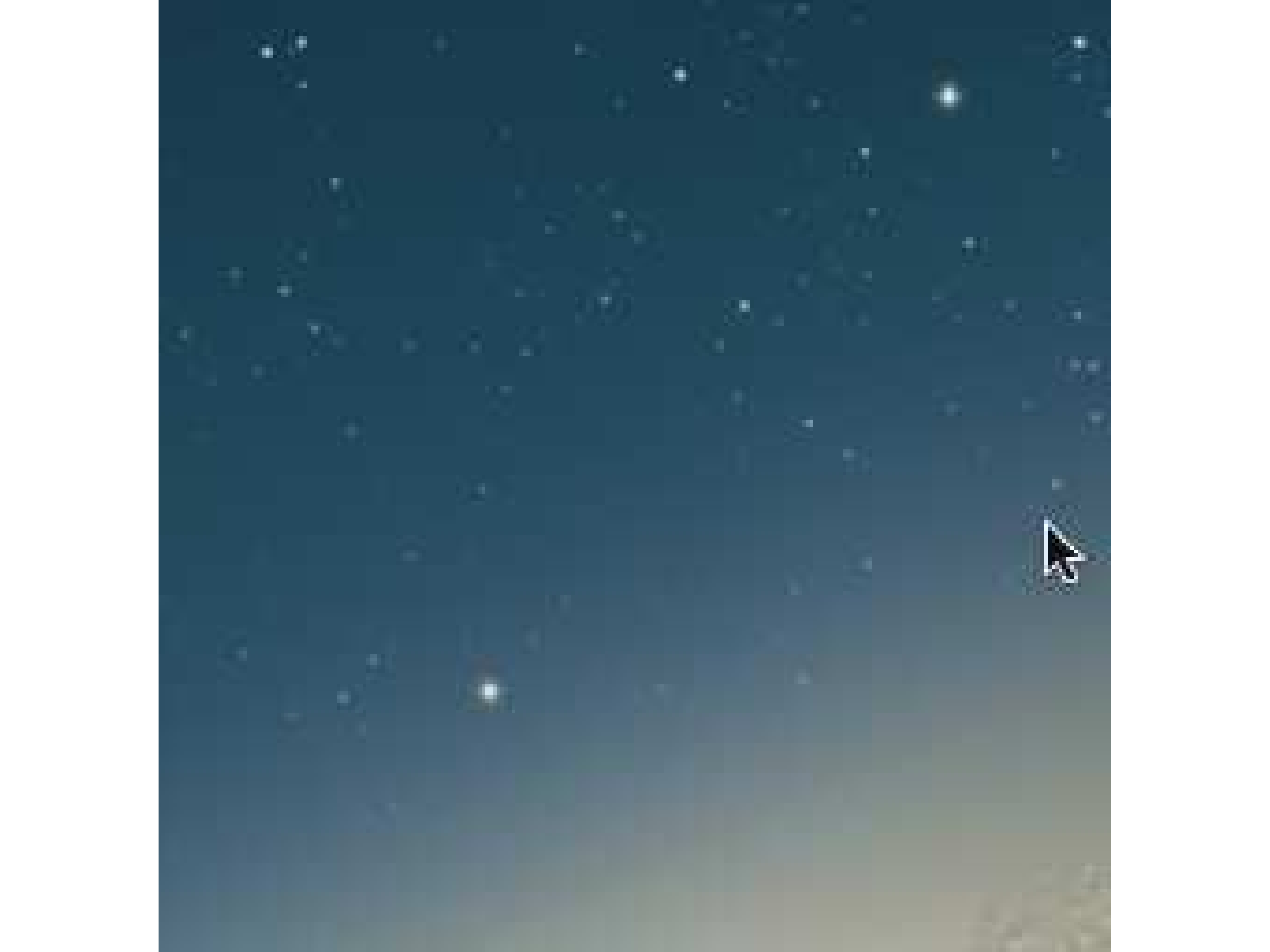}
\hspace{-0.3cm}	\includegraphics [scale=0.14] {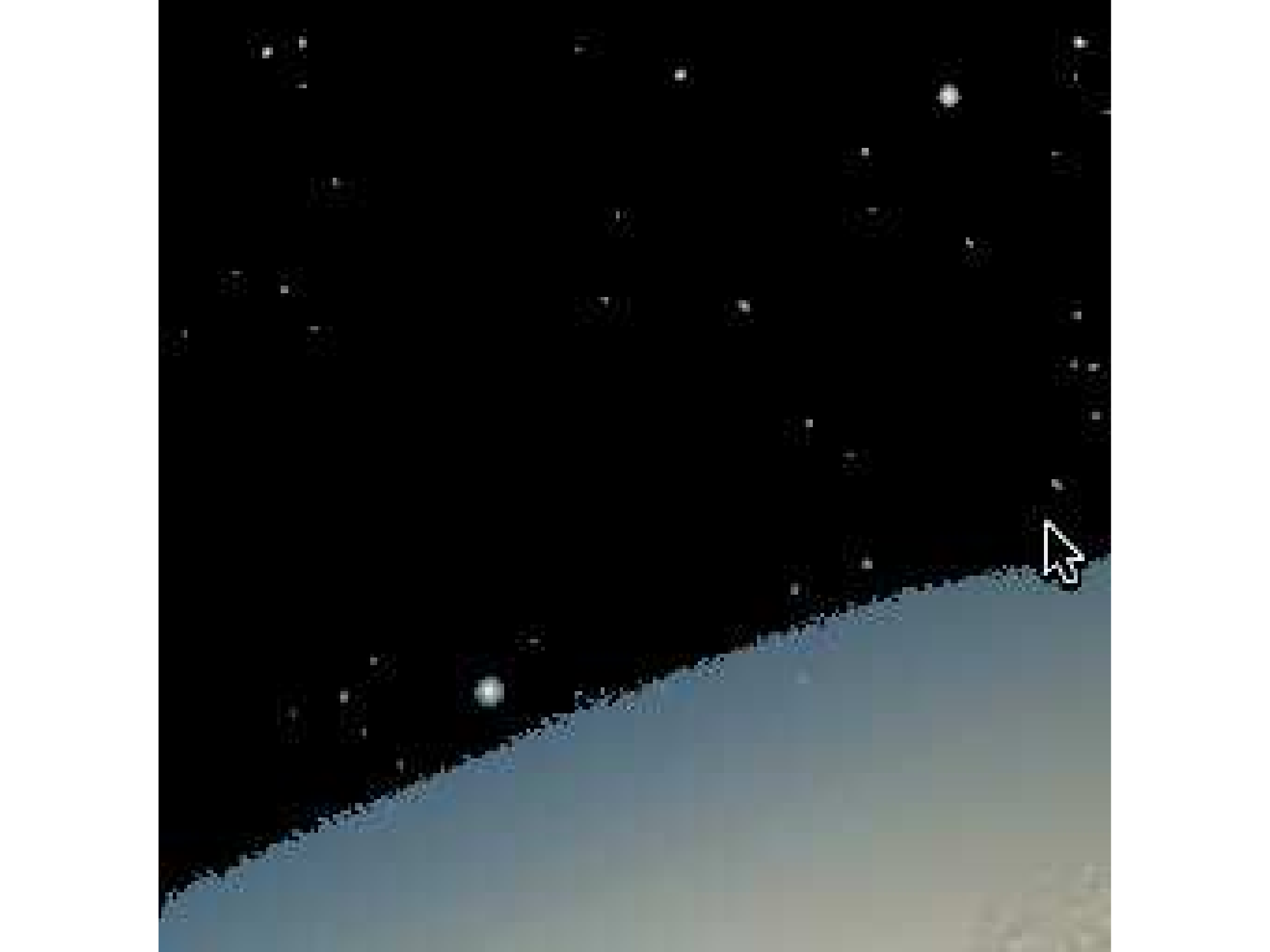}
  \hspace{-0.45cm}  \includegraphics [scale=0.14] {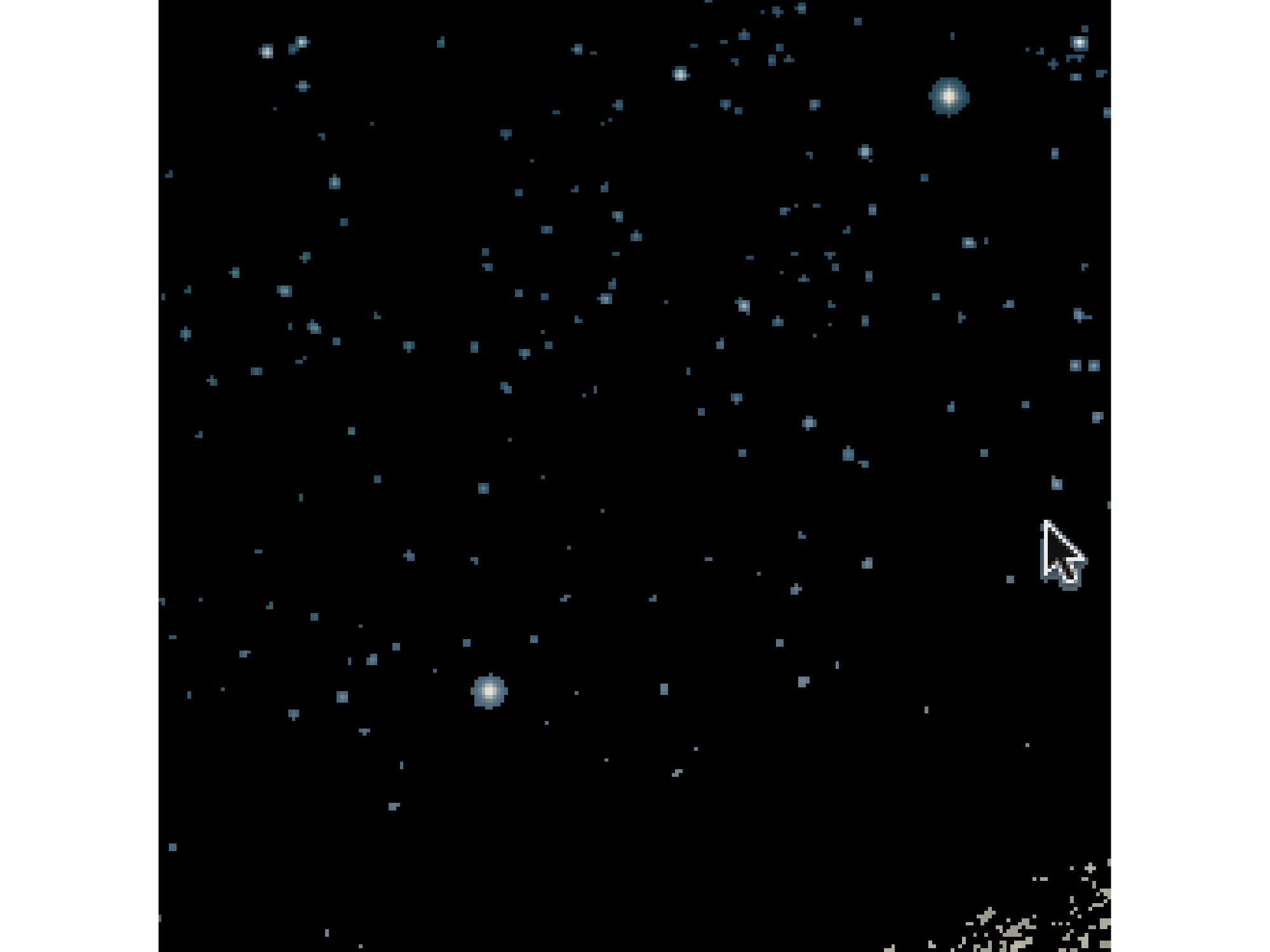} 
\end{center}
  \caption{Segmentation result for a block with smoothly changing background. The left, middle and right images denote the original image, segmented foreground by hierarchical clustering in DjVu \cite{djvu} and the robust regression algorithm respectively.  }
\end{figure}

Both of our proposed schemes assume the smooth background can be represented well by a few low frequency basis and the foreground is sparse. This is different from the prior image segmentation works \cite{bs1}-\cite{bs4} that assume an image (directly or after some geometric transformation) consists of a low rank component and a sparse component \cite{SLD}, and use the method of low-rank decomposition \cite{bs2}  to separate the low rank component and the sparse component.  Because the smooth backgrounds in screen content images may not always have low rank and the foreground may happen to have low rank patterns (e.g. horizontal and vertical lines), applying such decomposition and assuming the low rank component is the background and the sparse component is the foreground may not always yield satisfactory results.
Figure 2 shows the result of sparse and low-rank decomposition for three sample images in our dataset.
Note that each component could have negative values, so to better display the results we show each image on its effective range. For sparse component we show its absolute value, otherwise some of the detected texts in that layer will not be visible.
The foreground map for each case is derived by thresholding the corresponding sparse component. 
As can be seen, there are some horizontal and/or vertical structures in the low rank components that do not belong to the background for the first two images. 
Also for the second image, some of the foreground patterns are not segmented in the sparse component. 
Only for the third image, where the background actually has low rank and the foreground is very sparse and disconnected (hence lacks low-rank components), the low-rank decomposition method yields very good result. Therefore, the low-rank decomposition framework is not always appropriate for the background-foreground segmentation problem.  As shown in Section IV, the proposed approach that uses low frequency smooth bases to represent the background is valid for a variety of background patterns and works well for all the screen content images we have examined.

\begin{figure}[2 h]
\begin{center}
\hspace{-0.15cm}
  \hspace{0.01cm}  \includegraphics [scale=0.375] {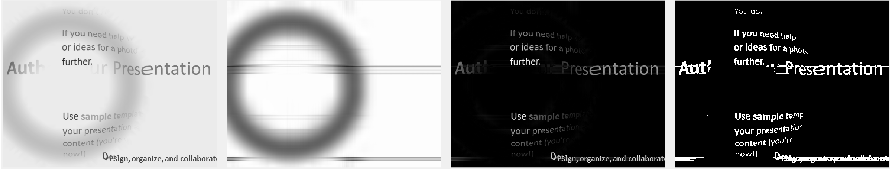} \\
 \vspace{0.21cm}	\includegraphics [scale=0.375] {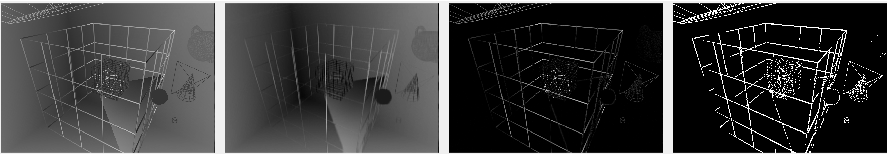} \\
 \vspace{0.2cm}	\includegraphics [scale=0.375] {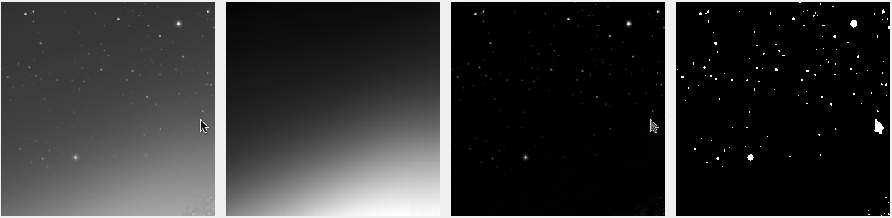}
\end{center}	
  \caption{Segmentation result using RPCA. The first, second, third, and fourth columns show the original images, the low-rank components, the sparse components, and the foreground masks respectively.}
\end{figure}

The structure of this paper is as follows: Section II presents the core idea of the proposed segmentation methods. The procedure of basis selection for background representation is explained in Section II.A. Section II.B briefly describes the first proposed scheme based on the RANSAC algorithm. The second proposed algorithm based on sparse decomposition is described in Section II.C. The final segmentation algorithm that includes both the core robust regression/sparse decomposition algorithm as well as preprocessing steps and quadtree decomposition is discussed in Section III. Section IV provides the experimental results for these algorithms. 
And finally the paper is concluded in Section V.

\section{Background Modeling and Background-Foreground Separation}
The main idea of this work lies in the fact that if an image only consists of background, it should be well represented with a few smooth basis functions. By well representation we mean that the approximated value at a pixel with the smooth functions should have an error less than a desired threshold at every pixel. Whereas if the image has some foreground pixels overlaid on top of a smooth background, those foreground pixels cannot be well represented using the smooth representation.
Since the foreground pixels cannot be modeled with this smooth representation they would usually have a large distortion by using this model.
Therefore the background segmentation task simplifies into finding the set of inlier pixels, which can be approximated well using this smooth model. 
Now some questions arise here:
\begin{enumerate}
\item What is  a good class of smooth models that can represent the background layer accurately and compactly?
\item How can we derive the model parameters, $\alpha_k$'s, such that they are not affected by foreground pixels, especially if we have many foreground pixels?
\end{enumerate}

For the first question, we divide each image into non-overlapping blocks of size $N \times N$, and represent each image block, denoted by $F(x,y)$, with a smooth model as a linear combination of a set of two dimensional smooth functions as $\sum_{k=1}^K \alpha_k P_k(x,y)$. Here low frequency two-dimensional DCT basis functions are used as $P_k(x,y)$, and the reason why DCT basis are used and how the number $K$ is chosen is explained in Section II.A. 
The 2-D DCT function is defined as:
\begin{equation*}
P_{u,v}(x,y)= \beta_u \beta_v cos((2x+1)\pi u/2N) cos((2y+1)\pi v/2N) 
\end{equation*}
where $u$ and $v$ denote the frequency indices of the basis and $\beta_u$ and $\beta_v$ are normalization factors and $x$ and $y$ denote spatial coordinate of the image pixel.
We order all the possible basis functions in the conventional zig-zag order in the (u,v) plane, and choose the first $K$ basis functions.

The second question is kind of a chicken and egg problem: To find the model parameters we need to know which pixel belongs to the background and to know which pixel belongs to background we need to know what the model parameters are. 
One simple way is to define some cost function, which measures the goodness of fit between the original pixel intensities and the ones predicted by the smooth model, and then minimize the cost function. If we use the $\ell_p$-norm  of the fitting error ($p$ can be 0, 1, or 2), the problem can be written as:
\begin{gather*}
\{\alpha_1^*,...,\alpha_K^*\}= \argmin_{\alpha_1,...,\alpha_K}  \sum_{x,y} |F(x,y)- \sum_{k=1}^K \alpha_k P_k(x,y)|^p
\end{gather*}

We can also look at the 1D version of the above optimization problem by converting the 2D blocks of size $N \times N$ into a vector of length $N^2$, denoted by $\boldsymbol{f}$, by concatenating the columns and denoting $\sum_{k=1}^K \alpha_k P_k(x,y)$ as $ \textbf{P}\boldsymbol{\alpha}$ where $\textbf{P}$ is a matrix of size $N^2\times K$ in which the k-th column corresponds to the vectorized version of $P_k(x,y)$ and, $\boldsymbol{\alpha}=[\alpha_1,...,\alpha_K]^\text{T}$. 
Then the problem can be formulated as:
\begin{gather}
\alpha^*= \argmin_{\alpha} \| f -P\alpha \|_p
\end{gather}

If we use the $\ell_2$-norm (i.e. $p=2$) for the cost function, the problem is simply the least squares fitting problem and is very easy to solve. In fact it has a closed-form solution as below:
\begin{gather}
\alpha^*= \argmin_{\alpha} \| f-P\alpha \|_2 \Rightarrow \alpha= (P^T P)^{-1}P^T f
\end{gather}
But this formulation has a problem that the model parameters, $\boldsymbol{\alpha}$, can be adversely affected by foreground pixels. Especially in least-square fitting, by squaring the residuals, the larger residues will get larger weights in determining the model parameters. 
We propose two approaches for deriving the model parameters, one based on robust regression and the other one based on sparse decomposition. 
The proposed schemes are more robust to outliers compared to least-square fitting.
These algorithms are described in Sections II.B and II.C respectively.

For color images, in general we need to check all color components to see whether each can be modeled by a smooth function. But we found through simulation results that typically if the luminance is smooth, the other two chrominance components are also smooth. Therefore, we first find the set of pixels whose luminance component can be modeled by a smooth function by using the luminance component for $F(x,y)$. Once we find these pixels, we check whether their chrominance components can also be modeled by a smooth function.

\subsection{ Basis Selection}
To find a good set of bases for background, we applied Karhunen-Loeve transform \cite{KLT} to a training set of smooth background images, and the derived bases turn out to be very similar to 2D DCT and 2D orthonormal polynomials.
Therefore we compared these two sets of basis functions, the DCT basis and the orthonormal polynomials which are known to be efficient for smooth image representation. The two dimensional DCT basis are outer-products of 1D DCT basis, and are well known to be very efficient for representing natural images \cite{DCT}.
To derive 2D orthonormal polynomials over an image block of size $N \times N$, we start with the $N$ 1D vectors obtained by evaluating the simple polynomials $f_n(x)=x^n$, at $x=\{1,2,...,N \}$, for $n=0, 1,.., N-1$ and orthonormalize them using Gram-Schmidt process to get $N$ orthonormal bases.
After deriving the 1D polynomial bases, we construct 2D orthonormal polynomial bases using the outer-product of 1D bases.

To compare DCT and orthonormal polynomial bases, we collected many smooth background blocks of size $64 \times 64$ from several images and tried to represent those blocks with the first $K$ polynomials and DCT basis functions in zigzag order. 
Because each block contains only smooth background pixels, we can simply apply least squares fitting to derive the model coefficients using Eq (2).
Then we use the resulting model to predict pixels' intensities and find the mean squared error (MSE) for each block. The reconstruction RMSEs (root MSE) as a function of the number of used bases, $K$, for both DCT and polynomials are shown in Figure 3. As we can see DCT has slightly smaller RMSE, so it is preferred over orthonormal polynomials.

\begin{figure}[1 h]
\vspace{-0.2cm}
\begin{center}
    \includegraphics [scale=0.5] {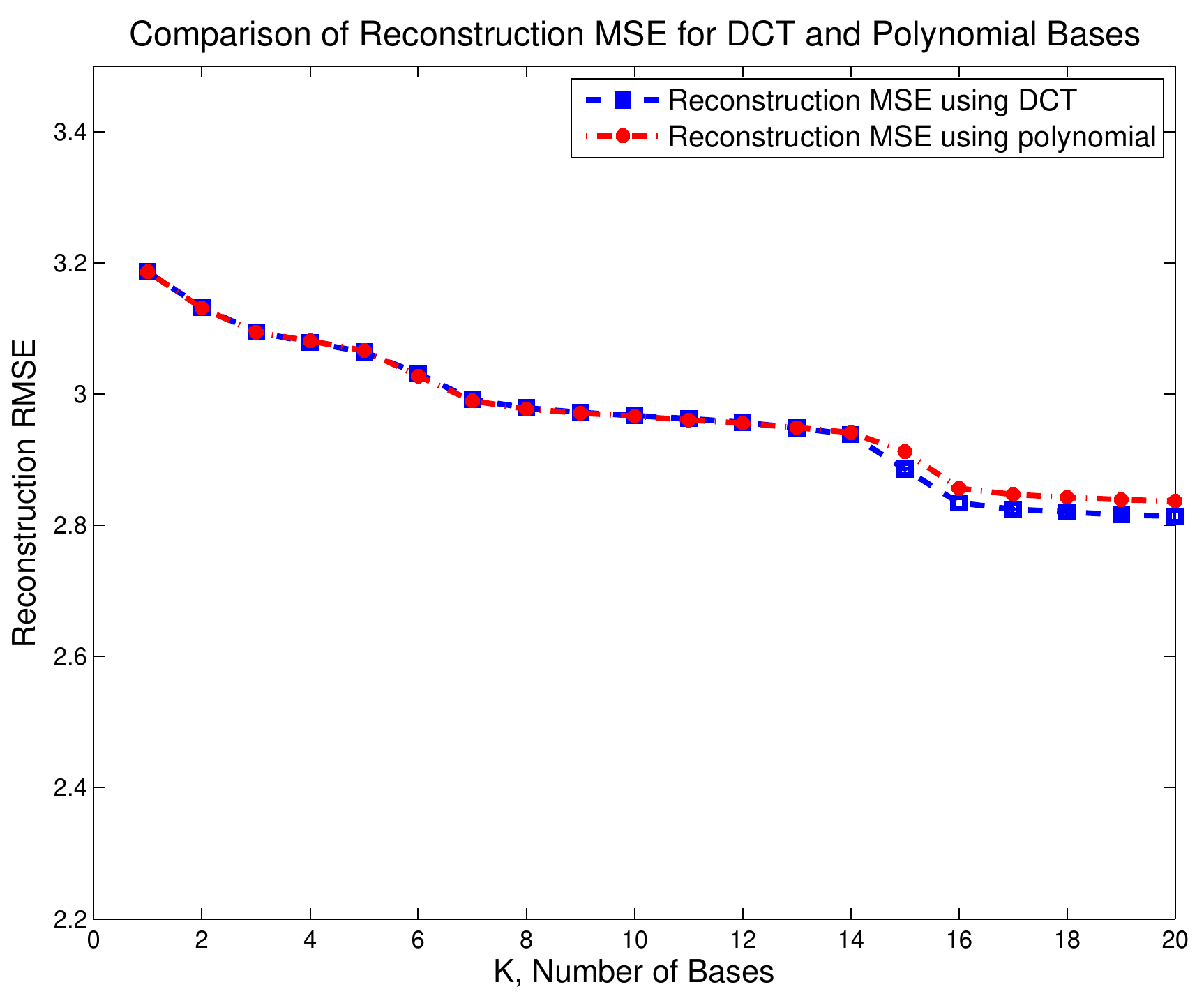}
\end{center}
  \caption{Background reconstruction RMSE vs. the number of bases.}
\end{figure}

We would like to note that for more complicated background patterns, one could use the hybrid linear models \cite{hybrid1}, \cite{hybrid2} to represent the background using a union of subspaces.
But for screen content images, the background can usually be well-represented by a few low frequency DCT bases.


\subsection{First Approach: Robust Regression Based Segmentation}
Robust regression is a form of regression analysis, which is developed to overcome some limitations of traditional algorithms \cite{robustregression}. The performance of most of the traditional regression algorithms can be significantly affected if the assumptions about underlying data-generation process are violated and they are highly sensitive to the presence of outliers. The outlier can be thought as any data-point or observation which does not follow the same pattern as the rest of observations. The robust regression algorithms are designed to find the right model for a dataset even in the presence of outliers. They basically try to remove the outliers from dataset and use the inliers for model prediction.

RANSAC \cite{RANSAC} is a popular robust regression algorithm. It is an iterative approach that performs the parameter estimation by minimizing the number of outliers (which can be thought as minimizing the $\ell_0$-norm).
RANSAC repeats two iterative procedures to find a model for a set of data. In the first step, it takes a subset of the data and derives the parameters of the model only using that subset. The cardinality of this subset is the smallest sufficient number to determine the model parameters. In the second step, it tests the model derived from the first step against the entire dataset to see how many samples can be modeled consistently. A sample will be considered as an outlier if it has a fitting error larger than a threshold that defines the maximum allowed deviation. RANSAC repeats the procedure a fixed number of times and at the end, it chooses the model with the largest consensus set (the set of inliers) as the optimum model. There is an analogy between our segmentation framework and model fitting in RANSAC. We can think of foreground pixels as outliers for the smooth model representing the background. Therefore RANSAC can be used to perform foreground segmentation task.

The proposed RANSAC algorithm for foreground/background segmentation of a block of size $N \times N$ is as follows:
\begin{enumerate}
\item Select a subset of $K$ randomly chosen pixels. Let us denote this subset by $S=\{(x_l,y_l), \ l=1,2,\ldots,K\}$.
\item Fit the model $\sum_{k=1}^K \alpha_k P_k(x,y) $ to the pixels $(x_l,y_l) \in S$ and find the $\alpha_k$'s. This is done by solving the set of $K$ linear equations $\sum_k \alpha_k P_k(x_l,y_l) = F(x_l,y_l), \ l=1,2,\ldots,K$.
\item Test all $N^2$ pixels $F(x,y)$ in the block against the fitted model. Those pixels that can be predicted with an error less than $\epsilon_{in}$ will be considered as the inliers.
\item Save the consensus set of the current iteration if it has a larger size than the largest consensus set identified so far.
\item If the inlier ratio, which is the ratio of inlier pixels to the total number of pixels, is more than 95\%, stop the algorithm.
\item Repeat this procedure up to $M_{\rm iter}$ times.
\end{enumerate}
After this procedure is finished, the pixels in the largest consensus set will be considered as inliers or equivalently background. 
The final result of RANSAC can be refined by refitting over all inliers once more  and finding all pixels with error less than $\epsilon_{in}$.
To boost the speed of the RANSAC algorithm, we stop once we found a consensus set which has an inlier ratio more than $0.95$. 

The segmentation results by RANSAC is usually very good, but it is computationally demanding. For blocks that can be easily segmented with other methods, RANSAC may be an overkill. Therefore, we propose a segmentation algorithm that has different modes in Section III.

\subsection{Second Approach: Sparse Decomposition Algorithm}
Sparse representation has been used for various applications in recent years, including face recognition, super-resolution, morphological component analysis, denosing, image restoration and sparse coding \cite{wright}-\cite{sp_coding}.
In this work, we explored the application of sparse decomposition for image segmentation.
As we mentioned earlier, the smooth background regions can be well represented with a few smooth basis functions, whereas the high-frequency component of the image belonging to the foreground, cannot be represented with this smooth model. 
But using the fact that foreground pixels occupy a relatively small percentage of the images we can model the foreground with a sparse component overlaid on background. 
Therefore it is fairly natural to think of mixed content image as a superposition of two components, one smooth and the other one sparse, as shown below:
\begin{equation}
F(x,y)= \sum_{k=1}^K \alpha_k P_k(x,y) + S(x,y)
\end{equation}
where $\sum_{i=1}^K \alpha_i P_i(x,y)$ and $S(x,y)$ correspond to the smooth background region and foreground pixels respectively. 
Therefore we can use sparse decomposition techniques to separate these two components.
After decomposition, those pixels with large value in the $S$ component will be considered as foreground.
We will denote this algorithm as "SD", for notation brevity.

To have a more compact notation, we will look at the 1D version of this problem. Denoting the 1D version of $S(x,y)$ by $s$, 
Eq. (3) can be written as:
\begin{equation}
f= {P\alpha}+s
\end{equation}

Now to perform image segmentation, we need to impose some prior knowledge about background and foreground to our optimization problem. 
Since we do not know in advance how many basis functions to include for the background part, we allow the model to choose from a large set of bases that we think are sufficient to represent the most "complex" background, while minimizing coefficient $\ell_0$ norm to avoid overfitting of the smooth model on the foreground pixels. Because if we do not restrict the parameters, we may end up with a situation that even some of the foreground pixels are represented with this model (imagine the case that we use a complete set of bases for background representation).
Therefore the number of nonzero components of $\alpha$ should be small (i.e. $\| \alpha \|_0$ should be small).
On the other hand we expect the majority of the pixels in each block to belong to the background component, therefore the number of nonzero components of $s$ should be small.
And the last but not the least one is that foreground pixels typically form connected components in an image, therefore we can add a regularization term which promotes the connectivity of foreground pixels. Here we used total variation of the foreground component to penalize isolated points in foreground.
Putting all of these priors together we will get the following optimization problem:
\begin{equation}
\begin{aligned}
& \underset{s, \alpha}{\text{minimize}}
& & \|\alpha \|_0+ \lambda_1 \| s \|_0+ \lambda_2 TV(s)   \\
& \text{subject to}
& &  f=P \alpha+s
\end{aligned}
\end{equation}
where $\lambda_1$ and $\lambda_2$ are some constants which need to be tuned. 
For the first two terms since $\ell_0$ is not convex, we use its approximated $\ell_1$ version to have a convex problem.
For the total variation we can use either the isotropic or the anisotropic version of 2D total variation \cite{TV}. To make our optimization problem simpler, we have used the anisotropic version in this algorithm, which is defined as:
\begin{equation}
\begin{aligned}
TV(s)=  \sum_{i,j} |S_{i+1,j}-S_{i,j}|+|S_{i,j+1}-S_{i,j}|
\end{aligned}
\end{equation}
After converting the 2D blocks into 1D vector, we can denote the total variation as below:
\begin{equation}
\begin{aligned}
TV(s)= \| D_xs \|_1+\| D_ys \|_1= \|Ds\|_1
\end{aligned}
\end{equation}
where $D=[ D_x',D_y']'$. Then we will get the following problem:
\begin{equation}
\begin{aligned}
& \underset{s, \alpha}{\text{minimize}}
& & \|\alpha \|_1+ \lambda_1 \| s \|_1+ \lambda_2 \|Ds\|_1   \\
& \text{subject to}
& &  P \alpha+s=f
\end{aligned}
\end{equation}
From the constraint in the above problem, we get $s= f-P\alpha$ and then we derive the following unconstrained problem:
\begin{equation}
\begin{aligned}
& \underset{ \alpha}{\text{min}}
& & \|\alpha \|_1+ \lambda_1 \| f-P\alpha \|_1+ \lambda_2 \|Df-DP\alpha\|_1
\end{aligned}
\end{equation}
This problem can be solved with different approaches, such as alternating direction method of multipliers (ADMM) \cite{ADMM}, majorization minimization \cite{seles1}, proximal algorithm \cite{patrick1} and iteratively reweighted least squares minimization \cite{IRWL}. Here we present the formulation using ADMM algorithm.\\ \\
\textbf{ADMM for solving L1 optimization:}\\
ADMM is a variant of the augmented Lagrangian method that uses the partial update for dual variable. 
To solve (9) with ADMM, we introduce the auxiliary variable $y,z$ and $x$  and convert the original problem into the following form:
\begin{equation}
\begin{aligned}
& \underset{\alpha, y, z, x}{\text{minimize}}
& & \|y \|_1+ \lambda_1 \| z \|_1+ \lambda_2 \|x\|_1   \\
& \text{subject to} & &  y=\alpha \\
& & & z=f-P \alpha \\
& & & x=Df-DP\alpha
\end{aligned}
\end{equation}
Then the augmented Lagrangian for the above problem can be formed as:
\begin{align*}
&L_{\rho_1,\rho_2,\rho_3}(\alpha,y,z,x)= \|y \|_1+ \lambda_1 \| z \|_1+ \lambda_2 \|x\|_1+ u_1^t(y- \alpha)+ \nonumber \\
&u_2^t(z+P\alpha-f)+ u_3^t(x+DP\alpha-Df)+\frac{\rho_1}{2} \| y- \alpha \|_2^2+  \nonumber \\
&\frac{\rho_2}{2} \| z+P\alpha-f \|_2^2+\frac{\rho_3}{2} \| x+DP\alpha-Df \|_2^2
\end{align*}
where $u_1$, $u_2$ and $u_3$ denote the dual variables.
Then, we can find the update rule of each variable by setting the gradient of the objective function w.r.t. to the primal variables to zero and using dual descent for dual variables as shown in Algorithm 1,
\begin{algorithm}
  \caption{pseudo-code for ADMM updates of problem (10)}\label{euclid}
  \begin{algorithmic}[1]
      \For{\texttt{$k$=1:$k_{max}$}}
        \State $\alpha^{k+1}= \underset{\alpha}{\text{argmin}} \ L_{\rho_{1:3}}(\alpha,y^k,z^k,x^k, u_1^k, u_2^k, u_3^k)$
        \Statex \qquad \qquad $ = A^{-1} \big[ u_1^{k}-P^tu_2^{k}-P^tD^tu_3^{k}+\rho_1y^k  $  
        \Statex \qquad \qquad $ +\rho_2P^t(f-z^{k})+\rho_3P^tD^t(Df-x^{k}) \big]   $         
        \State $y^{k+1}= \underset{y}{\text{argmin}} \ L_{\rho_{1:3}}(\alpha^{k+1},y,z^k,x^k, u_1^k, u_2^k, u_3^k) $     
        \Statex \qquad \qquad $ =\text{Soft}(\alpha^k- \frac{1}{\rho_1} u_1^{k},\frac{1}{\rho_1}) $   
        \State $z^{k+1}= \underset{z}{\text{argmin}} \ L_{\rho_{1:3}}(\alpha^{k+1},y^{k+1},z,x^k, u_1^k, u_2^k, u_3^k)$        
        \Statex \qquad \qquad $ = \text{Soft}(f-P\alpha^{k+1}-\frac{1}{\rho_2} u_2^{k},\frac{\lambda_1}{\rho_2}) $
        \State $x^{k+1}= \underset{x}{\text{argmin}} \ L_{\rho_{1:3}}(\alpha^{k+1},y^{k+1},z^{k+1},x, u_1^k, u_2^k, u_3^k)$        
        \Statex \qquad \qquad $ = \text{Soft}(Df-DP\alpha^{k+1}-\frac{1}{\rho_3} u_3^{k},\frac{\lambda_2}{\rho_3}) $
        \State $u_1^{k+1}= u_1^{k}+ \rho_1 (y^{k+1}-\alpha^{k+1})$        
        \State $u_2^{k+1}= u_2^{k}+ \rho_2 (z^{k+1}+P\alpha^{k+1}-f)$
        \State $u_3^{k+1}= u_3^{k}+ \rho_3 (x^{k+1}+DP\alpha^{k+1}-Df)$                
      \EndFor
  \end{algorithmic}
\end{algorithm}
\\Here $A=(\rho_3 P^tD^tDP+\rho_2P^tP+\rho_1 I)$, and $\text{Soft}(.,\lambda)$ denotes the soft-thresholding operator applied elementwise and is defined as:
\begin{gather*}
\text{Soft}(x,\lambda)= \text{sign}(x) \ \text{max}(|x|-\lambda,0)
\end{gather*}
The setting for the parameters $\rho_{1:3}$ and the regularization weights $\lambda_{1:3}$ are explained in section IV.

After finding the values of $\alpha$, we can find the sparse component as $s=f-P\alpha$. 
Then those pixels with values less than an inlier threshold $\epsilon_{in}$ in $s$ will be considered as foreground.

To show the advantage of minimizing $\ell_1$ over $\ell_2$, and also sparse decomposition over both $\ell_1$ and $\ell_2$ minimization approaches, we provide the segmentation result using least square fitting (LSF), least absolute deviation fitting (LAD) and also sparse decomposition (SD) framework for a sample image consists of foreground texts overlaid on a constant background. The original image and the segmentation results using LSF, LAD and SD are shown in Figure 4. 
\begin{figure}[4 h]
\begin{center}
    \includegraphics [scale=0.8] {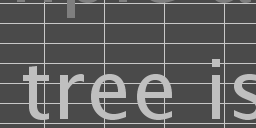} 
\end{center}
\vspace{0.02cm}  
\begin{center}
    \includegraphics [scale=0.8] {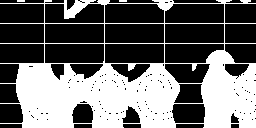} 
\end{center}
\vspace{0.02cm}      
\begin{center}
    \includegraphics [scale=0.8] {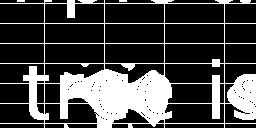}     
\end{center}
\vspace{0.02cm} 
\begin{center}
    \includegraphics [scale=0.8] {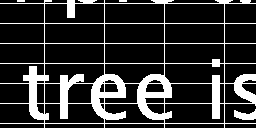}     
\end{center}
  \caption{ The original image (first row), the segmented foreground using least square fitting (second row), least absolute deviation (third row) and sparse decomposition (last row).}
\end{figure}

The reconstructed smooth model by these algorithms are shown in Figure 5. All methods used 10 DCT basis for representing the background and the same inlier threshold of 10 is used here.
\begin{figure}[5 h]
\begin{center}
    \includegraphics [scale=0.7] {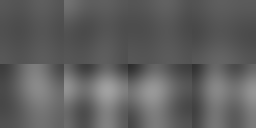} 
\end{center}     
\begin{center}
    \includegraphics [scale=0.7] {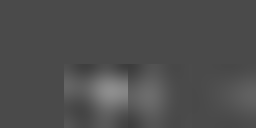}     
\end{center}
\begin{center}
    \includegraphics [scale=0.7] {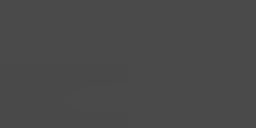}     
\end{center}
  \caption{ The reconstructed background layer using least square fitting (top image), least absolute deviation (middle image) and sparse decomposition (bottom image)}
\end{figure}

As we can see, the smooth model derived by LSF is largely affected by the foreground pixels. 
The ideal smooth model should have the same color as actual background (here gray), but because of the existence of many text pixels with white color the LSF solution tries to find a trade-off between fitting the texts and fitting the actual background, which results in inaccurate representation of either background or foreground in the regions around text. Therefore the regions around texts will have error larger than the inlier threshold and be falsely considered as the foreground pixels. 
The smooth model produced by the LAD approach was less affected by the foreground pixels than the LSF solution, because it minimizes the $\ell_1$ norm of the fitting error $s$. However, in blocks where there is a larger percentage of foreground pixels (bottom middle  and right regions),  LAD solution is still adversely affected by the foreground pixels. The SD approach yielded accurate solution in this example, because it considers  the $\ell_1$ norm of the fitting coefficient, $\alpha$, and the TV norm of $s$, in addition to  the  $\ell_1$ norm of $s$. Although the LAD solution leads to smaller $\ell_1$ norm of the fitting error, it also  leads to a much larger $\ell_1$ norm of $\alpha$ as well.  By minimizing all three terms, the SD solution obtains  a background model that uses predominantly only the DC basis, which represented the background accurately.

To confirm that the SD solution indeed has a smaller $\ell_1$ norm of $\alpha$, we show below the derived $\alpha$ values using each scheme. As we can see the derived $\alpha$ by SD has much smaller $\ell_0$ and $\ell_1$ norm than the other two.
\small
\begin{gather*}
\alpha^{LSF} \ = (7097,   -359,    19,   -882,    177,  -561,    863,  953,     113,   - 554) \\
\alpha^{LAD}= (5985,   -599,    201,  -859,  -13,   -96,    365,    39,    464,   -411) \\
\hspace{-2.9cm} \alpha^{SD}  \ \ = (4735,   -1,   0,   -4,   0,   -1,   0,    0,    0,    1)
\end{gather*}

\normalsize

\section{Overall Segmentation Algorithms}
We propose a segmentation algorithm that mainly depends on RANSAC/SD but it first checks if a block can be segmented using some simpler approaches and it goes to RANSAC/SD only if the block cannot be segmented using those approaches. These simple cases belong to one of these groups: pure background block, smoothly varying background only and text/graphic overlaid on constant background.
 
Pure background blocks are those in which  all pixels have similar intensities. These kind of blocks are common in screen content images. These blocks can be detected by looking at the standard deviation or  maximum absolute deviation of pixels' intensities. If the standard deviation is less than some threshold we declare that block as pure background.

Smoothly varying background only is a block in which the intensity variation over all pixels can be modeled well by a smooth function. Therefore we try to fit $K$ DCT basis to all pixels using least square fitting. If all pixels of that block can be represented with an error less than a predefined threshold, $\epsilon_{in}$, we declare it as smooth background.

The last group of simple cases is text/graphic overlaid on constant background. 
The images of this category usually have zero variance (or very small variances) inside each connected component.  
These images usually have a limited number of different colors in each block (usually less than 10) and the intensities in different parts are very different. We calculate the percentage of each different color in that block and the one with the highest percentage will be chosen as background and the other ones as foreground. 

When a block does not satisfy any of the above conditions, RANSAC/SD will be applied to separate the background and the foreground. If the segmentation is correct, the ratio of background pixels over the total number of pixels should be fairly large (greater than at least half ). When the ratio is small, the background of the block may be too complex to be presented by the adopted smooth function model. This may also happen when the block sits at the intersection of two smooth backgrounds. To overcome these problems, we apply the proposed method recursively using a quadtree structure. When the inlier ratio of the current block is less than $\epsilon_2$, we divide it into 4 smaller blocks and apply the proposed algorithm on each smaller block, until the smallest block size is reached.\\
The overall segmentation algorithm is summarized as follows:
\begin{enumerate}
\item Starting with block size $N=64$, if the standard deviation of pixels' intensities is less than $\epsilon_1$ (i.e. pixels in the block have very similar color intensity), then declare the entire block as background.
If not, go to the next step;
\item Perform least square fitting using all pixels. If all pixels can be predicted with an error less than $\epsilon_{in}$, declare the entire block as background. If not, go to the next step;
\item If the number of different colors (in terms of the luminance value) is less than $T_1$ and the intensity range is above $R$, declare the block as text/graphics over a constant background and find the background as the color in that block with the highest percentage of pixels. If not, go to the next step;
\item Use RANSAC/SD to separate background and foreground using the luminance component only. Verify that the corresponding chrominance components of background pixels can also be fitted using $K$ basis functions with an error less than $\epsilon_{in}$. If some of them cannot be fitted with this error, remove them from inliers set.
If the percentage of inliers is more than a threshold $\epsilon_2$ or $N$ is equal to 8, the inlier pixels are selected as background. If not go to the next step;
\item Decompose the current block of size $N \times N$ into 4 smaller blocks of size $\frac{N}{2} \times \frac{N}{2}$ and run the segmentation algorithm for all of them. Repeat until $N=8$.
\end{enumerate}

To show the advantage of quad-tree decomposition, we provide an example of the segmentation map without and with quad-tree decomposition in Figure 6. 
As we can see, using quadtree decomposition we get much better result compared to the case with no decomposition. When we do not allow a $64 \times 64$ block to be further divided, only a small percentage of pixels can be represented well by a smooth function, leaving many pixels as foreground. 
It is worth mentioning that the gray region on the top of the image is considered as foreground in the segmentation result without using quadtree decomposition. This is because the first row of $64 \times 64$ blocks contain two smooth background regions with relatively equal size.

\begin{figure}[6 h]
\begin{center}
    \includegraphics [scale=0.27] {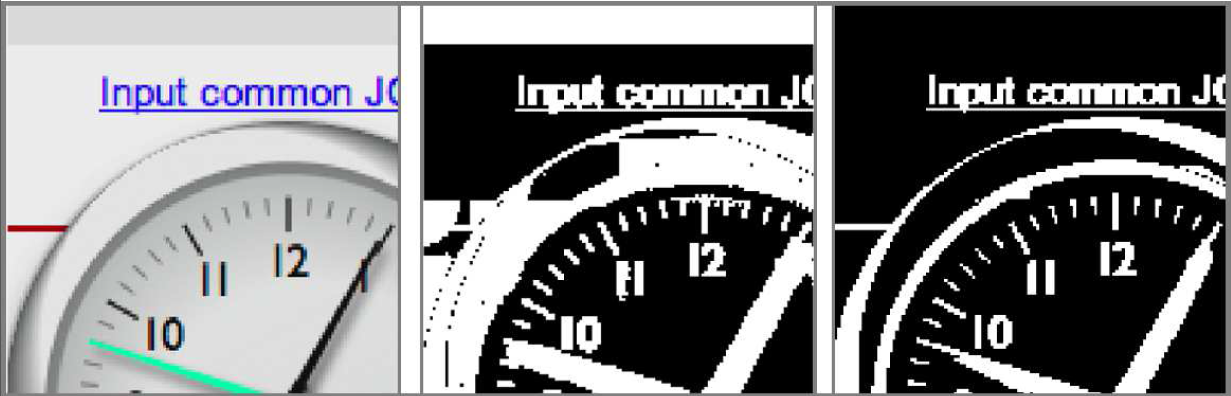} 
\end{center}
  \caption{Segmentation result for a sample image, middle and right images denote foreground map without and with quad-tree decomposition  using the RANSAC as the core algorithm. }
\end{figure}

\section{Experimental Results}
To enable rigorous evaluation of different algorithms, we have generated an annotated dataset consisting of 328 image blocks of size $64\times 64$, extracted from sample frames from HEVC test sequences for screen content coding \cite{SCC_data}. 
The ground truth foregrounds for these images are extracted manually by the author and then refined independently by another expert. This dataset is publicly available at \cite{our_dataset}.

Table 1 summarizes the parameter choices in the proposed algorithms.
The largest block size is chosen to be $N$=64, which is the same as the largest CU size in HEVC standard.
The thresholds used for preprocessing (steps 1-3) should be chosen conservatively to avoid segmentation errors. In our simulations, we have chosen them as $\epsilon_1=3$, $T_1=10$ and $R=50$, which achieved a good trade off between computation speed and segmentation accuracy. 
For the RANSAC algorithm, the maximum number of iteration is chosen to be 200.
For the sparse decomposition algorithm, the weight parameters in the objective function are tuned by testing on a validation set and are set to be $\lambda_1=10$ and $\lambda_2=4$.
The ADMM algorithm described in Algorithm 1 is implemented in MATLAB, which the code available in \cite{our_dataset}.
The number of iteration for ADMM is chosen to be 50 and the parameter $\rho_1$, $\rho_2$ and $\rho_3$ are all set to 1 as suggested in \cite{boyd}. 

\begin{table}[h]
\centering
  \caption{Parameters of our implementation}
  \centering
\begin{tabular}{|m{5cm}|m{1.2cm}|m{1.2cm}|}
\hline
Parameter description &  \ Notation & \ \ \  Value\\
\hline
Maximum block size & \ \ \ \ \ $N$ & \ \ \ \ \  64 \\
\hline
Inlier distortion threshold & \ \ \ \ \ $\epsilon_{in}$ & \ \ \ \ \ 10 \\ 
\hline
Background standard deviation threshold & \ \ \ \ \ $\epsilon_{1}$ & \ \ \ \ \ 3 \\
\hline
Qaud-tree decomposition threshold & \ \ \ \ \ $\epsilon_{2}$ & \ \ \ \ \ 0.5 \\
 \hline
Max number of colors for text over constant background & \ \ \ \ \ $T_1$ & \ \ \ \ \ 10 \\ 
\hline
Min intensity range for text over constant background & \ \ \ \ \ $R$ & \ \ \ \ \ 50 \\
\hline
Sparsity weight in SD algorithm & \ \ \ \ \ $\lambda_1$  & \ \ \ \  \ 10 \\
\hline
Total variation weight in SD algorithm & \ \ \ \ \ $\lambda_2$  & \ \ \ \ \ \ 4 \\
\hline
\end{tabular}
\label{TblComp}
\end{table}

To find the number of DCT basis functions, $K$, and inlier threshold, $\epsilon_{in}$, for RANSAC and sparse decomposition, we did a grid search over pairs of these parameters, in the range of 6 to 10 for $K$ and 5 to 15 for $\epsilon_{in}$, on some training images, and then chose the one which achieved the best result in terms of average F1-score (defined in Eq. 12). 
The parameter values that resulted in the best F1-score on our training images are shown in Table II.
\begin{table}[h]
\centering
  \caption{The chosen values for the inlier threshold and number of bases}
  \centering
\begin{tabular}{|m{3.2cm}|m{1.2cm}|m{1.2cm}|m{1.2cm}|}
\hline
Segmentation Algorithm  &  \ \  \ LAD & \ RANSAC & \ \ \ \ SD \\
\hline
Inlier threshold & \ \ \ \ 10 & \ \ \ \ 10 & \ \ \ \  10 \\
\hline
Number of bases & \ \ \ \ \ 6 & \ \ \ \ 10 & \ \ \ \ 10 \\
\hline
\end{tabular}
\label{TblComp}
\end{table}

Before showing the segmentation result of the proposed algorithms on the test images, we illustrate how the segmentation result varies by changing different parameters in RANSAC algorithm. 
The sparse decomposition algorithm would also have the same behavior.

To evaluate the effect of the distortion threshold, $\epsilon_{in}$, for inlier pixels in the final segmentation result, we show the foreground map derived by several different thresholds in Figure 7.  As we can see by increasing the threshold more and more pixels are considered as background.

\begin{figure}[h]
        \centering
        \vspace{-0.1cm}
        \begin{subfigure}[b]{0.24\textwidth}
                \includegraphics[width=\textwidth]{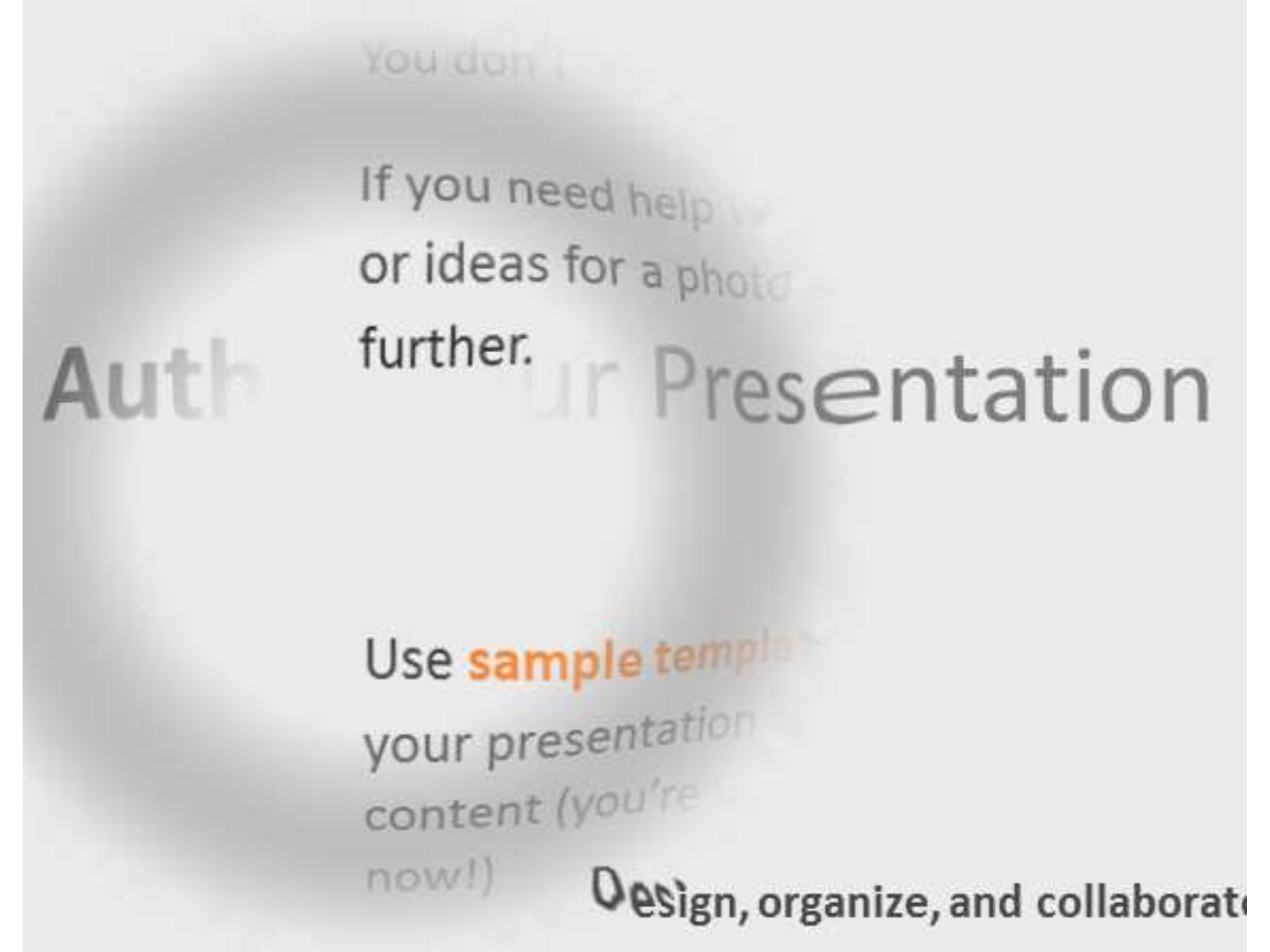}
                                \vspace{-0.5cm}
          \hspace{-2.5cm}    
        \end{subfigure}%
        ~ 
        \begin{subfigure}[b]{0.24\textwidth}
                \includegraphics[width=\textwidth]{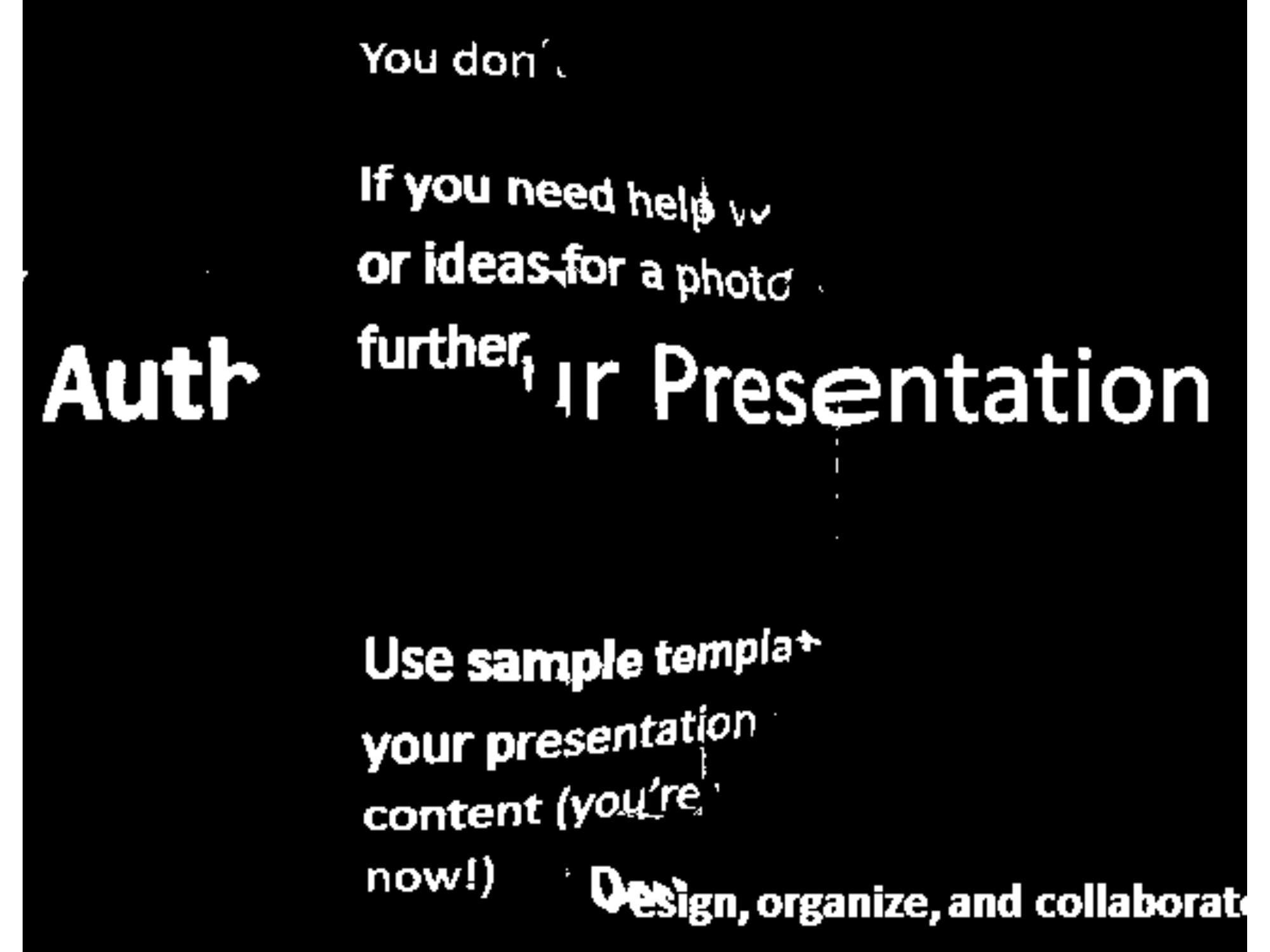}
                \vspace{-0.5cm}
            \hspace{-3cm} 
        \end{subfigure} \\[1ex]
        \begin{subfigure}[b]{0.24\textwidth}
                \includegraphics[width=\textwidth]{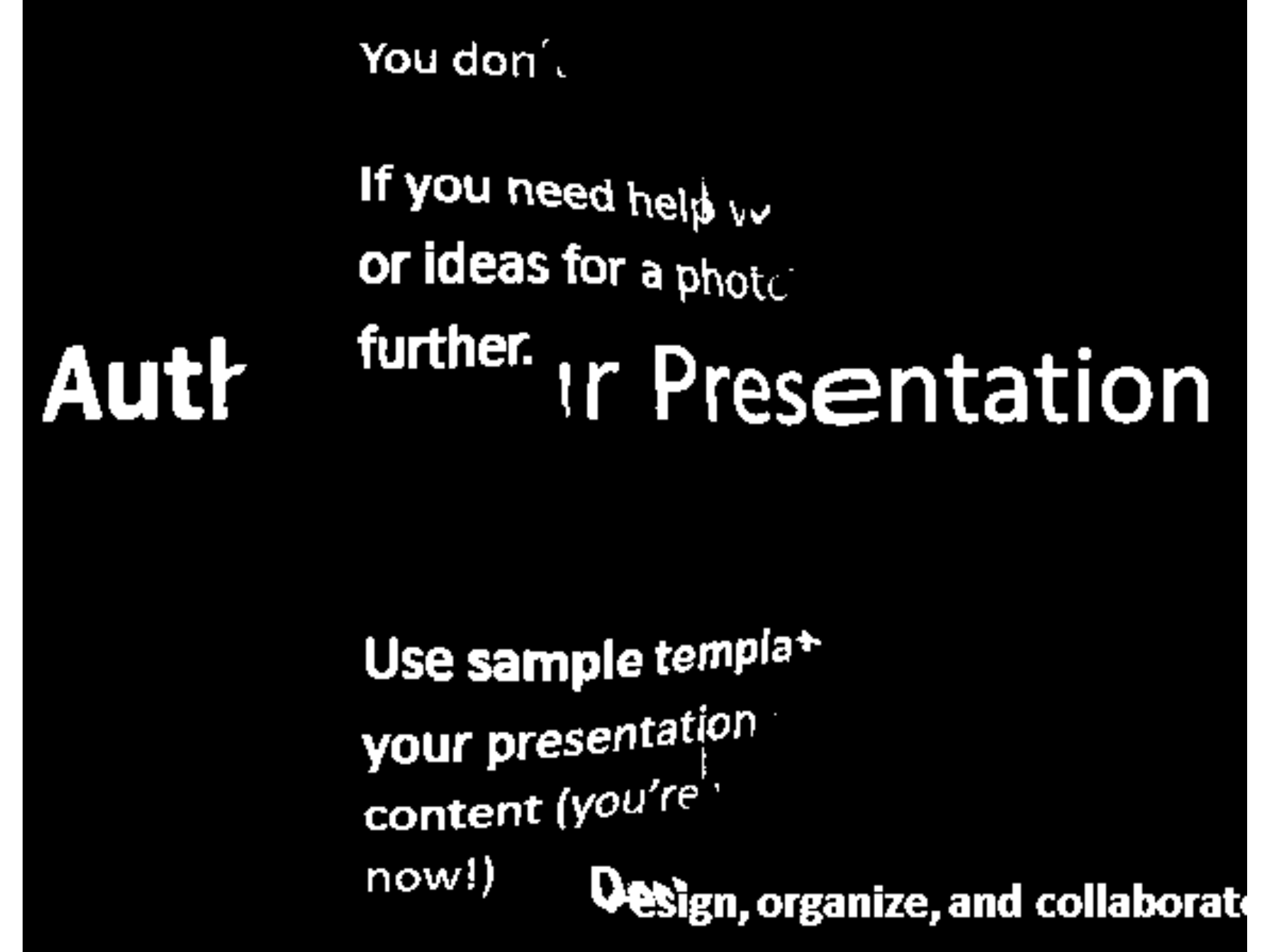}
                                \vspace{-0.5cm}
          \hspace{-2.5cm}    
        \end{subfigure}%
        ~ 
        \begin{subfigure}[b]{0.24\textwidth}
                \includegraphics[width=\textwidth]{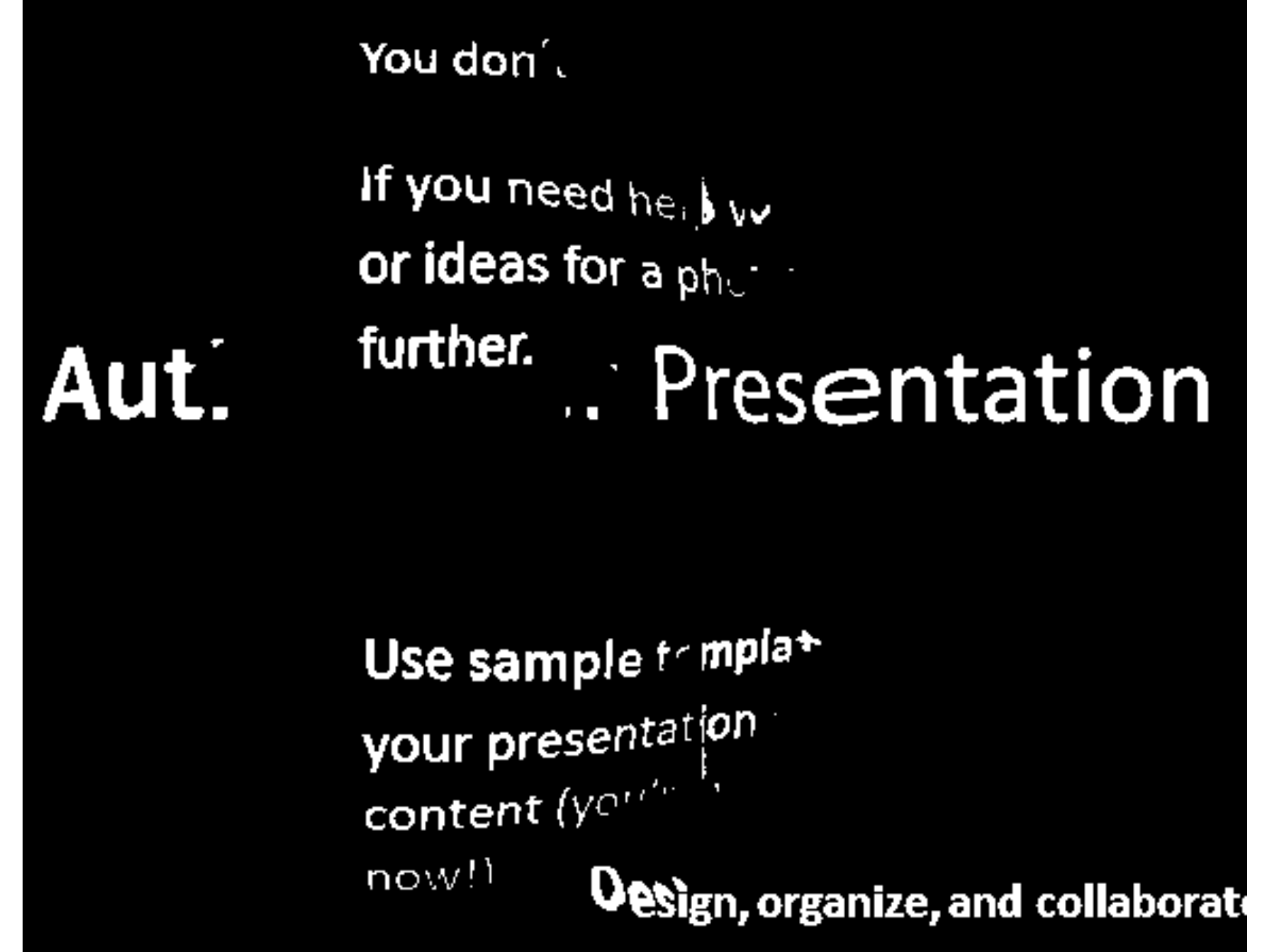}
                \vspace{-0.5cm}
            \hspace{-3cm} 
        \end{subfigure} \\[1ex]
        \begin{subfigure}[b]{0.24\textwidth}
                \includegraphics[width=\textwidth]{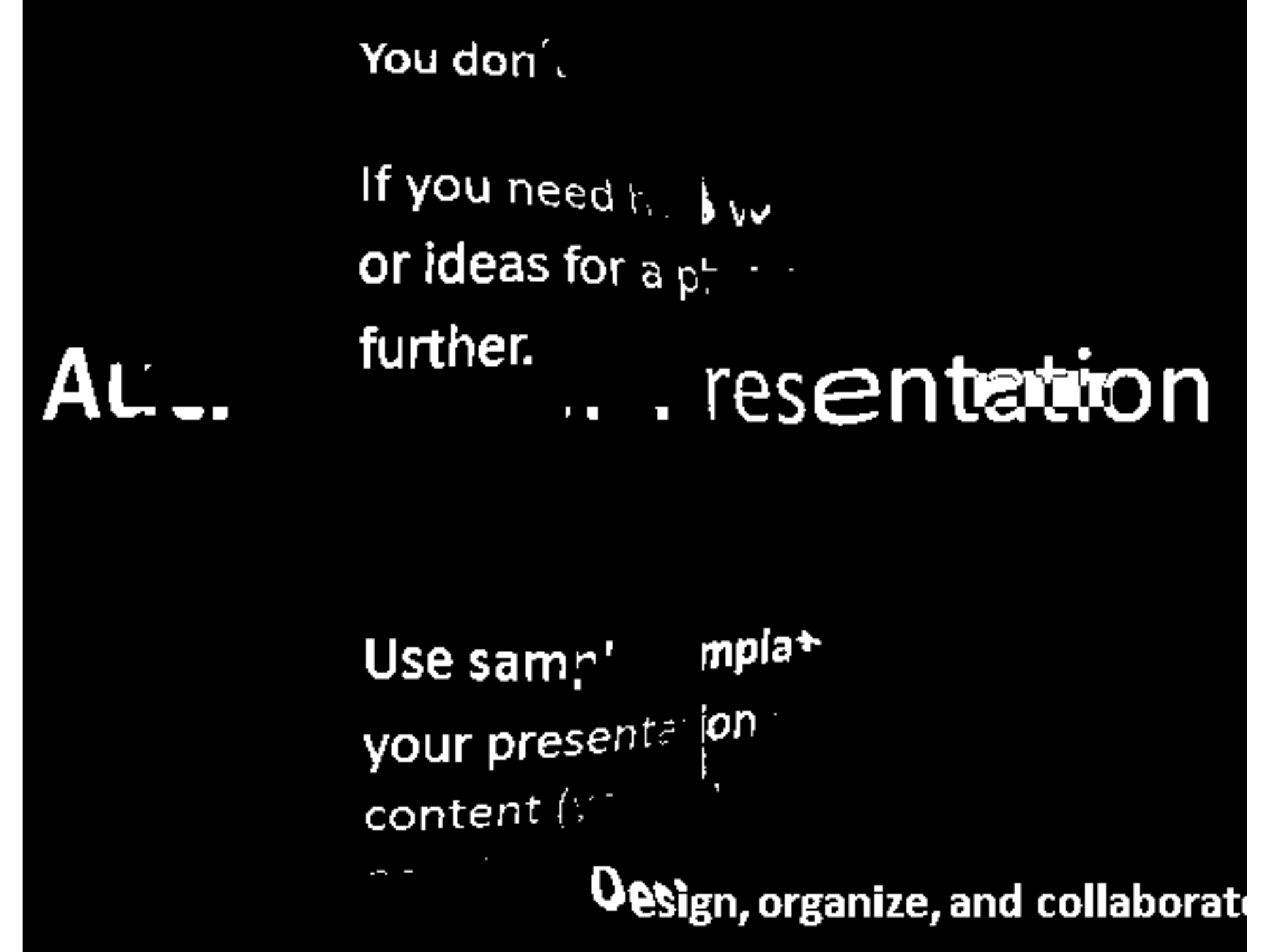}
                                \vspace{-0.5cm}
          \hspace{-2.5cm}    
        \end{subfigure}%
        ~ 
        \begin{subfigure}[b]{0.24\textwidth}
                \includegraphics[width=\textwidth]{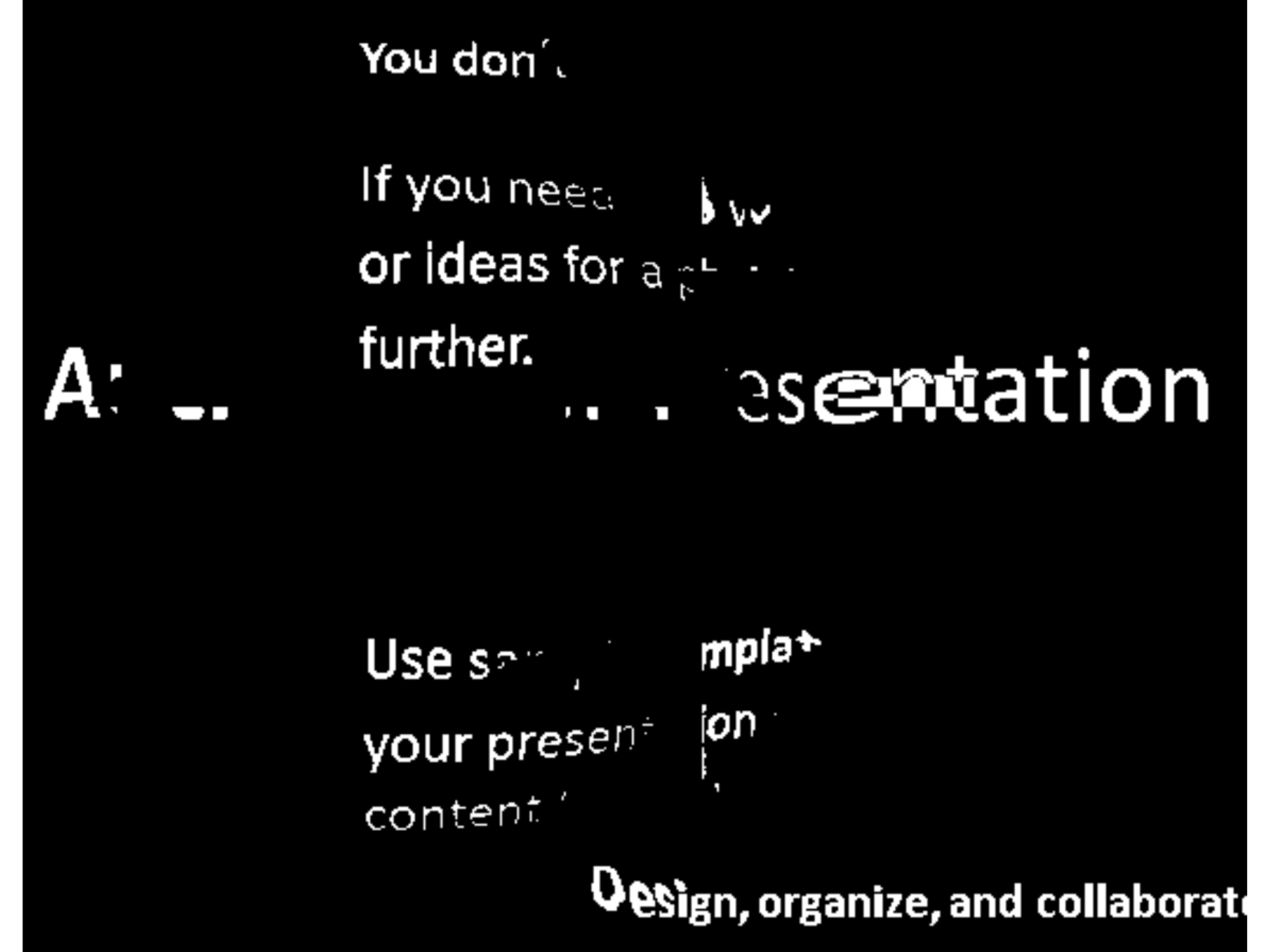}
                \vspace{-0.5cm}
            \hspace{-3cm} 
        \end{subfigure}%
                \caption{Segmentation results of the RANSAC method by varying the inlier threshold $\epsilon_{in}$. The foreground maps from left to right and top to bottom are obtained with $\epsilon_{in}$ setting to 5, 10, 25, 35, and 45, respectively.}
\end{figure}

To assess the effect of the number of basis, $K$, in the final segmentation result, we show the foreground map derived by several different number of basis functions using the RANSAC method in Figure 8.

\begin{figure}[h]
        \centering
        \vspace{-0.1cm}
        \begin{subfigure}[b]{0.24\textwidth}
                \includegraphics[width=\textwidth]{2Original_Image-eps-converted-to.pdf}
                                \vspace{-0.5cm}
          \hspace{-2.5cm}    
        \end{subfigure}%
        ~ 
        \begin{subfigure}[b]{0.24\textwidth}
                \includegraphics[width=\textwidth]{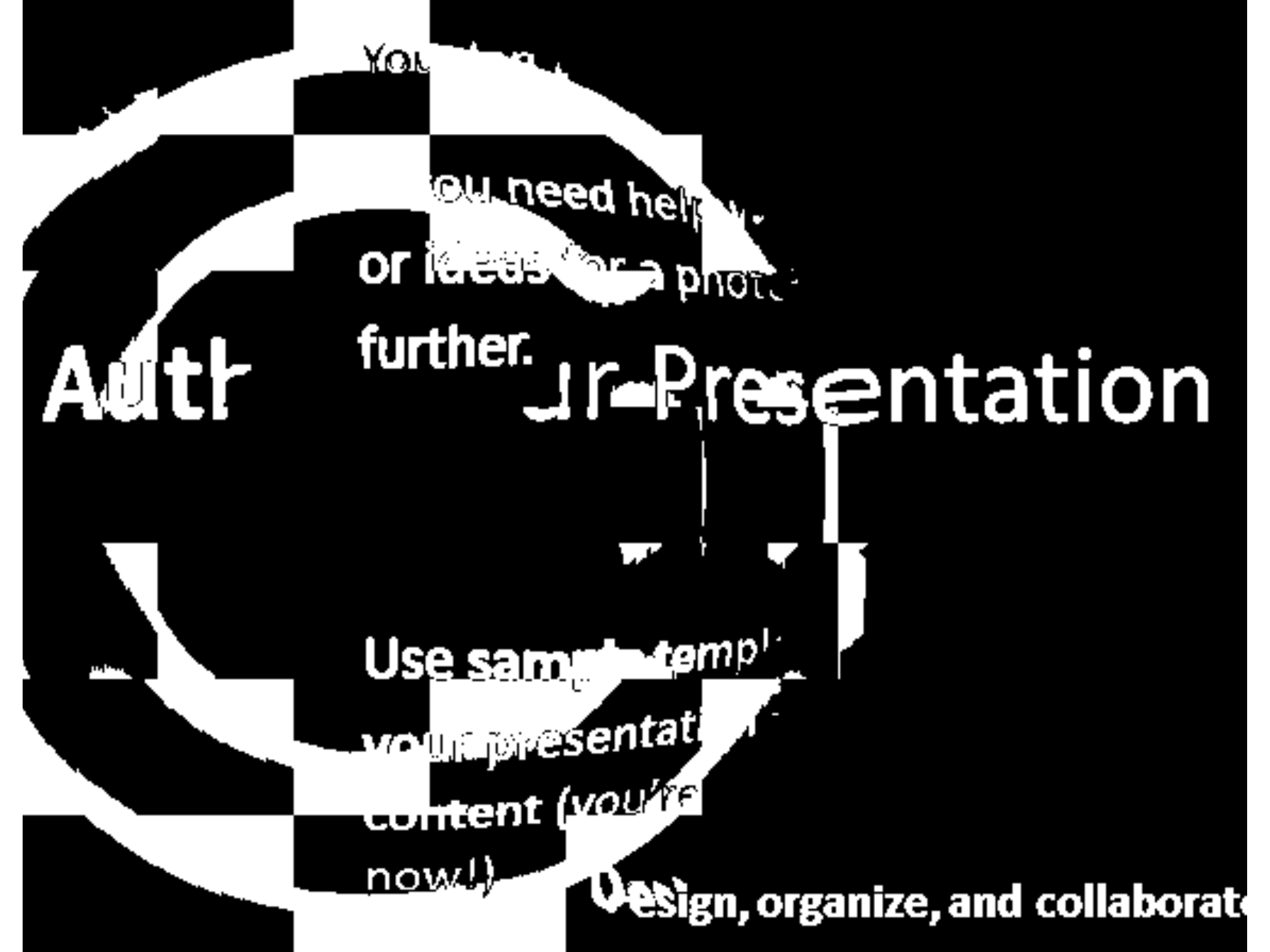}
                \vspace{-0.5cm}
            \hspace{-3cm} 
        \end{subfigure} \\[1ex]
        \begin{subfigure}[b]{0.24\textwidth}
                \includegraphics[width=\textwidth]{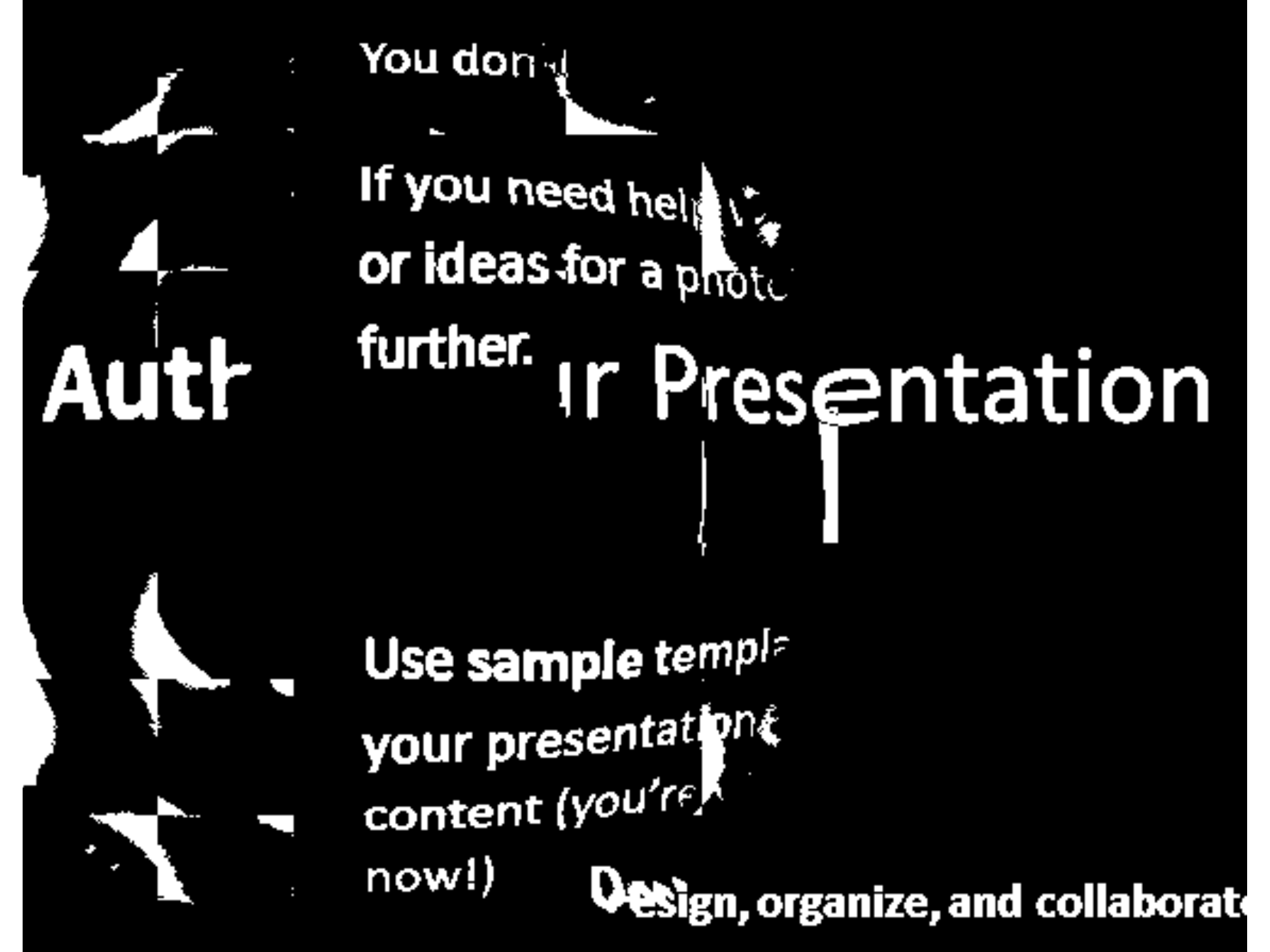}
                                \vspace{-0.5cm}
          \hspace{-2.5cm}    
        \end{subfigure}%
        ~ 
        \begin{subfigure}[b]{0.24\textwidth}
                \includegraphics[width=\textwidth]{2_map_basis10-eps-converted-to.pdf}
                \vspace{-0.5cm}
            \hspace{-3cm} 
        \end{subfigure} \\[1ex]
        \begin{subfigure}[b]{0.24\textwidth}
                \includegraphics[width=\textwidth]{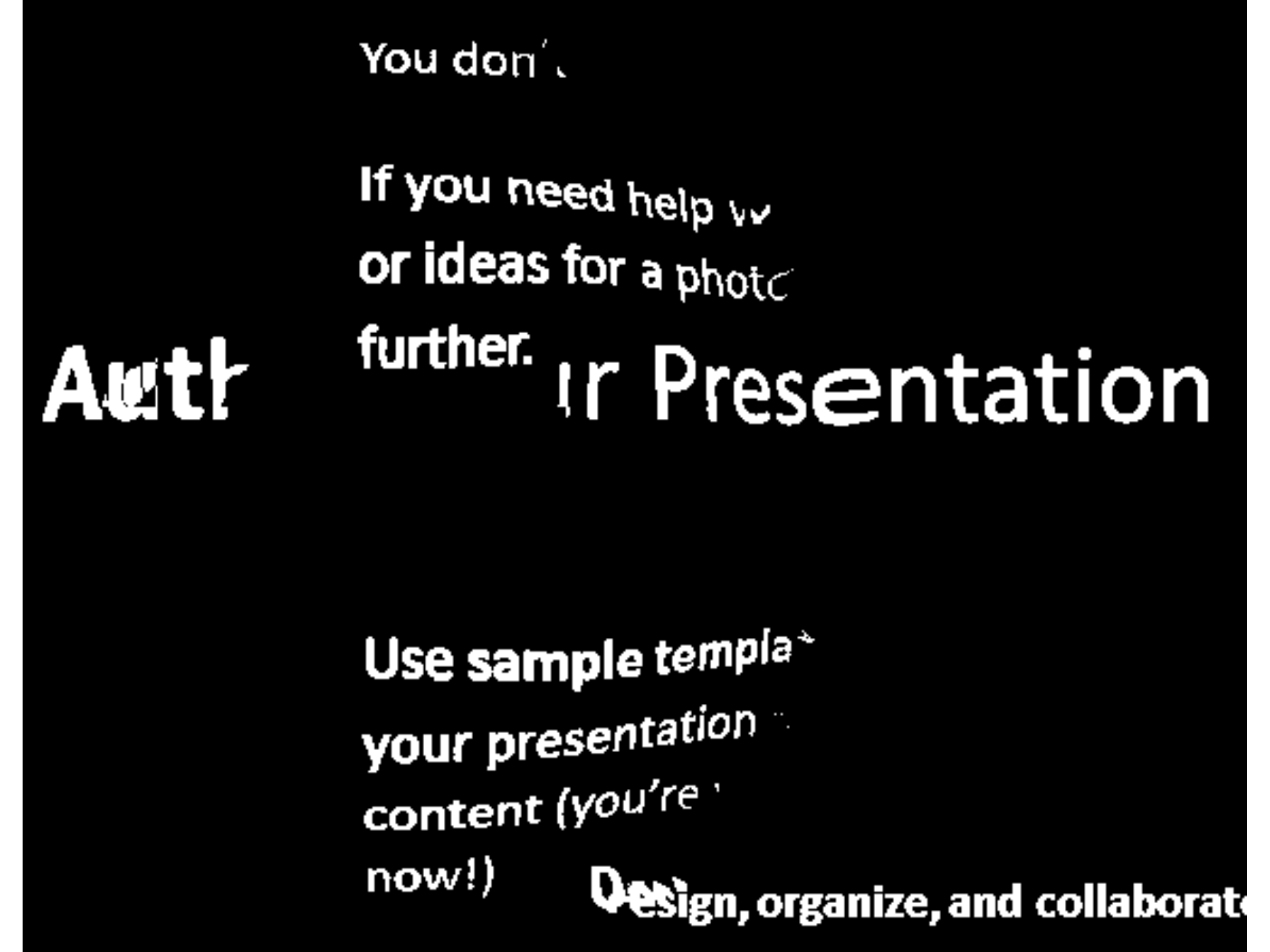}
                                \vspace{-0.5cm}
          \hspace{-2.5cm}    
        \end{subfigure}%
        ~ 
        \begin{subfigure}[b]{0.24\textwidth}
                \includegraphics[width=\textwidth]{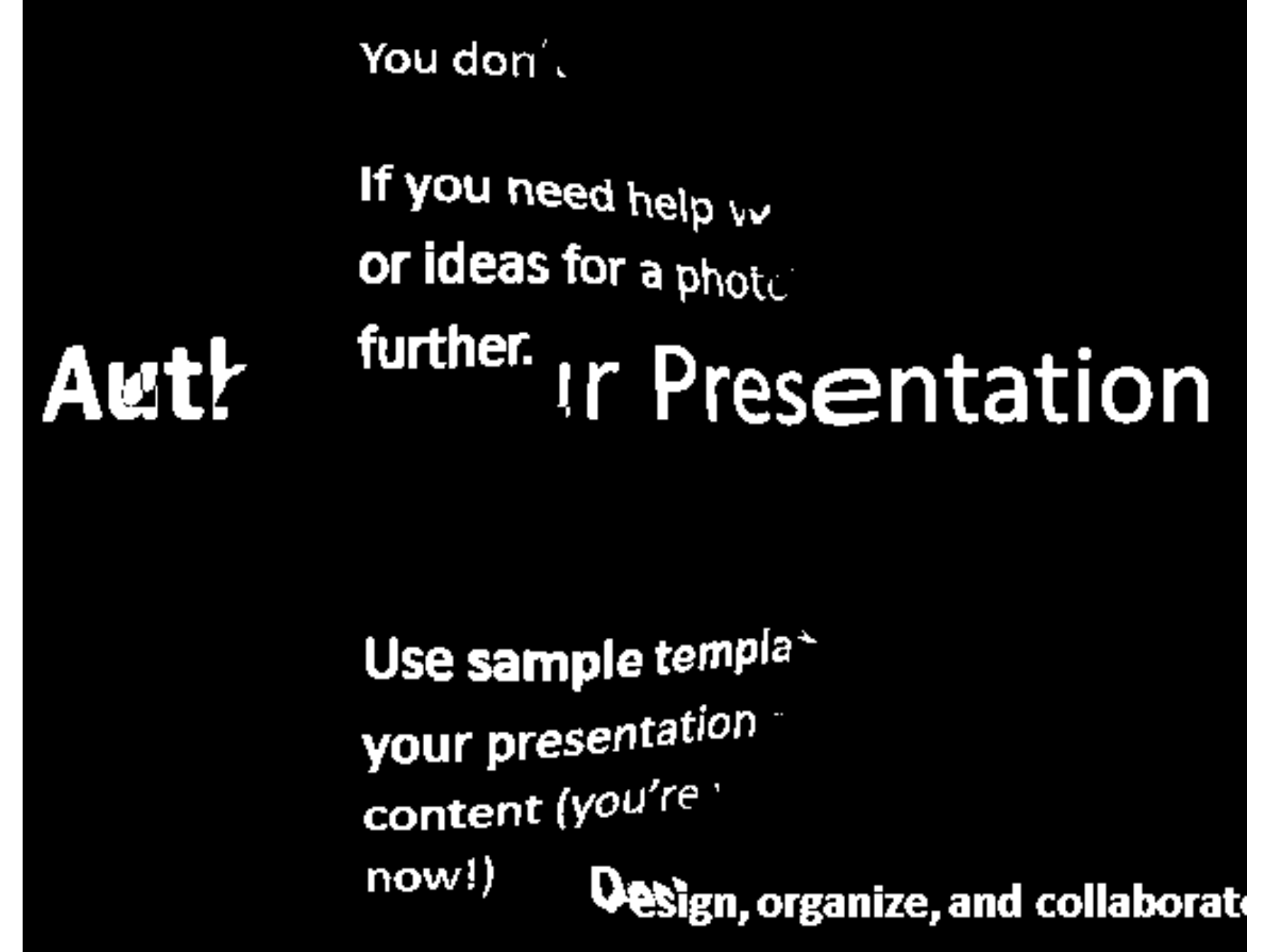}
                \vspace{-0.5cm}
            \hspace{-3cm} 
        \end{subfigure}%
                \caption{Segmentation results of the RANSAC method using different number of basis functions. The foreground maps from left to right and top to bottom are obtained with 2, 5, 10, 15, and 20 basis functions, respectively.}
\end{figure}

To illustrate the smoothness of the background layer and its suitability for being coded with transform-based coding, the filled background layer of a sample image is presented in Figure 9. The background holes (those pixels that belong to foreground layers) are filled by the predicted value using the smooth model, which is obtained using the least squares fitting to the detected background pixels. As we can see the background layer is very smooth and does not have any sharp edges.

\begin{figure}[3 h]
\begin{center}
\hspace{-0.1cm}
    \includegraphics [scale=0.3] {2Original_Image-eps-converted-to.pdf}
    \vspace{0.3cm}
\hspace{-0.18cm}	\includegraphics [scale=0.3] {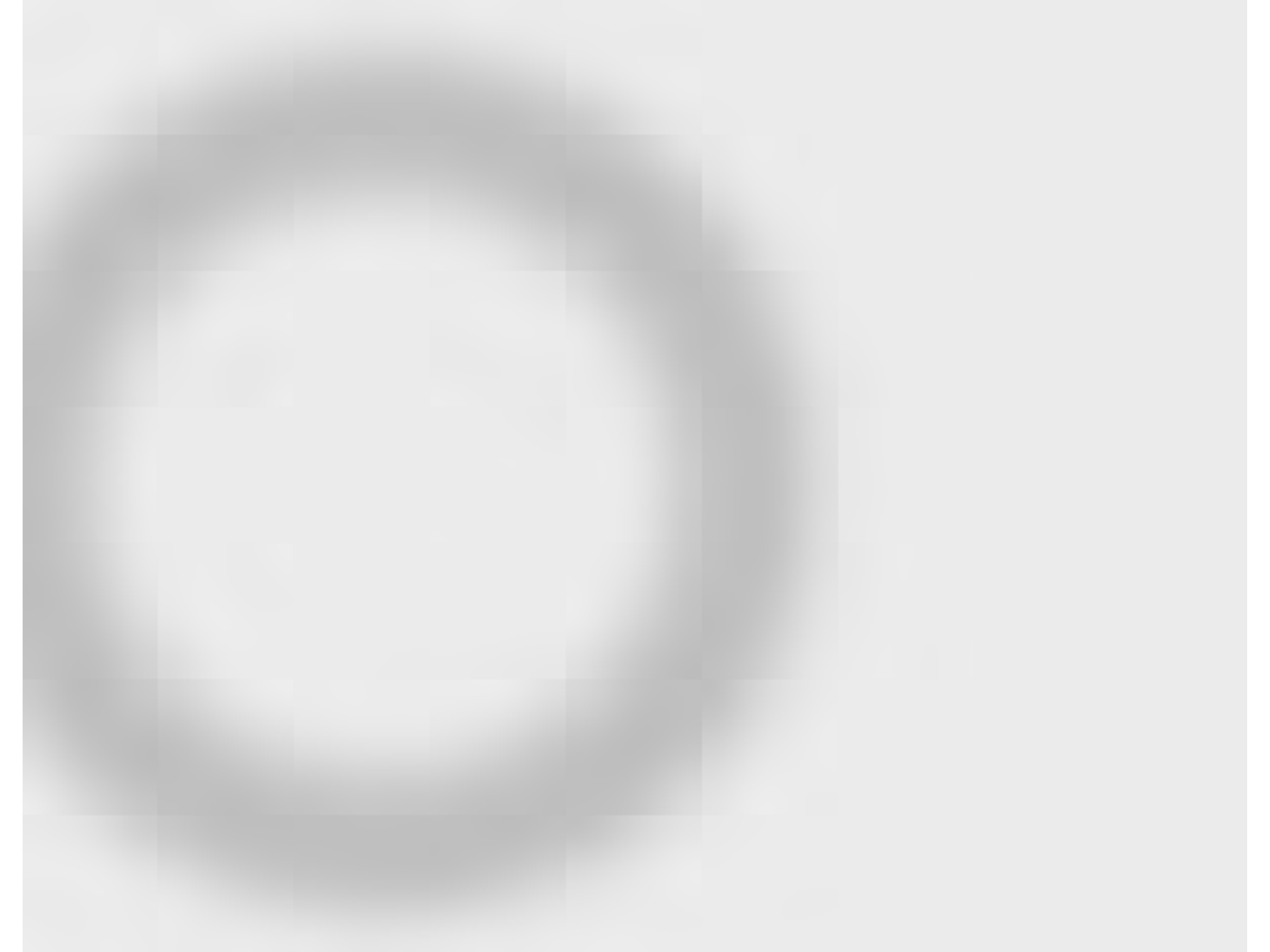}
\end{center}
  \caption{The reconstructed background of an image }
\end{figure}

We compare the proposed algorithms with hierarchical k-means clustering used in DjVu, SPEC, least square fitting, and LAD algorithms.
For SPEC, we have adapted the color number threshold and the shape primitive size threshold from the default value given in \cite{SPEC}  when necessary to give more satisfactory result. Furthermore, for blocks classified as text/graphics based on the color number, we segment the most frequent color and any similar color to it (i.e. colors whose distance from most frequent color is less than 10 in luminance) in the current block as background and the rest as foreground.
We have also provided a comparison with least square fitting algorithm result, so that the reader can see the benefit of minimizing the $\ell_0$ and $\ell_1$ norm over minimizing the $\ell_2$ norm.

To provide a numerical comparison between the proposed scheme and previous approaches, we have calculated the average precision and recall and F1 score (also known as F-measure) \cite{PRF} achieved by different segmentation algorithms over this dataset. 
The precision and recall are defined as:
\begin{gather}
 Precision= \frac{\text{TP}}{\text{TP+FP}} \ , 
\ \ \ \ Recall= \frac{\text{TP}}{\text{TP+FN}} 	\ 	, 
\end{gather}
where $\text{TP}, \text{FP}$ and $\text{FN}$ denote true positive, false positive and false negative respectively. In our evaluation, we treat a foreground pixel as positive. A pixel that is correctly identified as foreground (compared to the manual segmentation) is considered true positive. The same holds for false negative and false positive. 

The balanced F1 score is defined as the harmonic mean of precision and recall, i.e. 
\begin{gather}
\text{F1}= 2 \ \frac{Precision \times Recall}{ Precision+Recall} 
\end{gather}
The average precision, recall and F1 scores by different algorithms are given in Table III. 
As can be seen, the two proposed schemes achieve  much higher precision and recall than the DjVu and SPEC algorithms, and also provide noticeable gain over our prior LAD approach. Among the two proposed methods, sparse decomposition based algorithm achieved high precision, but lower recall than the RANSAC algorithm. 

\begin{table}[h]
\centering
  \caption{Segmentation accuracies of different algorithms}
  \centering
\begin{tabular}{|m{3.4cm}|m{1.2cm}|m{1.2cm}|m{1.2cm}|}
\hline
Segmentation Algorithm  &  \  \ Precision & \ \  Recall & \  F1 score\\
\hline
SPEC \cite{SPEC} & \ \ \ 50\% & \ \ \  64\% & \ \ \ 56\% \\
\hline
 DjVu \cite{djvu} & \ \ \ 64\% & \ \ \ 69\% & \ \ \ 66\% \\
\hline
 Least square fitting & \ \ \ 79\% & \ \ \ 60\% & \ \  \ 68\% \\ 
\hline
 Least Absolute Deviation \cite{LAD} & \ \ \  90.5\% & \ \ \  87\% & \ \ \  88.7\% \\
\hline
 RANSAC based segmentation & \ \ \ 91\%  & \ \ \ 90\%  & \ \ \  90.4\%\\
 \hline
 Sparse Decomposition Alg. & \ \ \ 94\%  & \ \ \ 87.2\%  & \ \ \  90.5\%\\
\hline
\end{tabular}
\label{TblComp}
\end{table}

The results for 5 test images (each consisting of multiple 64x64 blocks) are shown in Figure 10. Each test image is a small part of a frame from a HEVC SCC test sequence.
\begin{figure*}
        \centering
        \vspace{-0.5cm}
        \begin{subfigure}[b]{0.18\textwidth}
                \includegraphics[width=\textwidth]{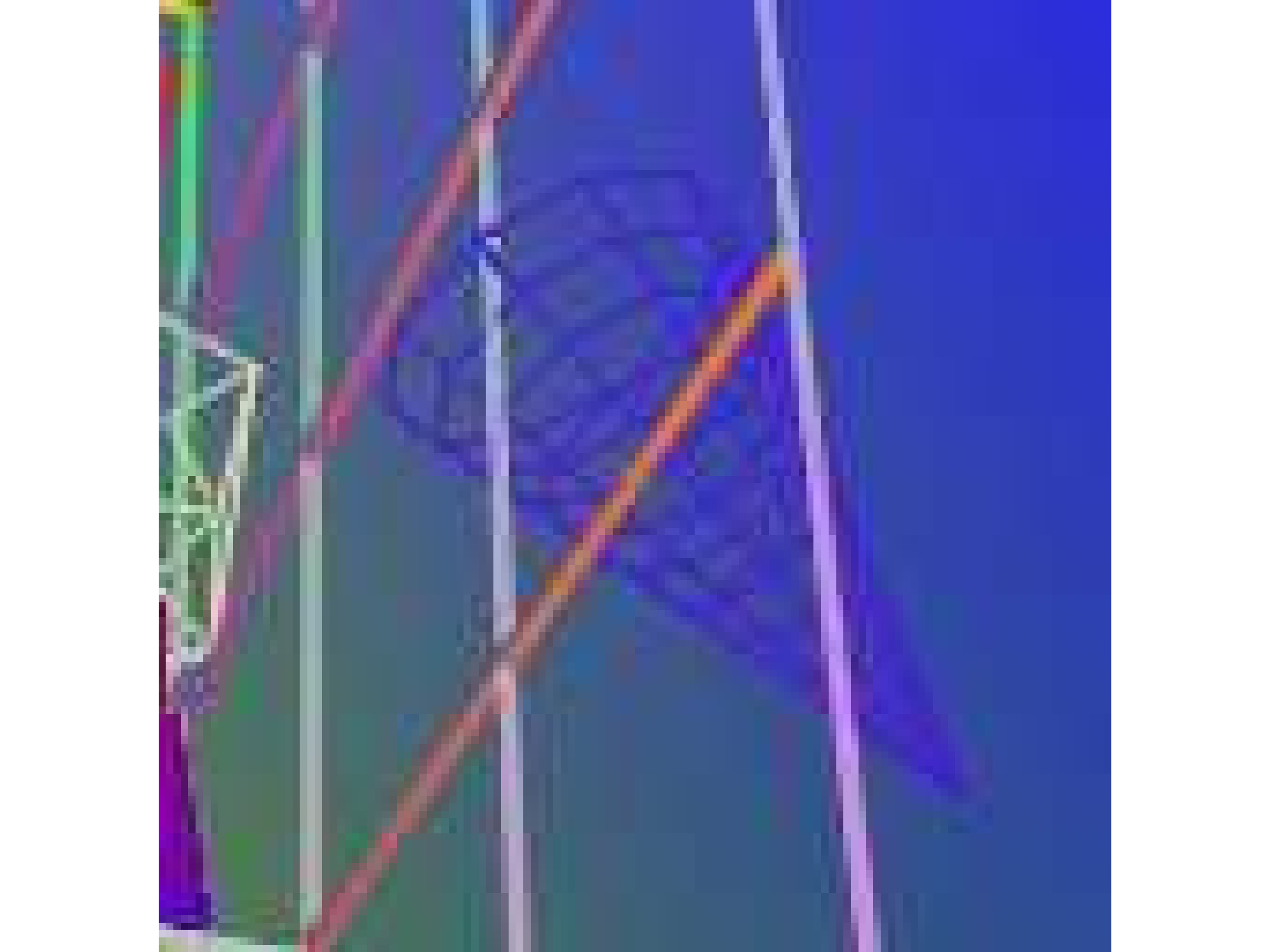}
                                \vspace{-0.5cm}
          \hspace{-2.5cm}    
        \end{subfigure}%
        ~ 
        \begin{subfigure}[b]{0.18\textwidth}
                \includegraphics[width=\textwidth]{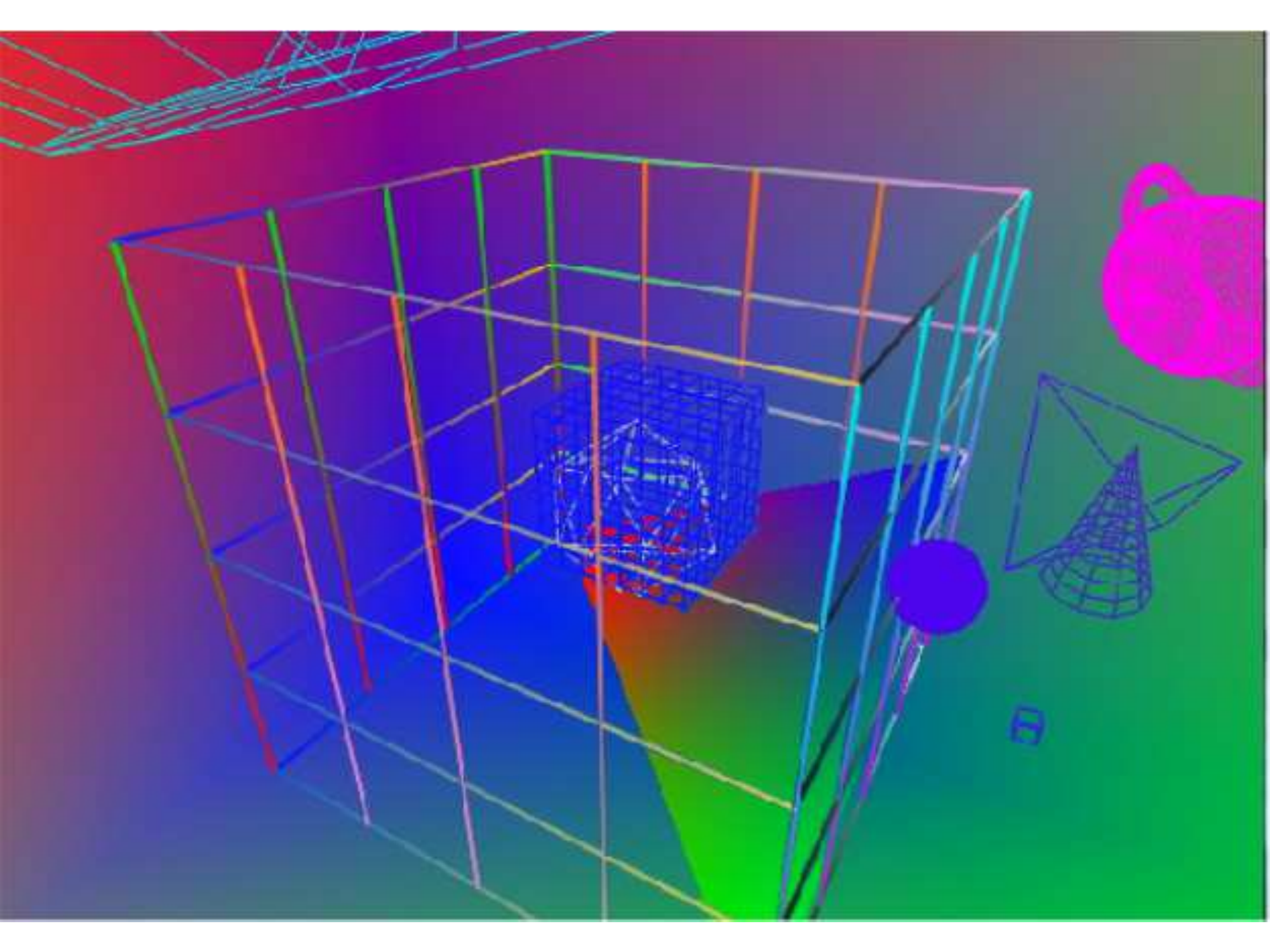}
                \vspace{-0.5cm}
            \hspace{-3cm} 
        \end{subfigure}%
        ~ 
        \begin{subfigure}[b]{0.18\textwidth}
                \includegraphics[width=\textwidth]{8Original_Image-eps-converted-to.pdf}
                \vspace{-0.45cm}
            \hspace{-3cm} 
        \end{subfigure}%
        \begin{subfigure}[b]{0.18\textwidth}
			~ 
                \includegraphics[width=\textwidth]{2Original_Image-eps-converted-to.pdf}
                \vspace{-0.45cm}
            \hspace{-5cm} 
        \end{subfigure}%
        \begin{subfigure}[b]{0.18\textwidth}
                \includegraphics[width=\textwidth]{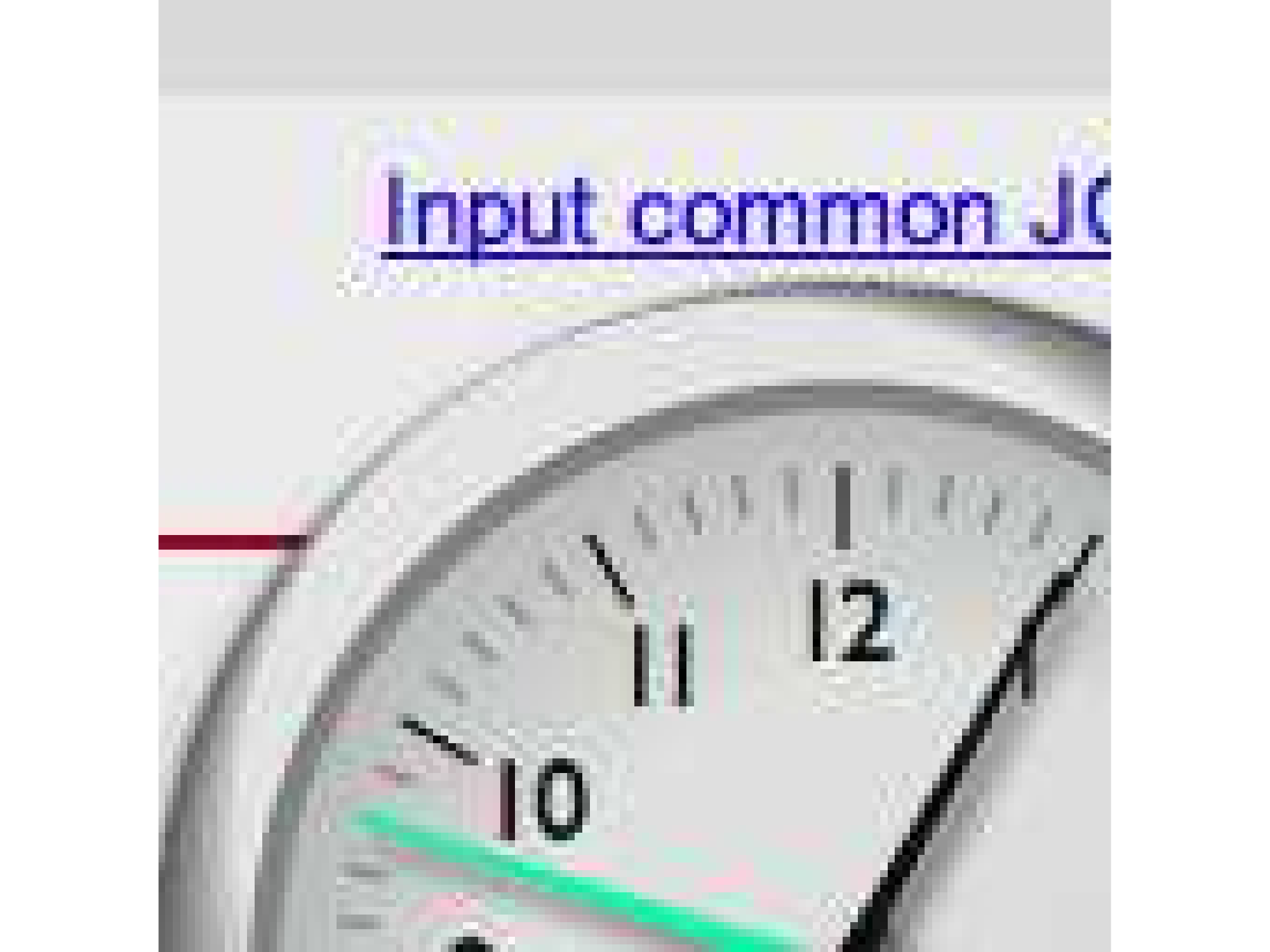}
                 \vspace{-0.45cm}
              \hspace{-4.8cm}
        \end{subfigure}
         \\[1ex]
        \begin{subfigure}[b]{0.18\textwidth}
                \includegraphics[width=\textwidth]{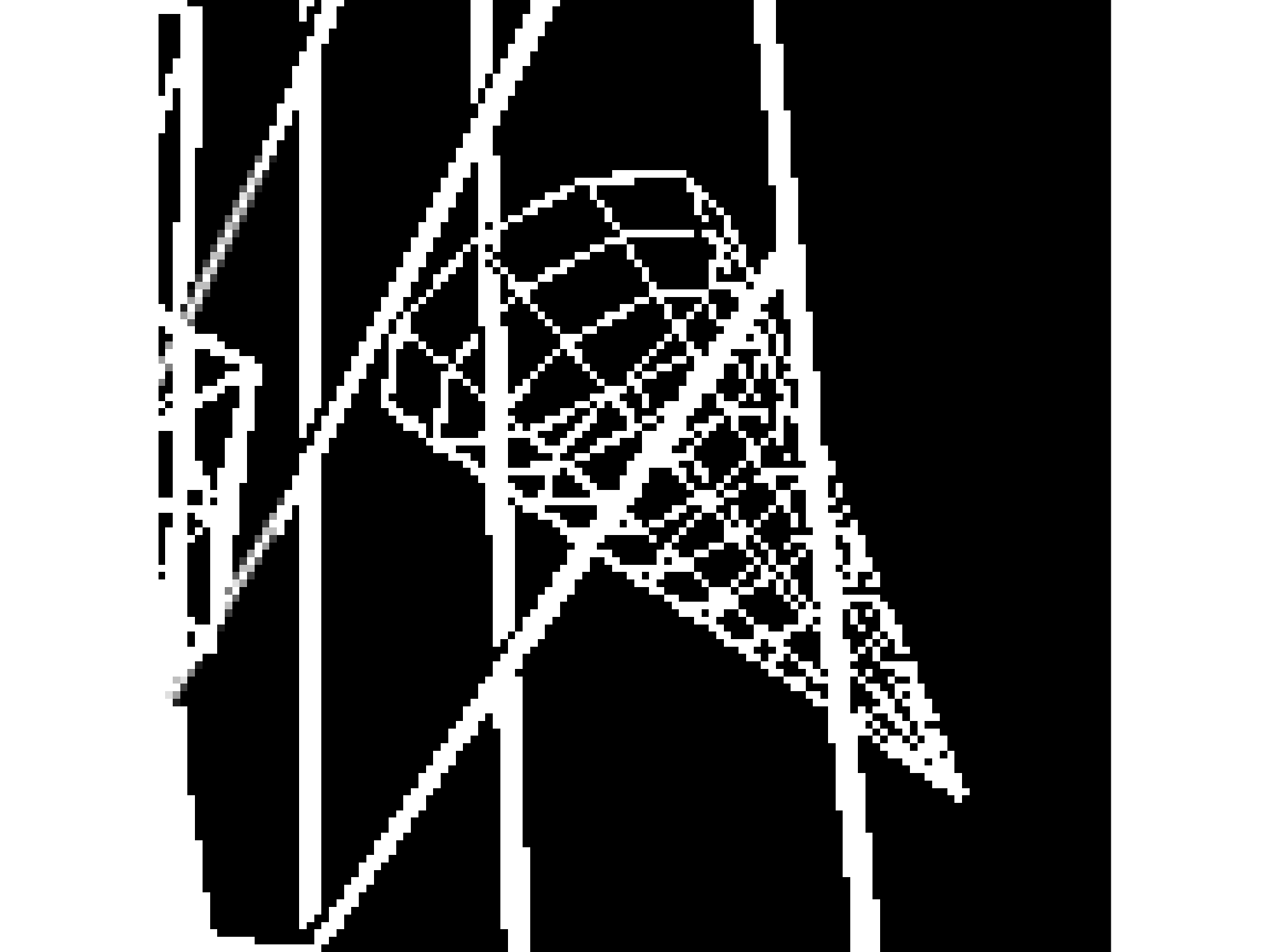}
                                \vspace{-0.5cm}
          \hspace{-2.5cm}    
        \end{subfigure}%
        ~ 
        \begin{subfigure}[b]{0.18\textwidth}
                \includegraphics[width=\textwidth]{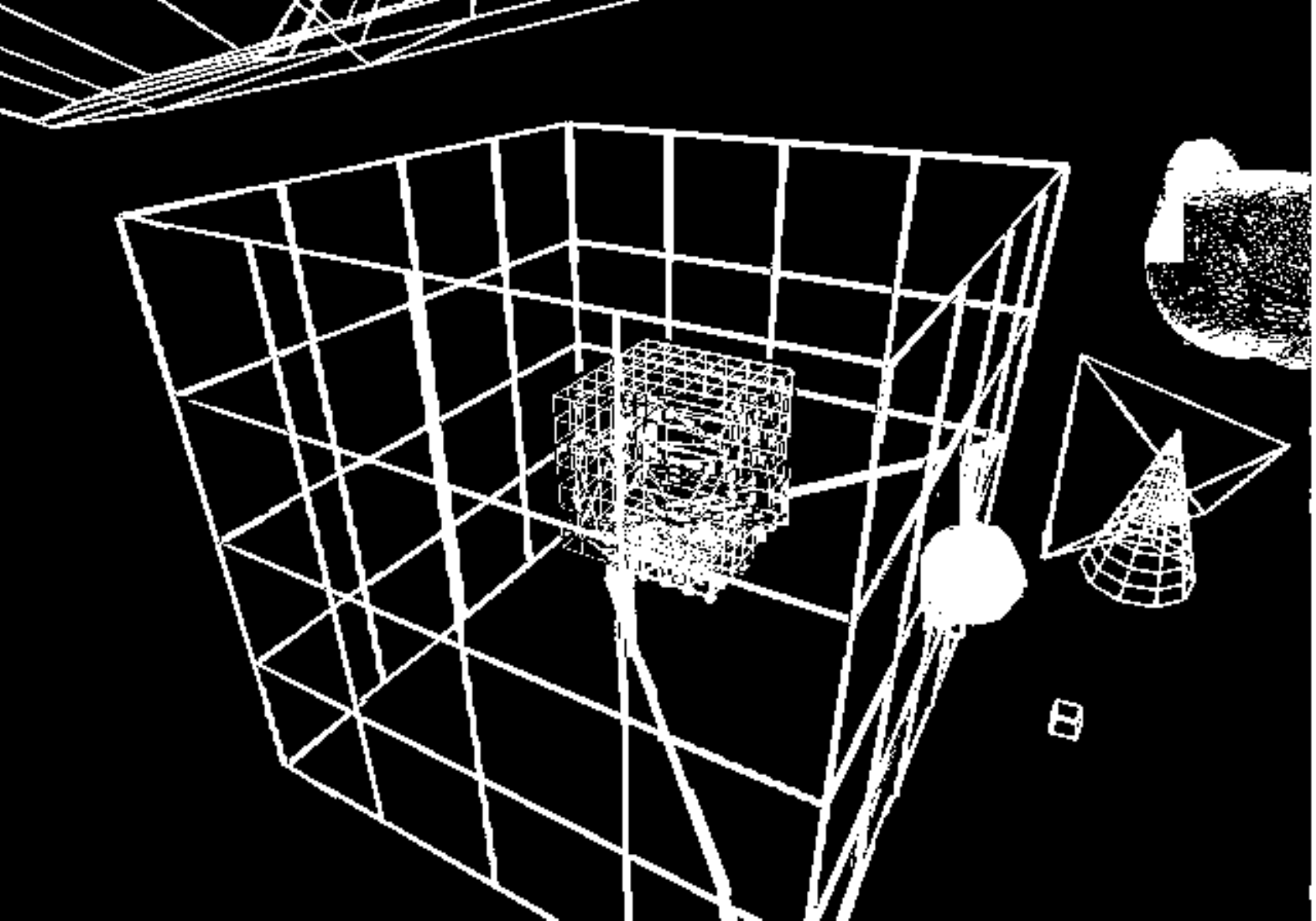}
                \vspace{-0.5cm}
            \hspace{-3cm} 
        \end{subfigure}%
        ~ 
        \begin{subfigure}[b]{0.18\textwidth}
                \includegraphics[width=\textwidth]{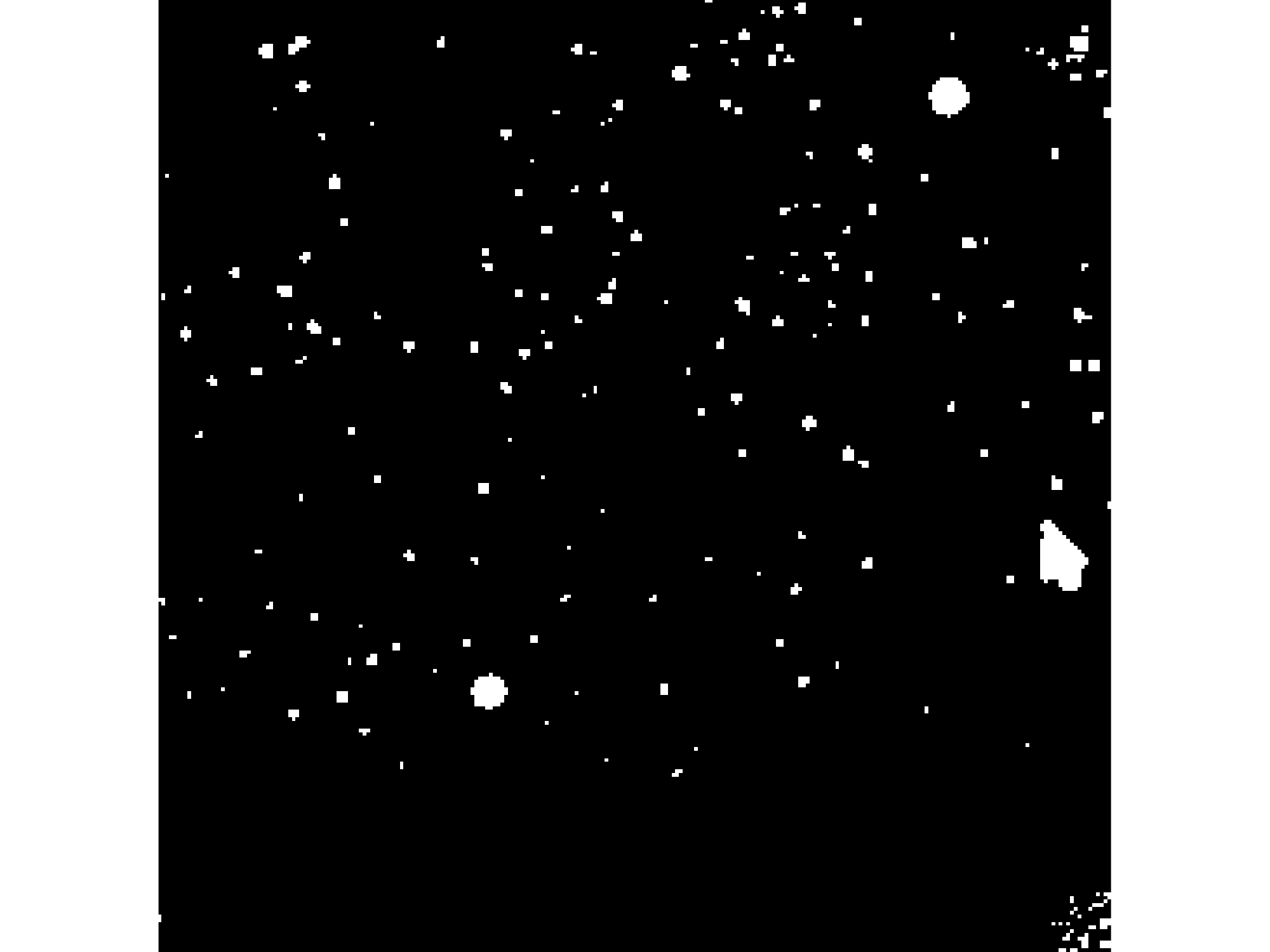}
                \vspace{-0.45cm}
            \hspace{-3cm} 
        \end{subfigure}%
        \begin{subfigure}[b]{0.18\textwidth}
			~ 
                \includegraphics[width=\textwidth]{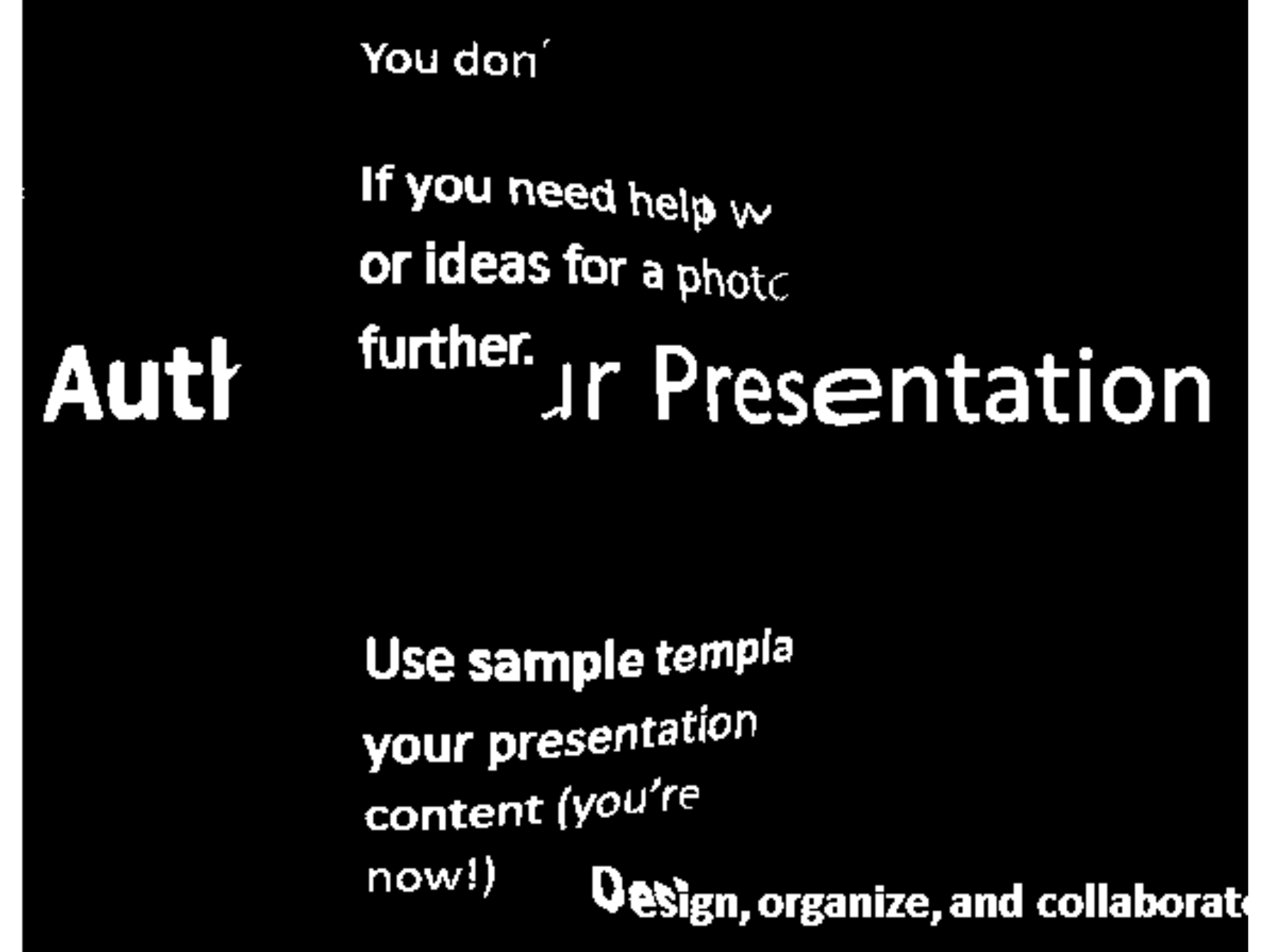}
                \vspace{-0.45cm}
            \hspace{-3cm} 
        \end{subfigure}%
        \begin{subfigure}[b]{0.18\textwidth}
                \includegraphics[width=\textwidth]{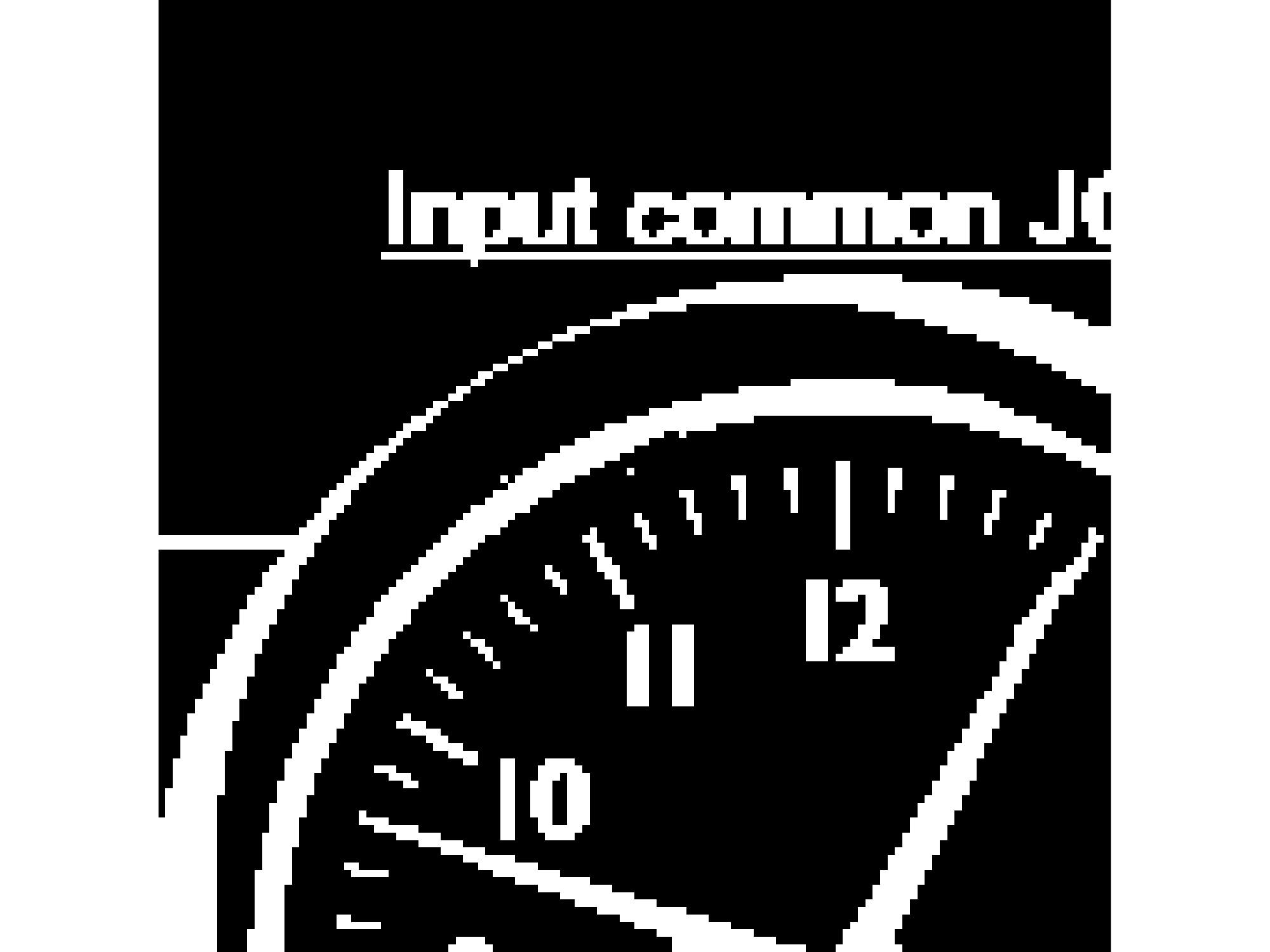}
                 \vspace{-0.45cm}
              \hspace{-4.8cm}
        \end{subfigure} \\[1ex]
        \begin{subfigure}[b]{0.18\textwidth}
                \includegraphics[width=\textwidth]{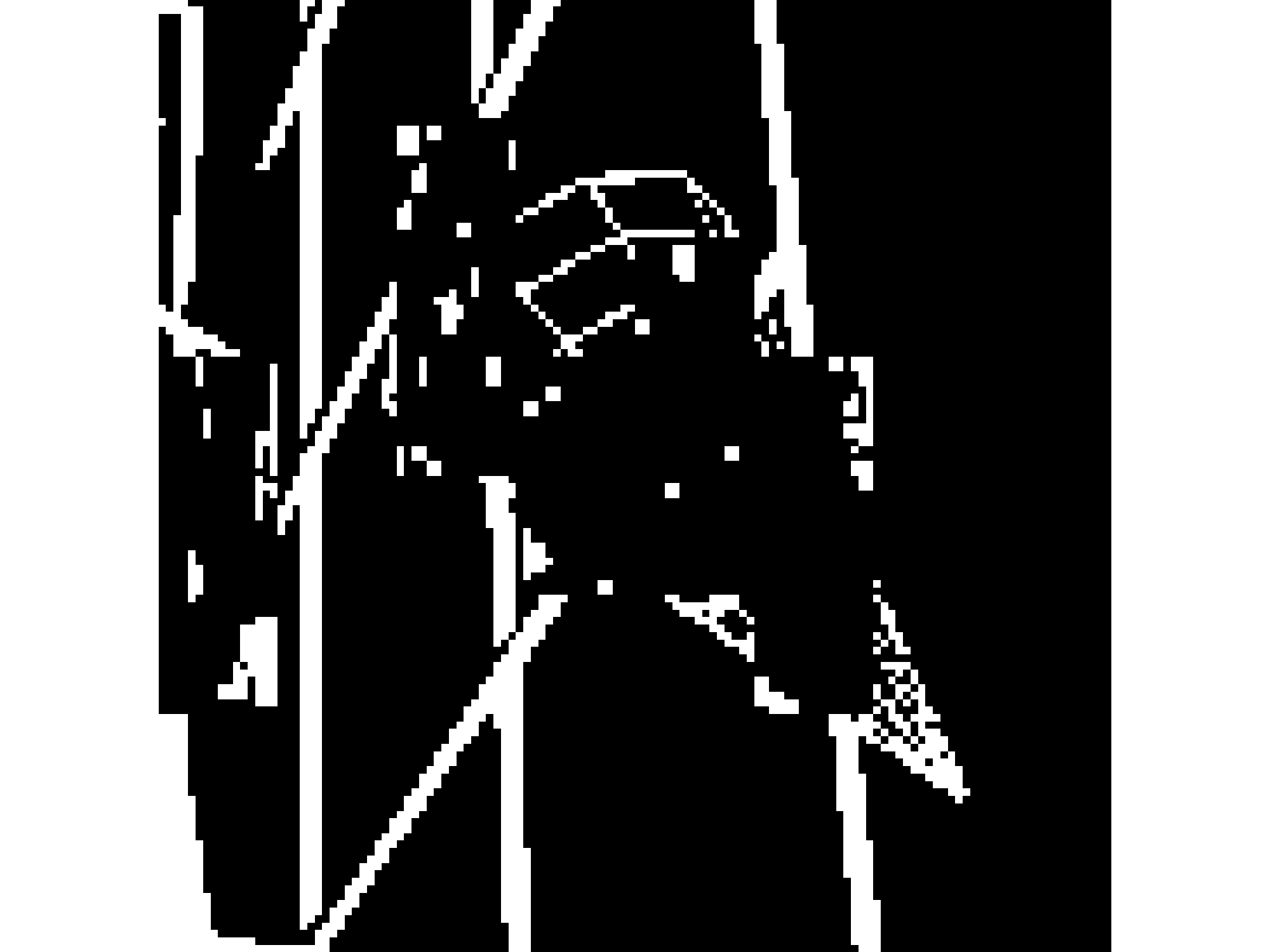}
                                \vspace{-0.5cm}
          \hspace{-2.5cm}    
        \end{subfigure}%
        ~ 
        \begin{subfigure}[b]{0.18\textwidth}
                \includegraphics[width=\textwidth]{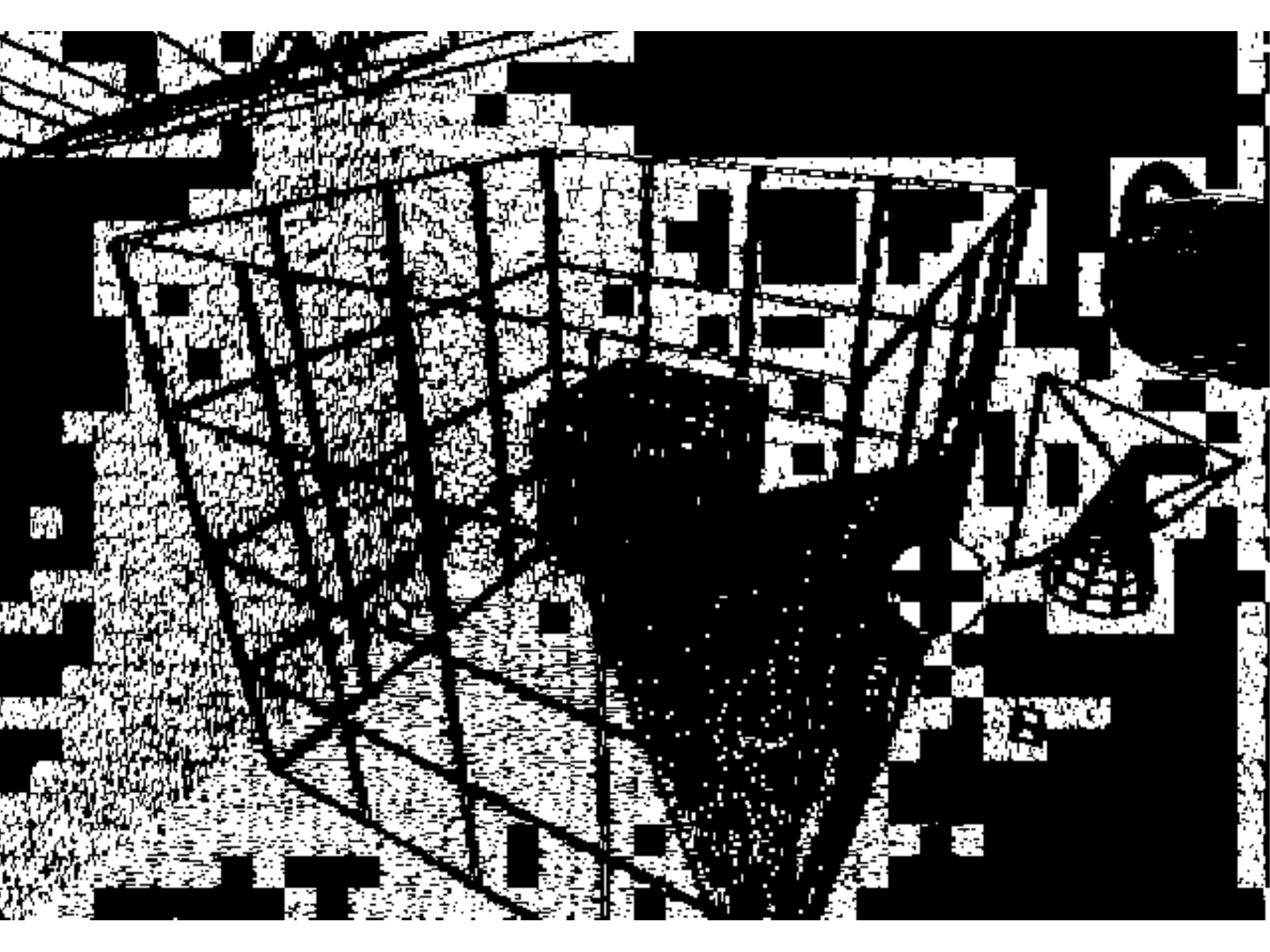}
                \vspace{-0.5cm}
            \hspace{-3cm} 
        \end{subfigure}%
        ~ 
        \begin{subfigure}[b]{0.18\textwidth}
                \includegraphics[width=\textwidth]{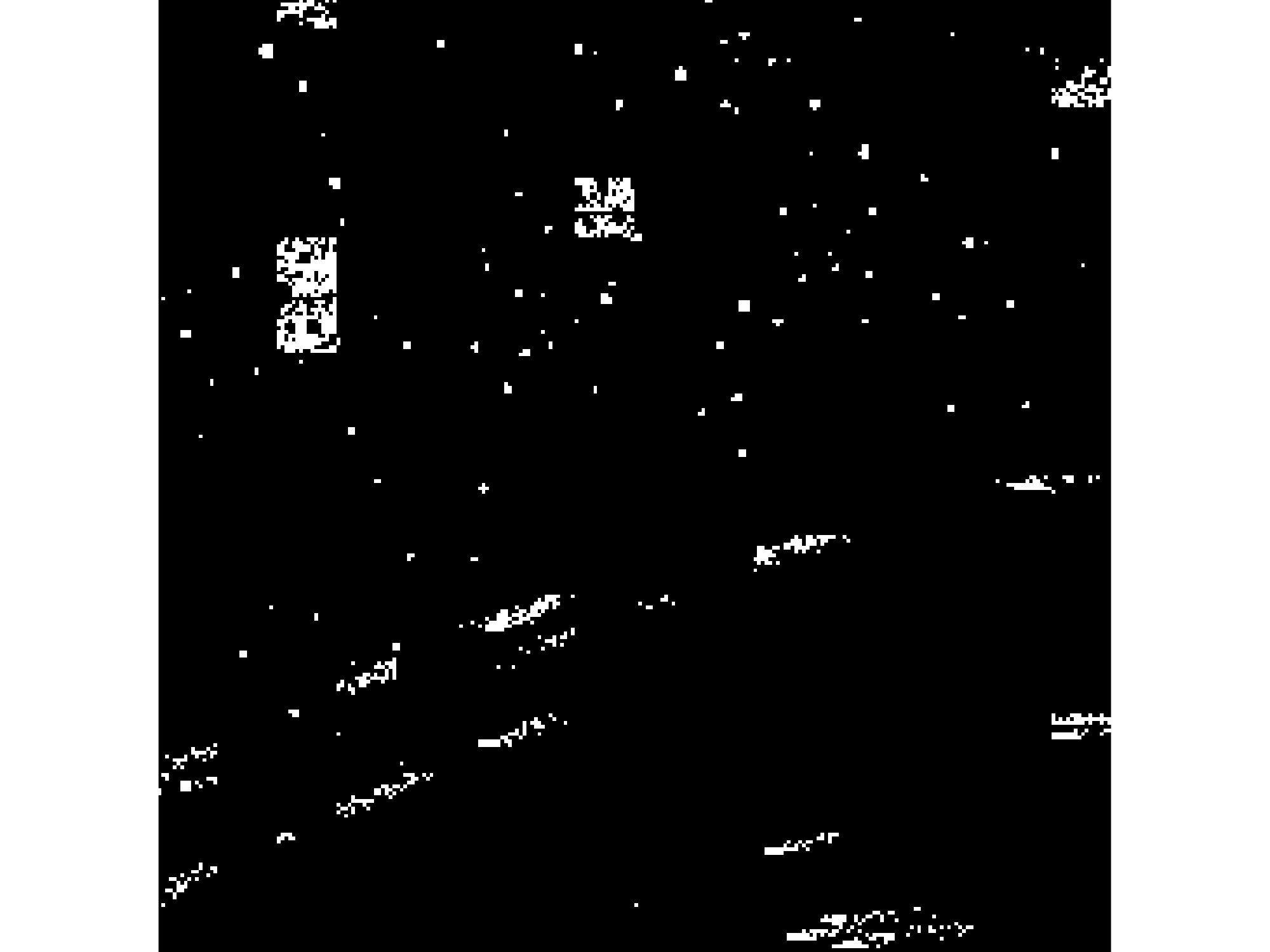}
                \vspace{-0.45cm}
            \hspace{-3cm} 
        \end{subfigure}%
        \begin{subfigure}[b]{0.18\textwidth}
			~ 
                \includegraphics[width=\textwidth]{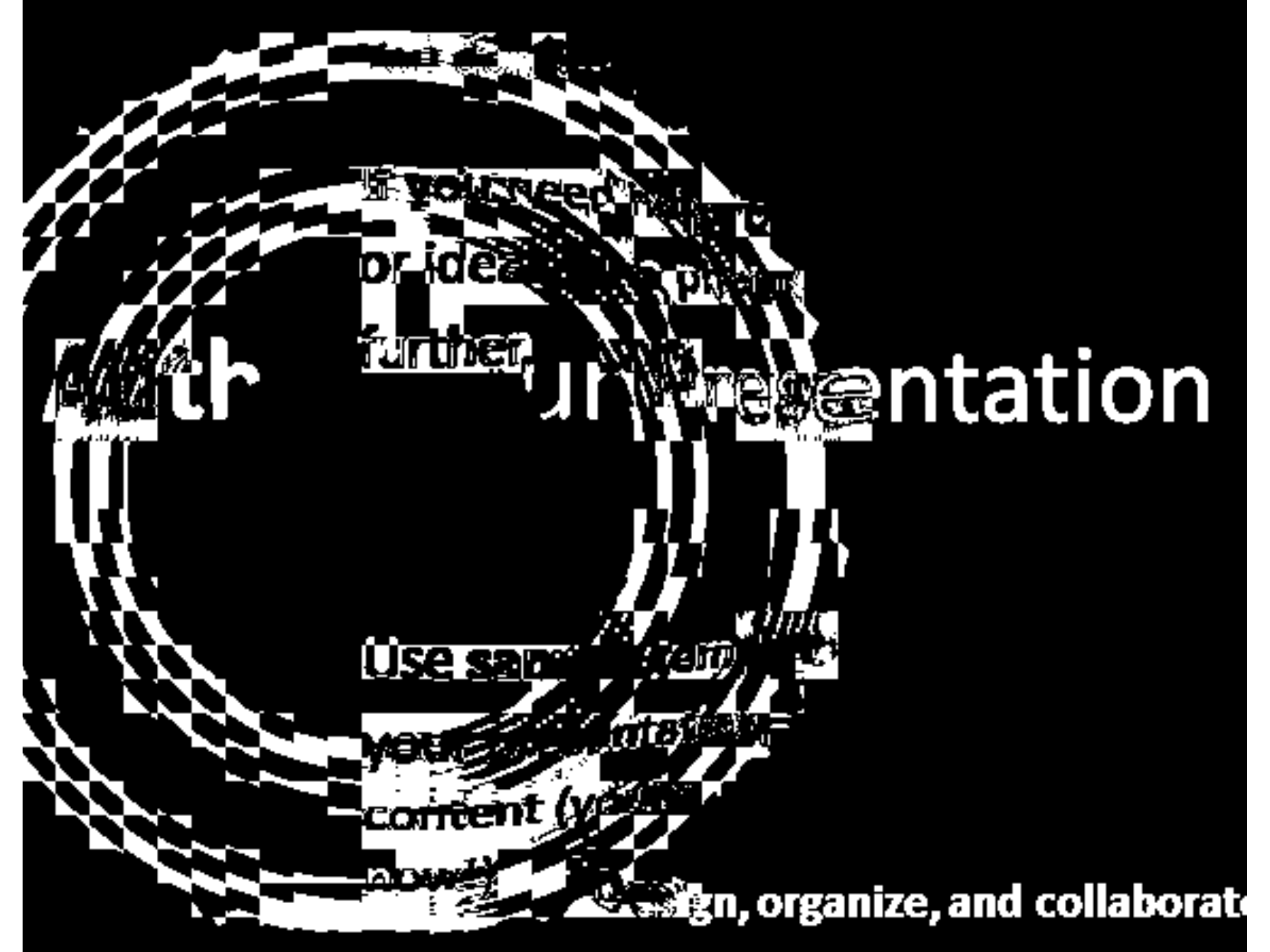}
                \vspace{-0.45cm}
            \hspace{-3cm} 
        \end{subfigure}%
        \begin{subfigure}[b]{0.18\textwidth}
                \includegraphics[width=\textwidth]{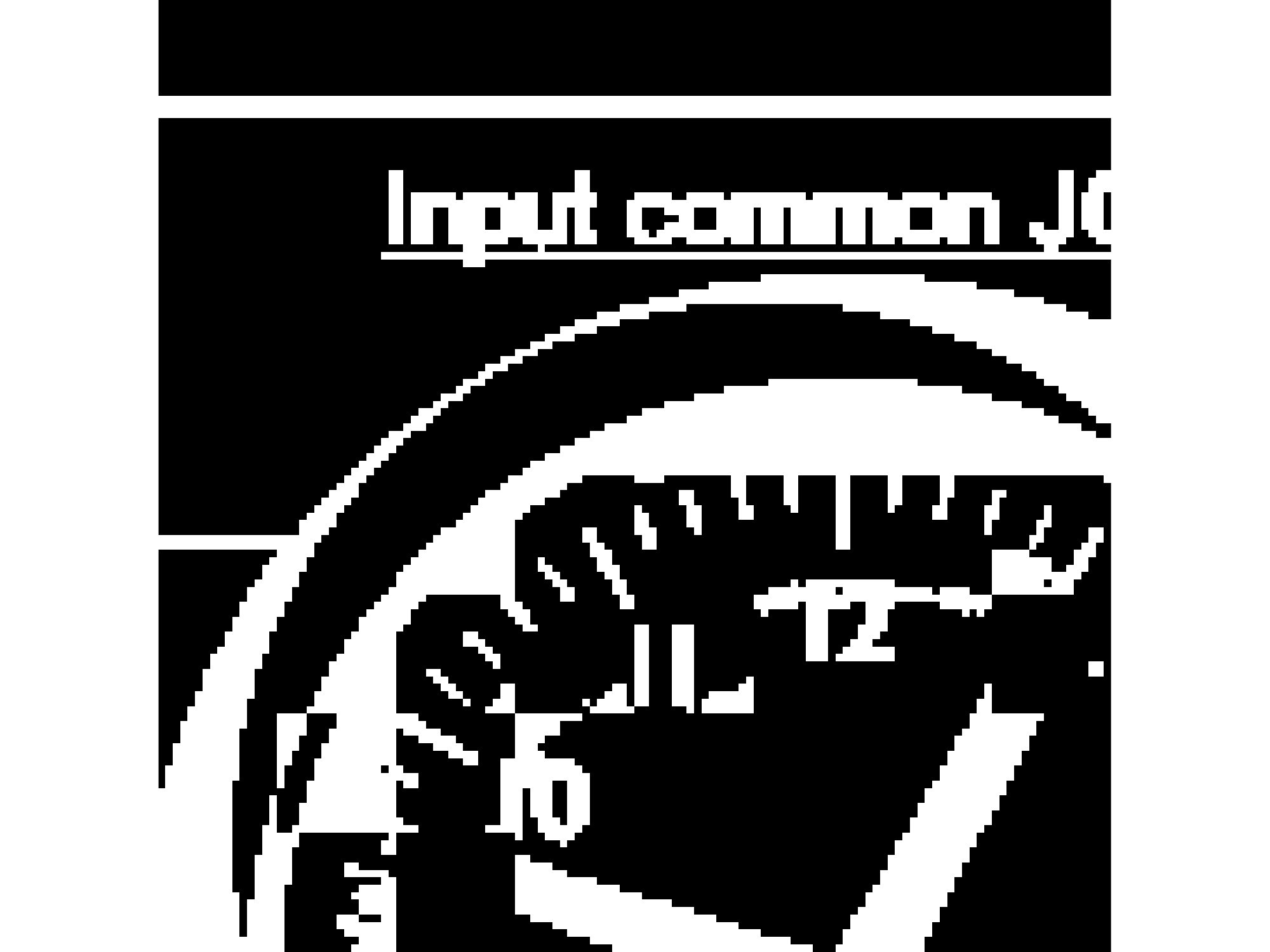}
                 \vspace{-0.45cm}
              \hspace{-4.8cm}
        \end{subfigure} \\[1ex]
        \begin{subfigure}[b]{0.18\textwidth}
                \includegraphics[width=\textwidth]{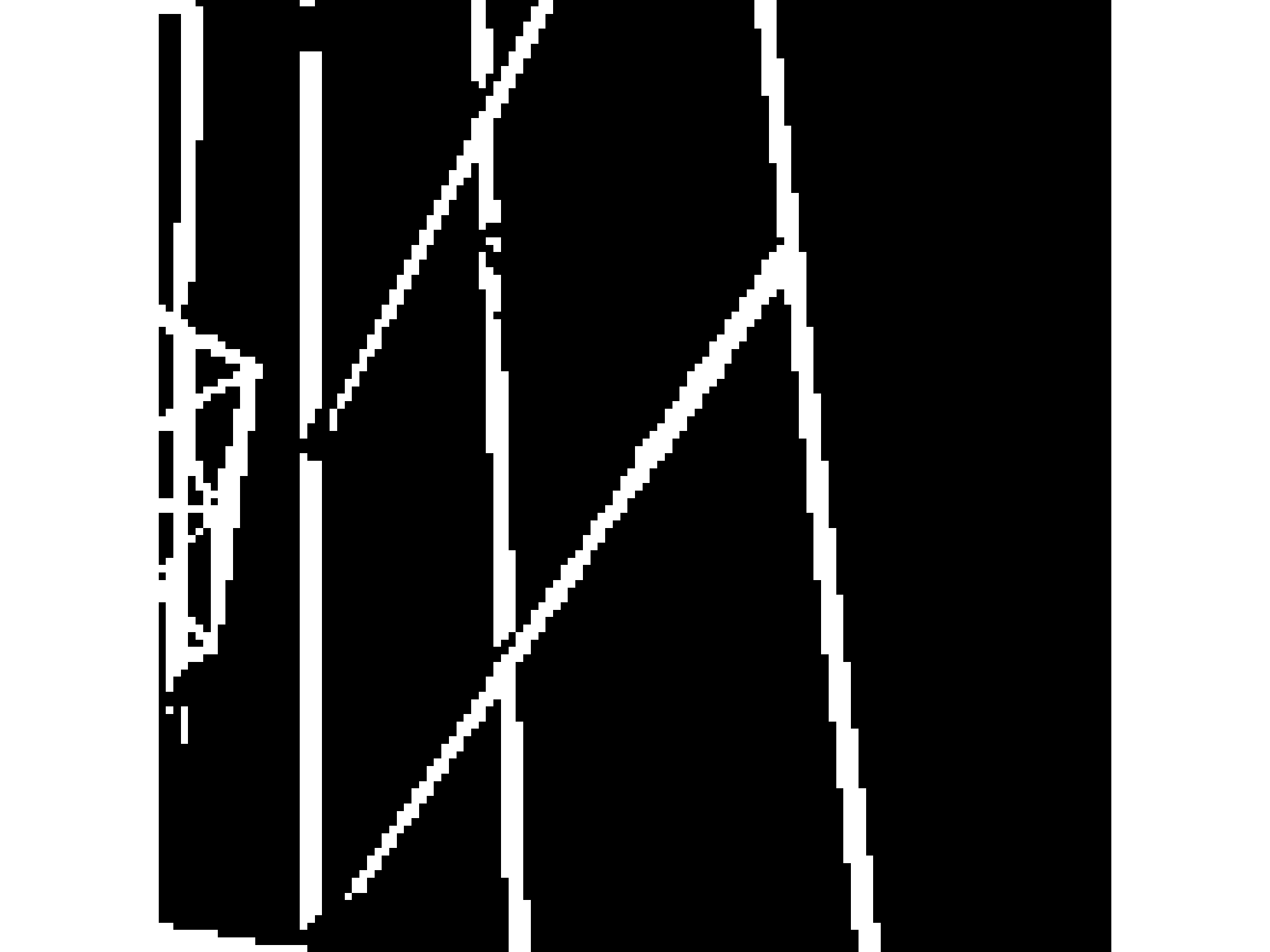}
                                \vspace{-0.5cm}
          \hspace{-2.5cm}    
        \end{subfigure}%
        ~ 
        \begin{subfigure}[b]{0.18\textwidth}
                \includegraphics[width=\textwidth]{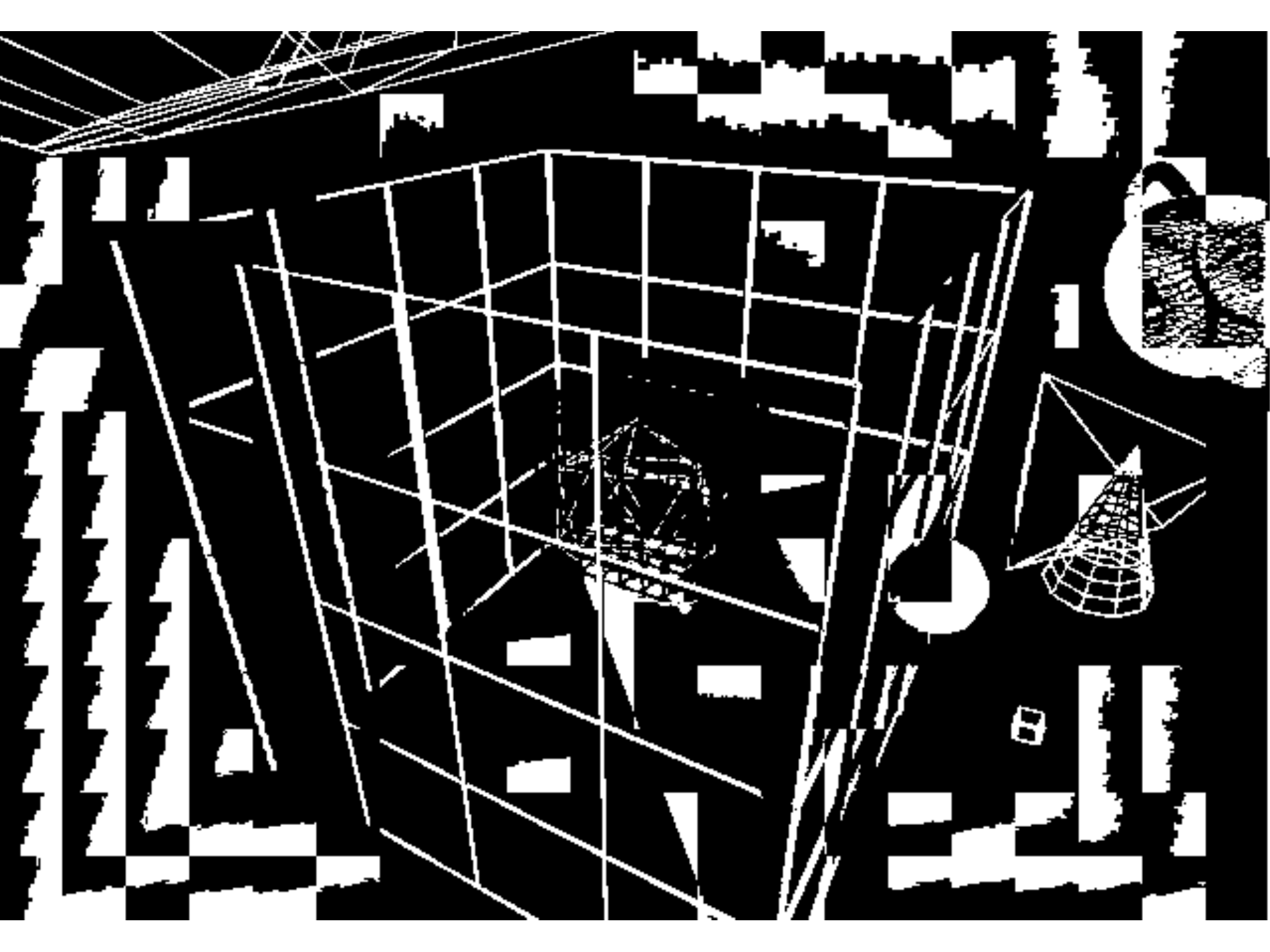}
                \vspace{-0.5cm}
            \hspace{-3cm} 
        \end{subfigure}%
        ~ 
        \begin{subfigure}[b]{0.18\textwidth}
                \includegraphics[width=\textwidth]{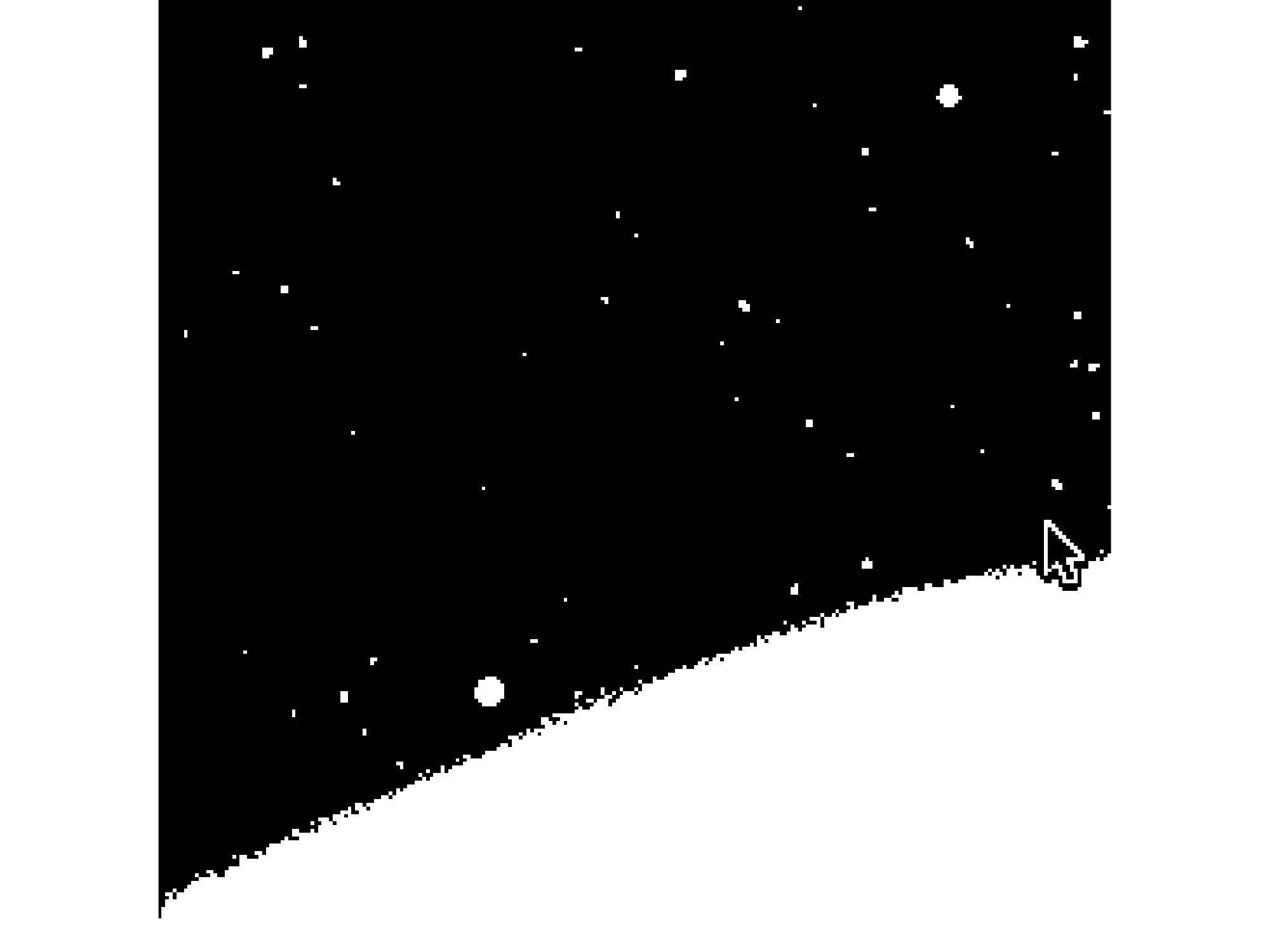}
                \vspace{-0.45cm}
            \hspace{-3cm} 
        \end{subfigure}%
        \begin{subfigure}[b]{0.18\textwidth}
			~ 
                \includegraphics[width=\textwidth]{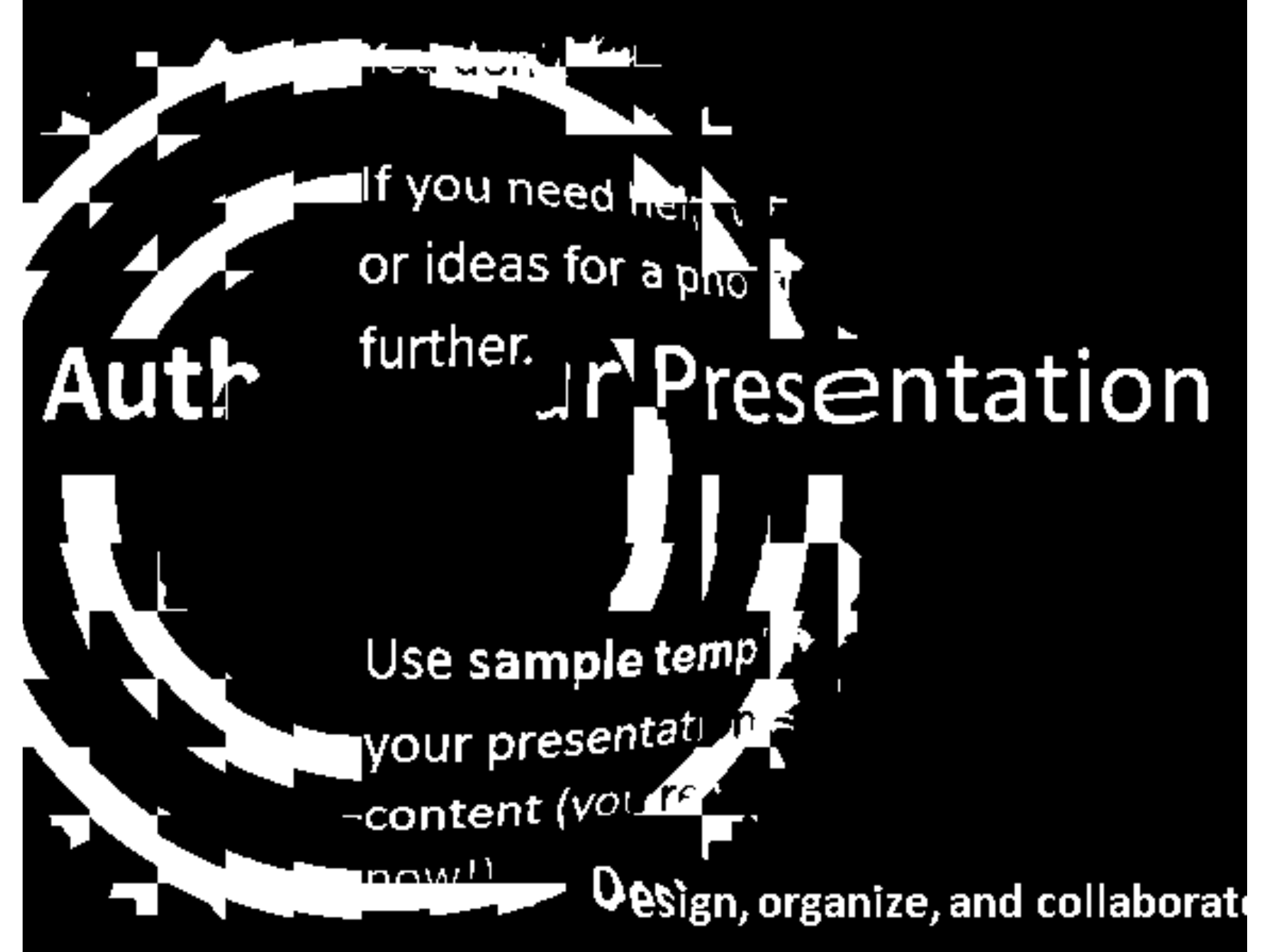}
                \vspace{-0.45cm}
            \hspace{-3cm} 
        \end{subfigure}%
        \begin{subfigure}[b]{0.18\textwidth}
                \includegraphics[width=\textwidth]{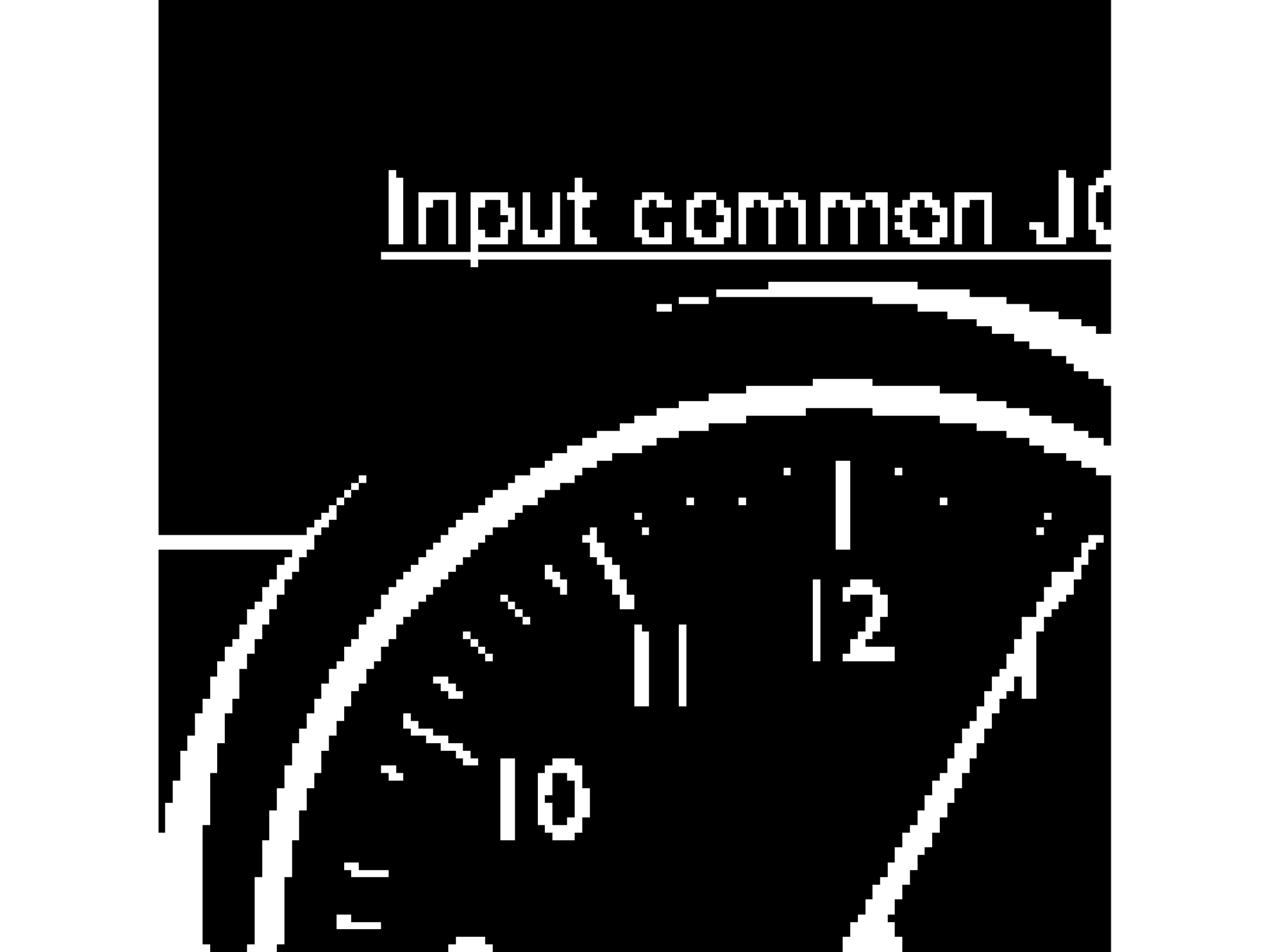}
                 \vspace{-0.45cm}
              \hspace{-4.8cm}
        \end{subfigure} \\[1ex]
        \begin{subfigure}[b]{0.18\textwidth}
                \includegraphics[width=\textwidth]{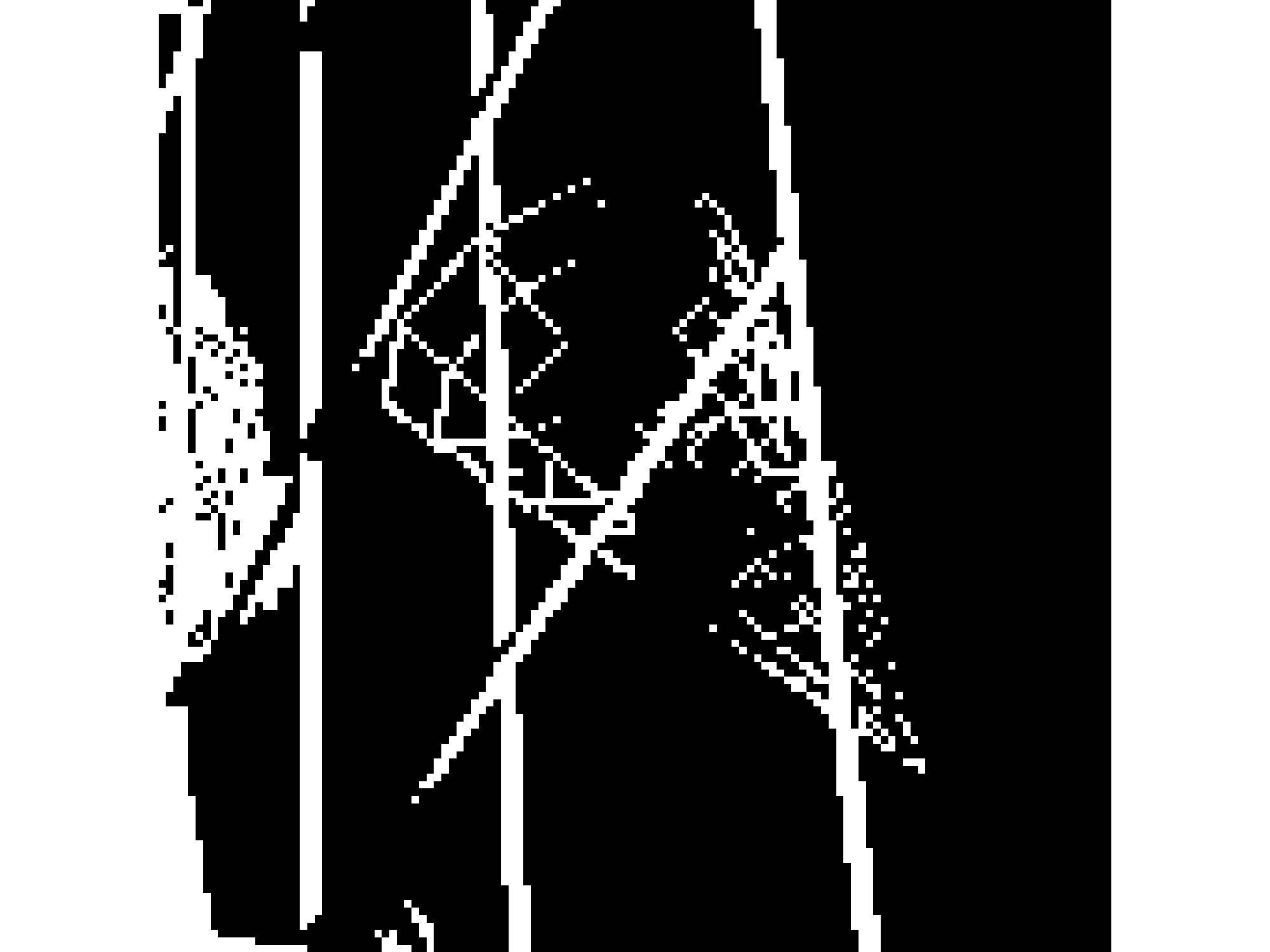}
                                \vspace{-0.5cm}
          \hspace{-2.5cm}    
        \end{subfigure}%
        ~ 
        \begin{subfigure}[b]{0.18\textwidth}
                \includegraphics[width=\textwidth]{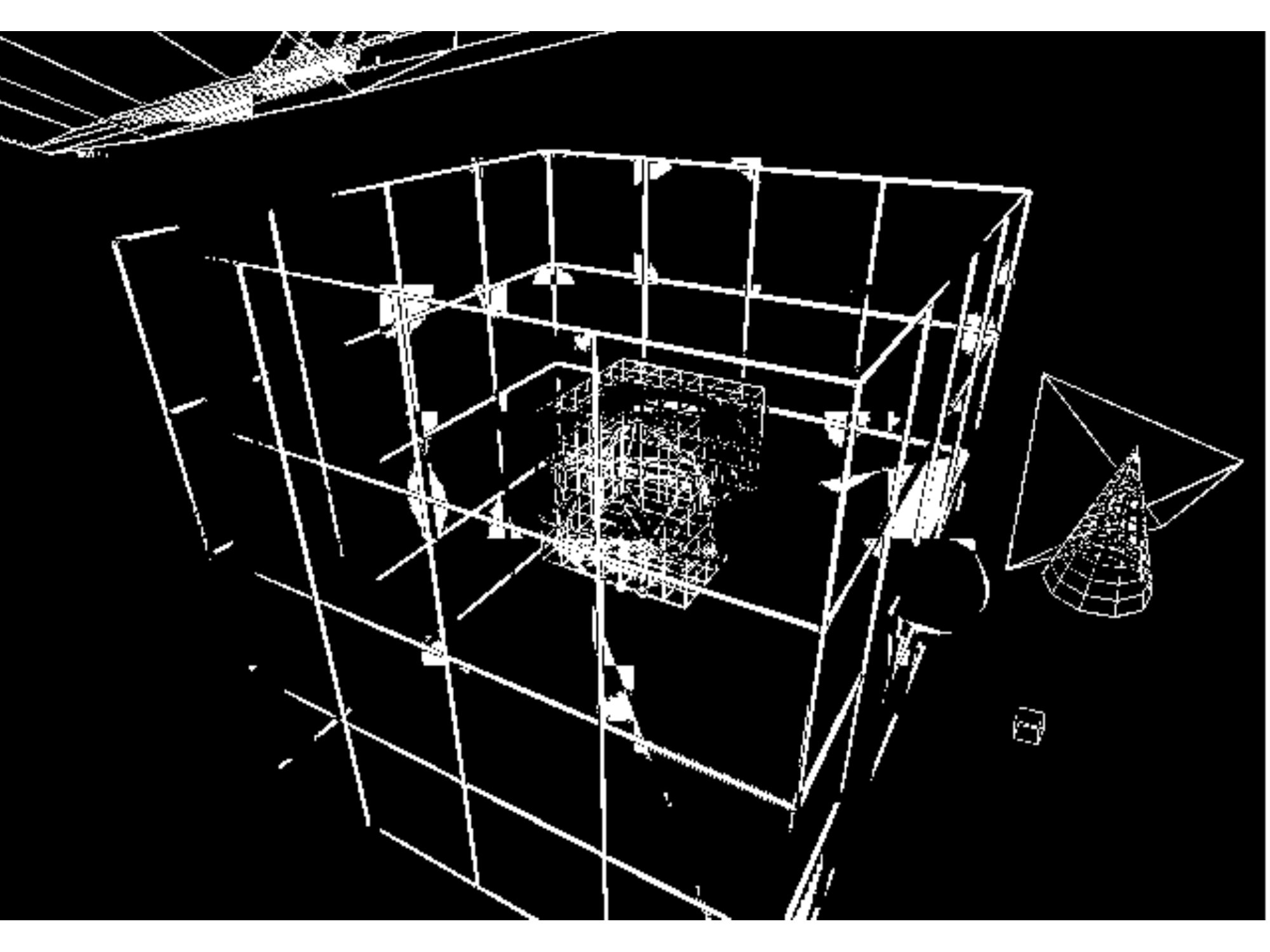}
                \vspace{-0.5cm}
            \hspace{-3cm} 
        \end{subfigure}%
        ~ 
        \begin{subfigure}[b]{0.18\textwidth}
                \includegraphics[width=\textwidth]{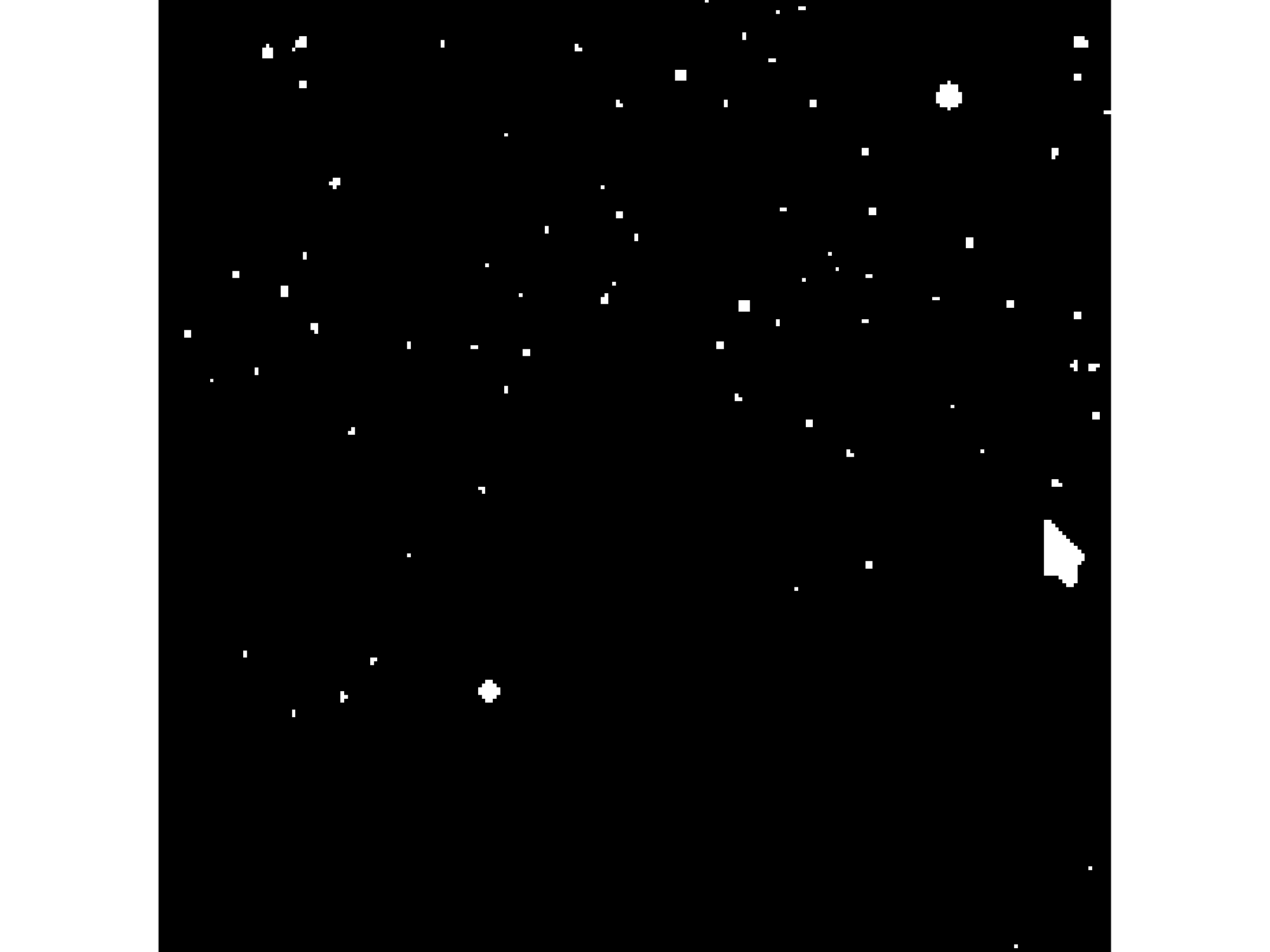}
                \vspace{-0.45cm}
            \hspace{-3cm} 
        \end{subfigure}%
        \begin{subfigure}[b]{0.18\textwidth}
			~ 
                \includegraphics[width=\textwidth]{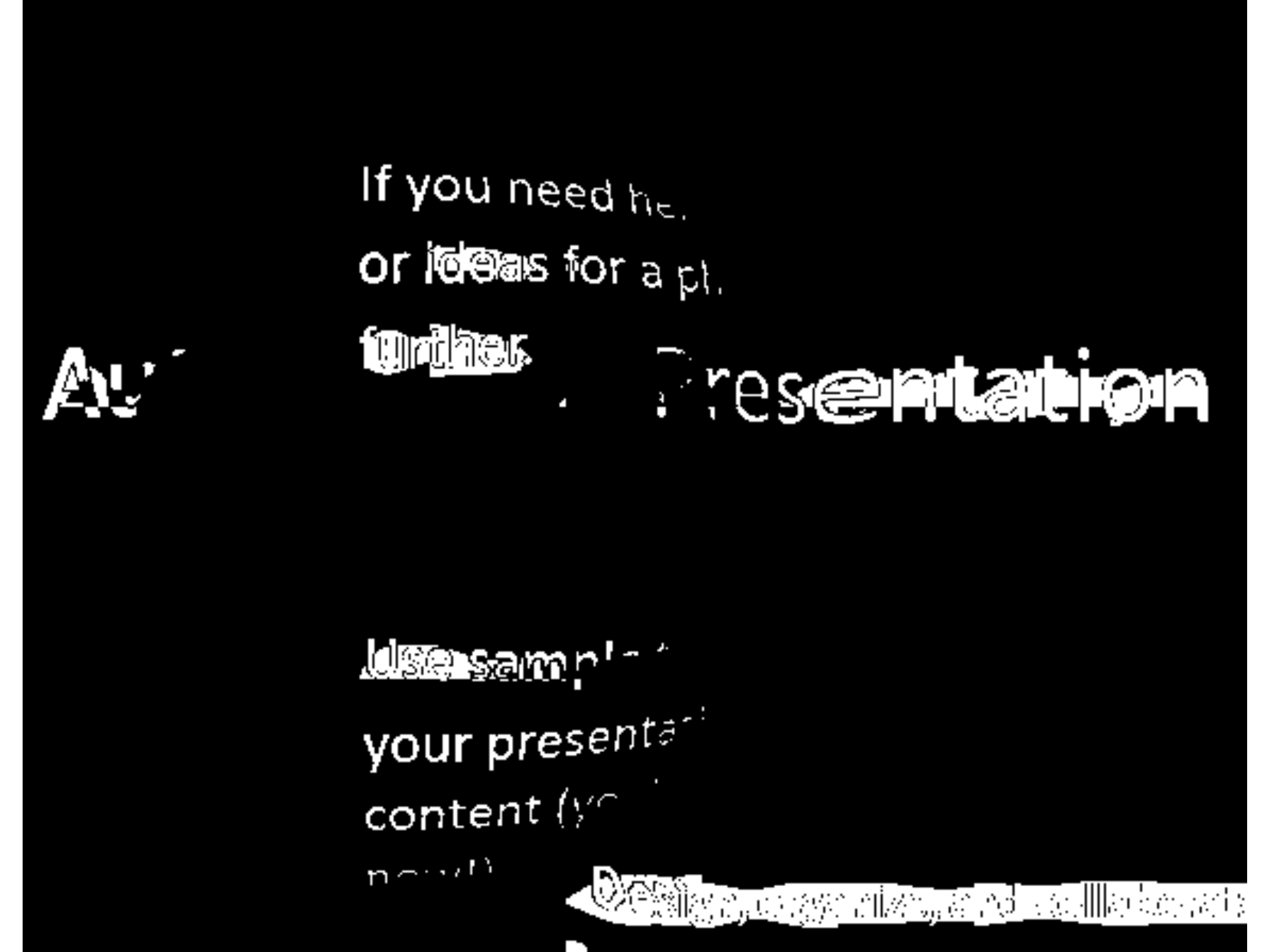}
                \vspace{-0.45cm}
            \hspace{-3cm} 
        \end{subfigure}%
        \begin{subfigure}[b]{0.18\textwidth}
                \includegraphics[width=\textwidth]{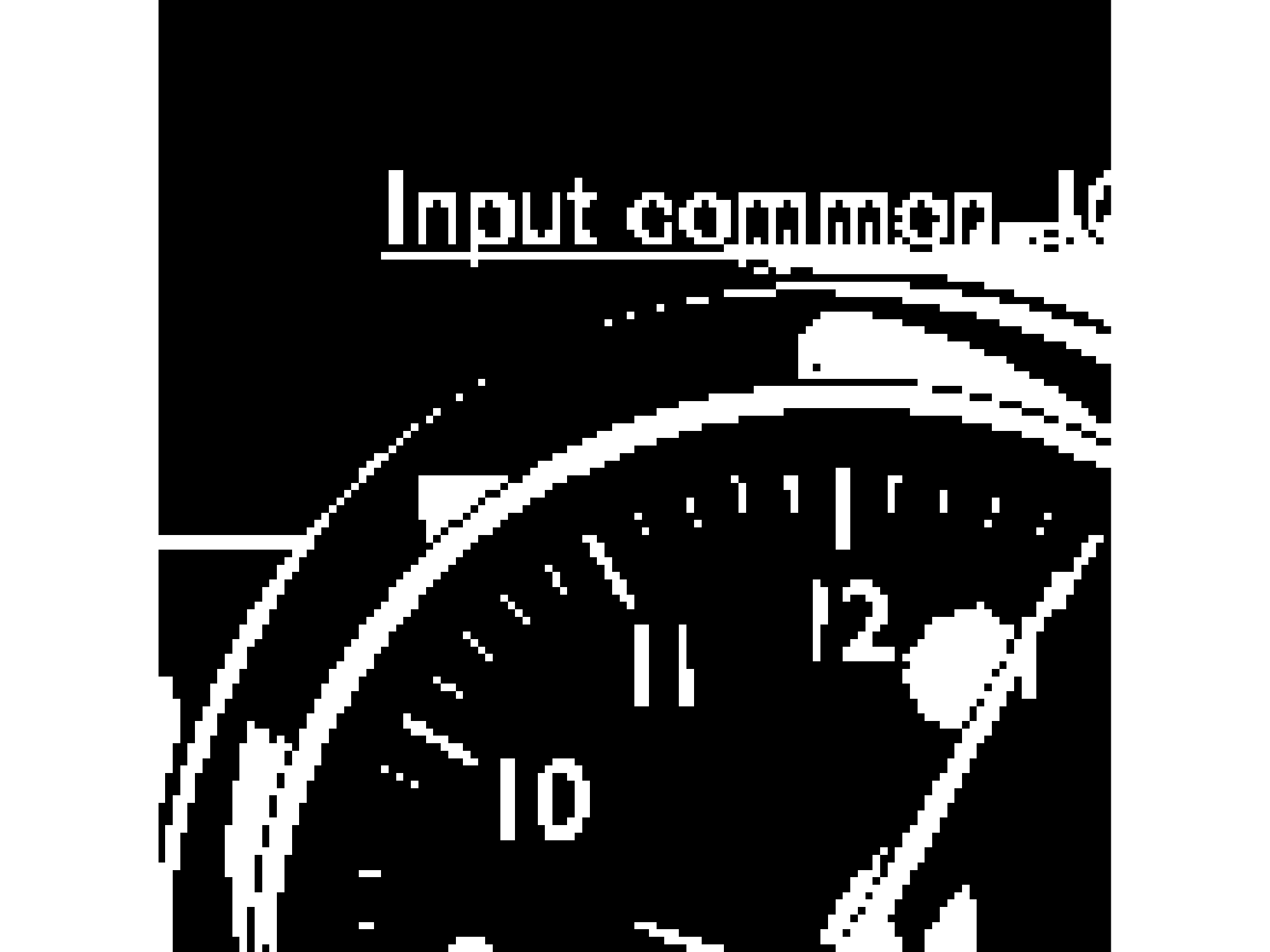}
                 \vspace{-0.45cm}
              \hspace{-4.8cm}
        \end{subfigure} \\[1ex]        
        \begin{subfigure}[b]{0.18\textwidth}
                \includegraphics[width=\textwidth]{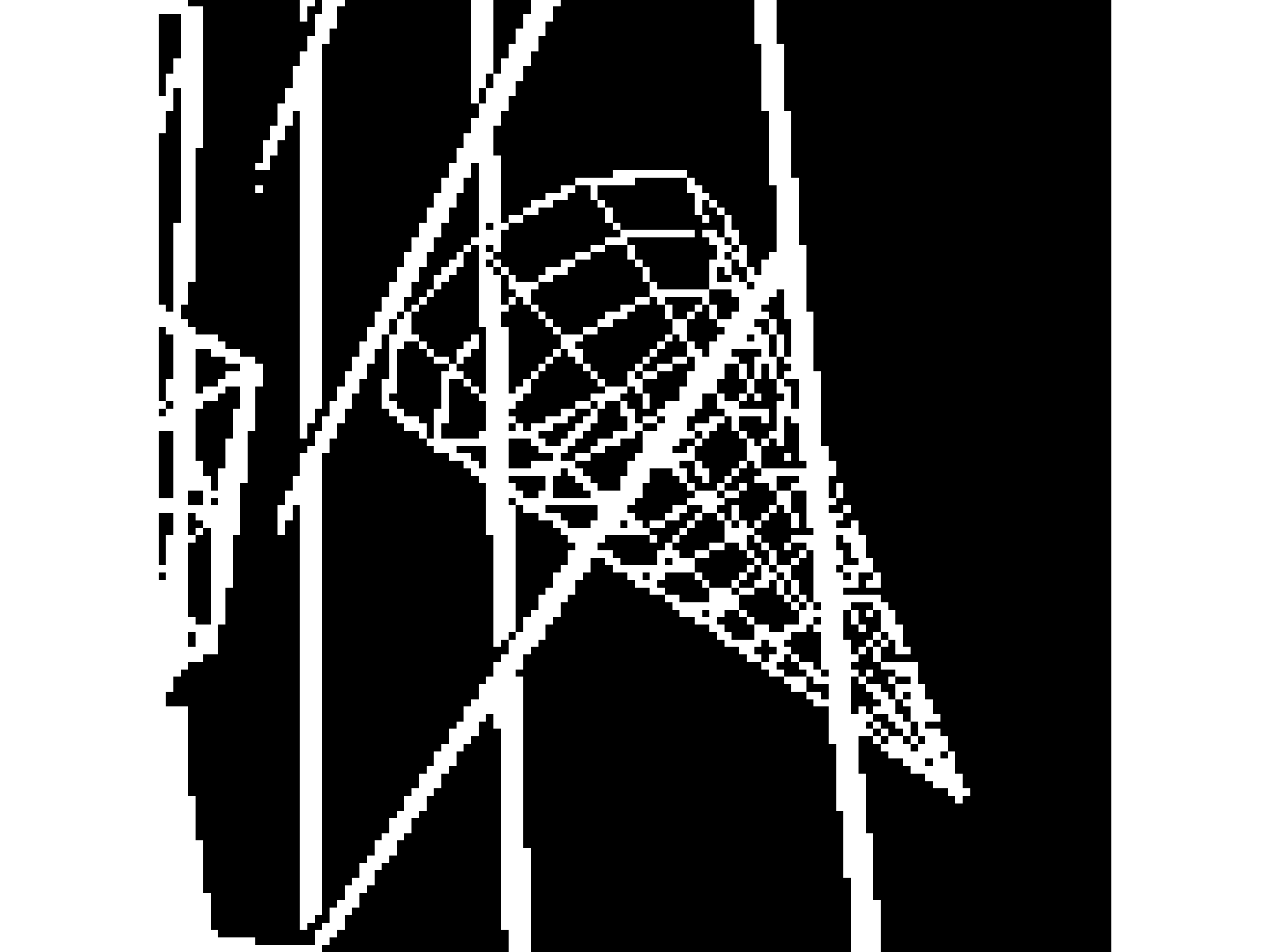}
                                \vspace{-0.5cm}
          \hspace{-2.5cm}    
        \end{subfigure}%
        ~ 
        \begin{subfigure}[b]{0.18\textwidth}
                \includegraphics[width=\textwidth]{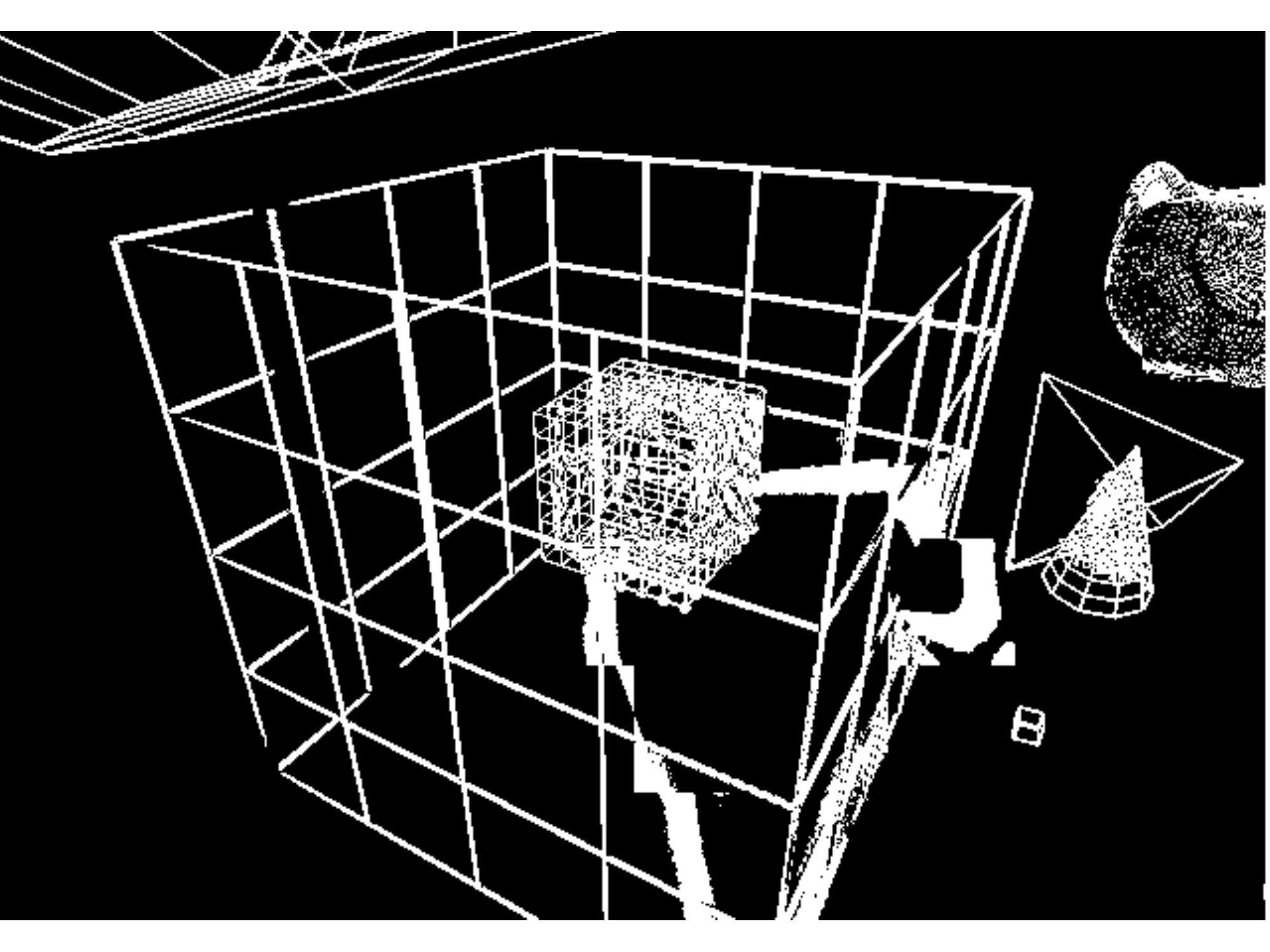}
                \vspace{-0.5cm}
            \hspace{-3cm} 
        \end{subfigure}%
        ~ 
        \begin{subfigure}[b]{0.18\textwidth}
                \includegraphics[width=\textwidth]{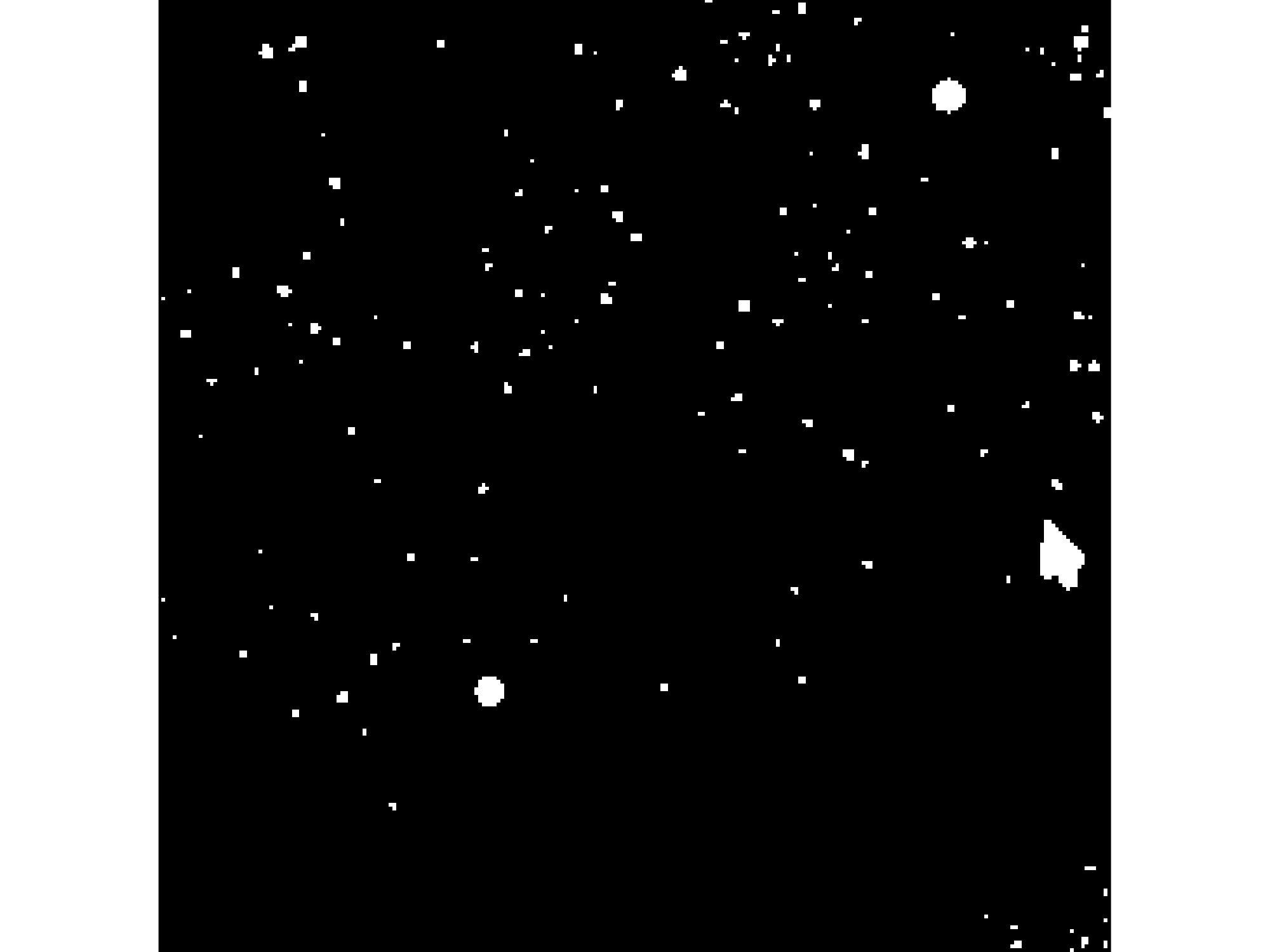}
                \vspace{-0.45cm}
            \hspace{-3cm} 
        \end{subfigure}%
        \begin{subfigure}[b]{0.18\textwidth}
			~ 
                \includegraphics[width=\textwidth]{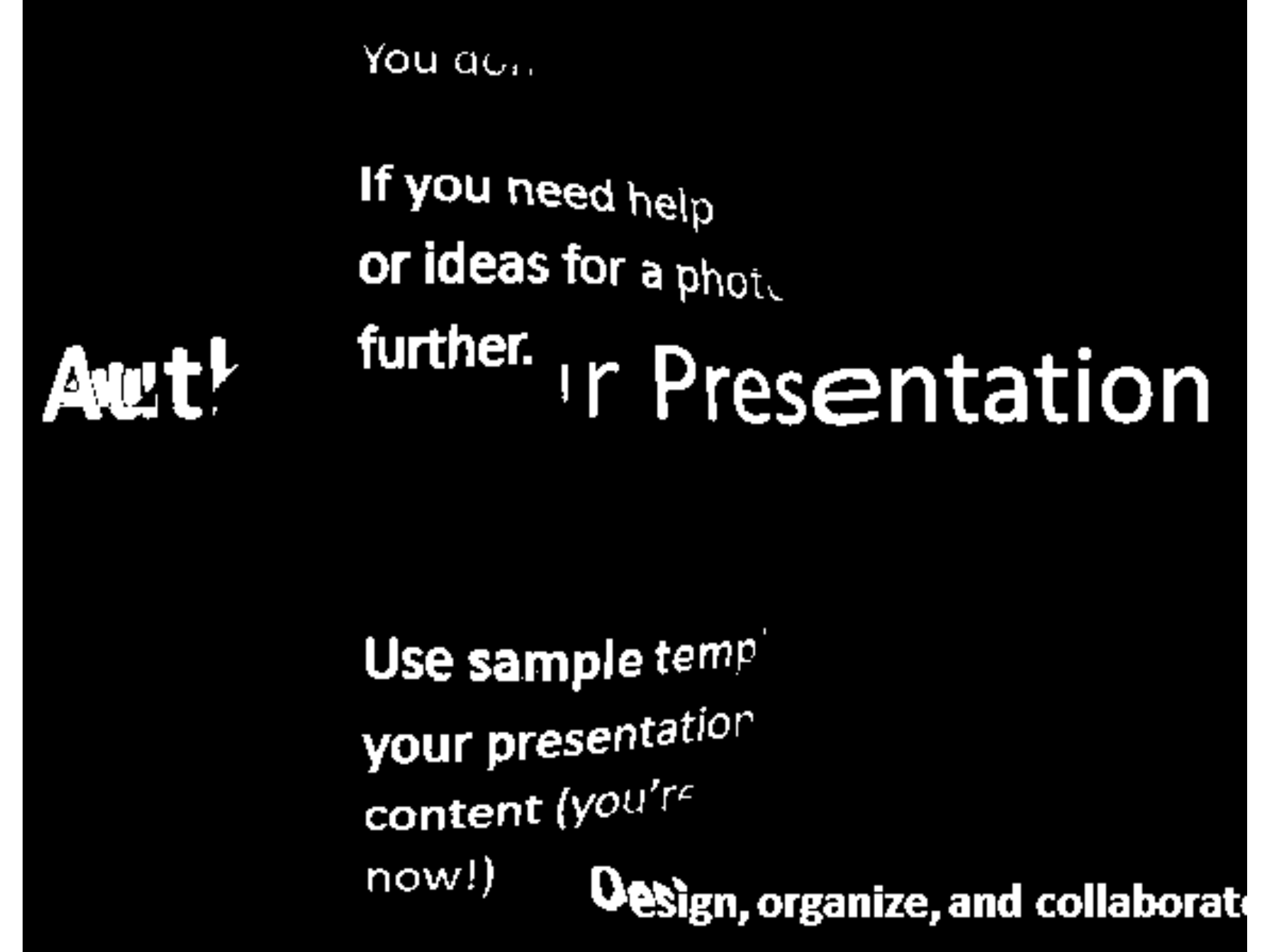}
                \vspace{-0.45cm}
            \hspace{-3cm} 
        \end{subfigure}%
        \begin{subfigure}[b]{0.18\textwidth}
                \includegraphics[width=\textwidth]{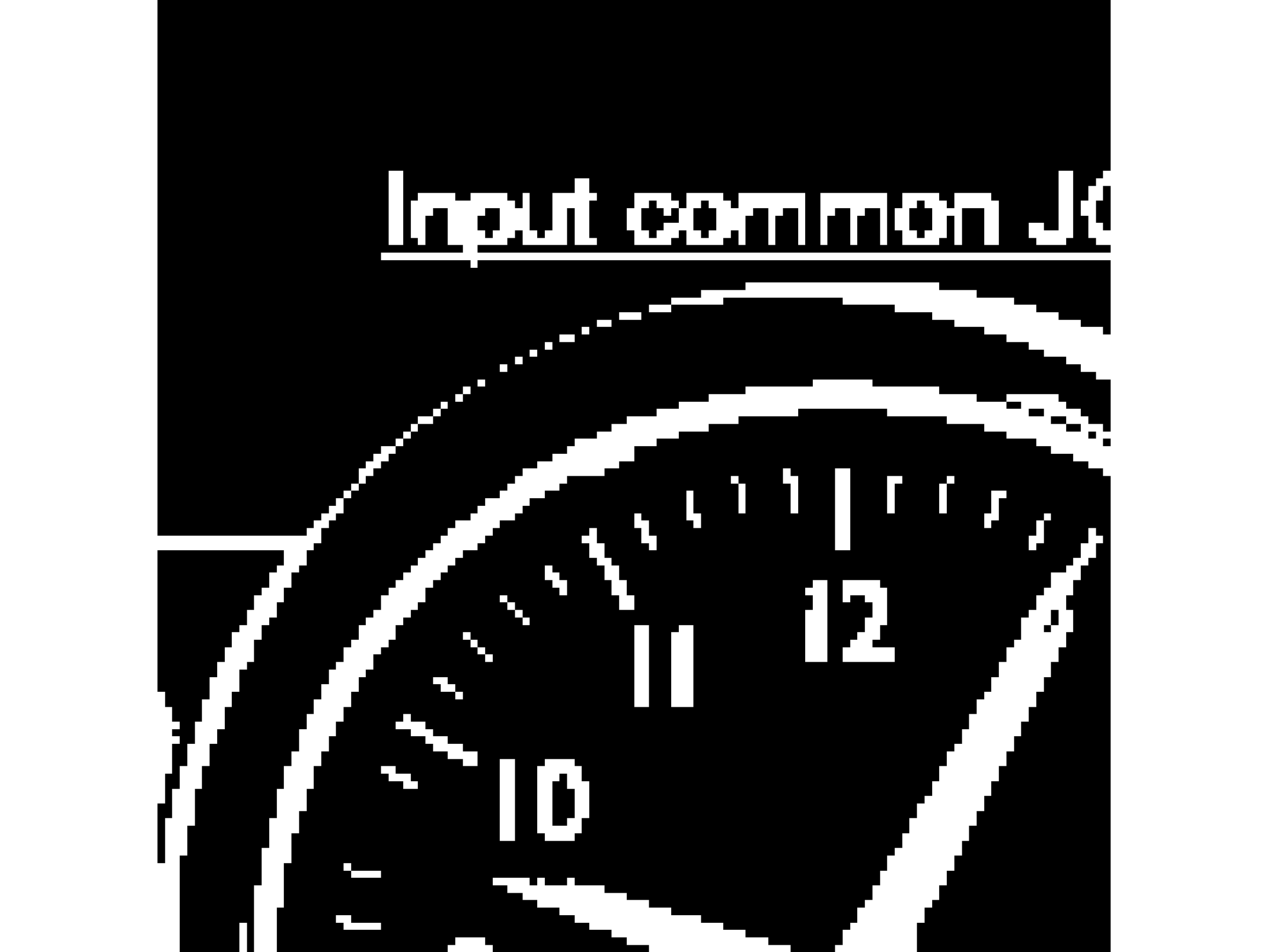}
                 \vspace{-0.45cm}
              \hspace{-4.8cm}
        \end{subfigure} \\[1ex]        
        \begin{subfigure}[b]{0.18\textwidth}
                \includegraphics[width=\textwidth]{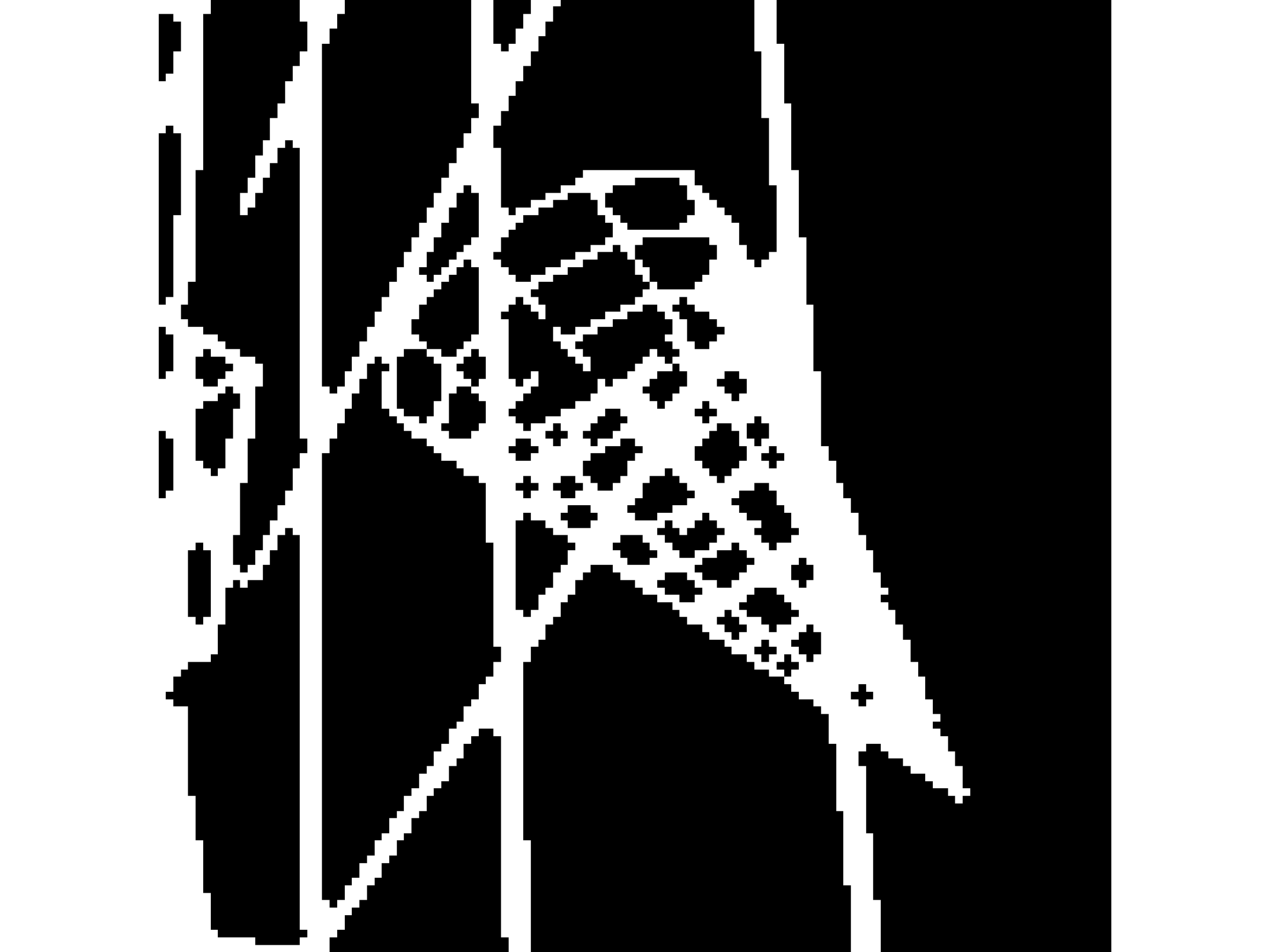}
                                \vspace{-0.5cm}
          \hspace{-2.5cm}    
        \end{subfigure}%
        ~ 
        \begin{subfigure}[b]{0.18\textwidth}
                \includegraphics[width=\textwidth]{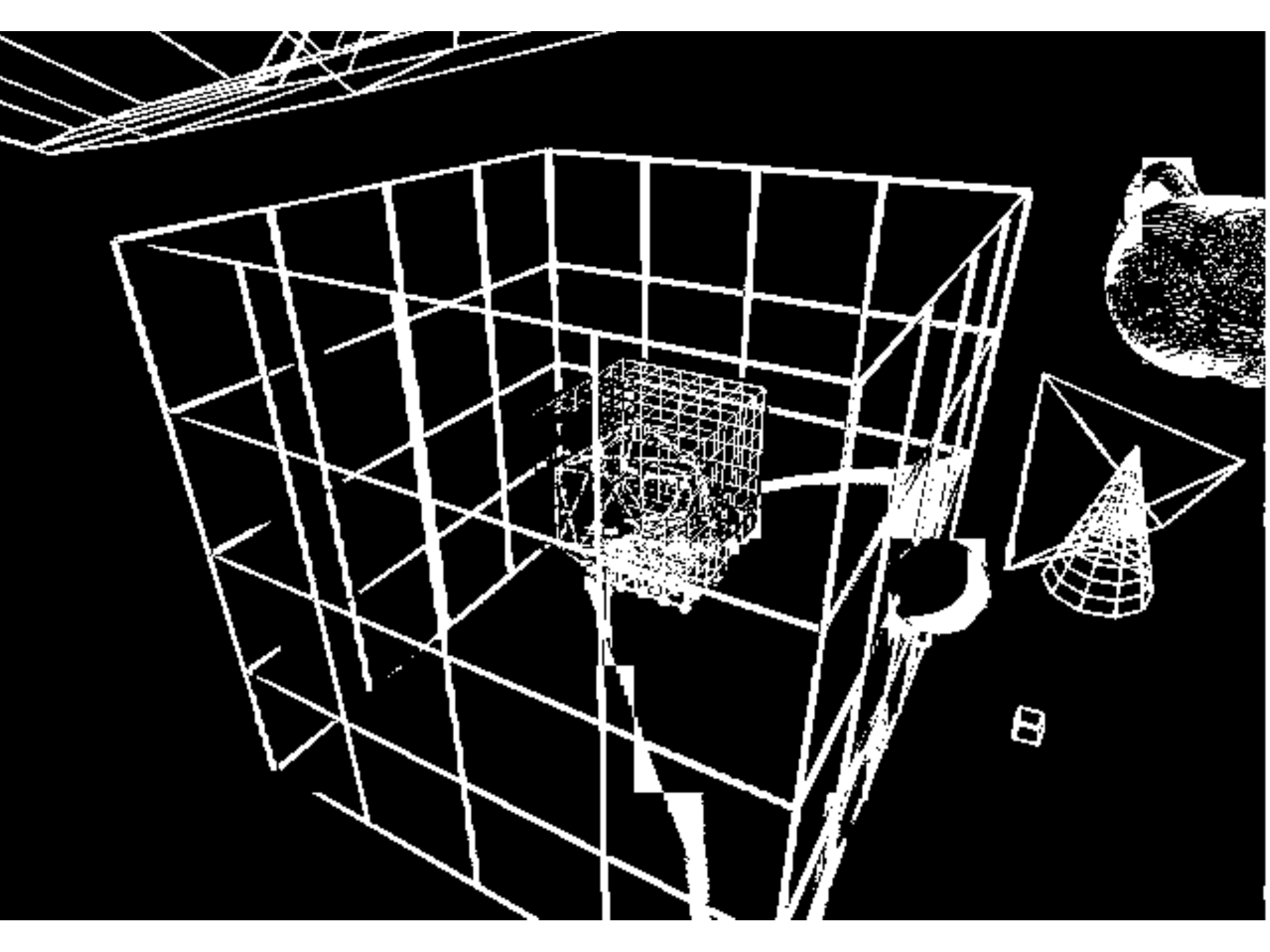}
                \vspace{-0.5cm}
            \hspace{-3cm} 
        \end{subfigure}%
        ~ 
        \begin{subfigure}[b]{0.18\textwidth}
                \includegraphics[width=\textwidth]{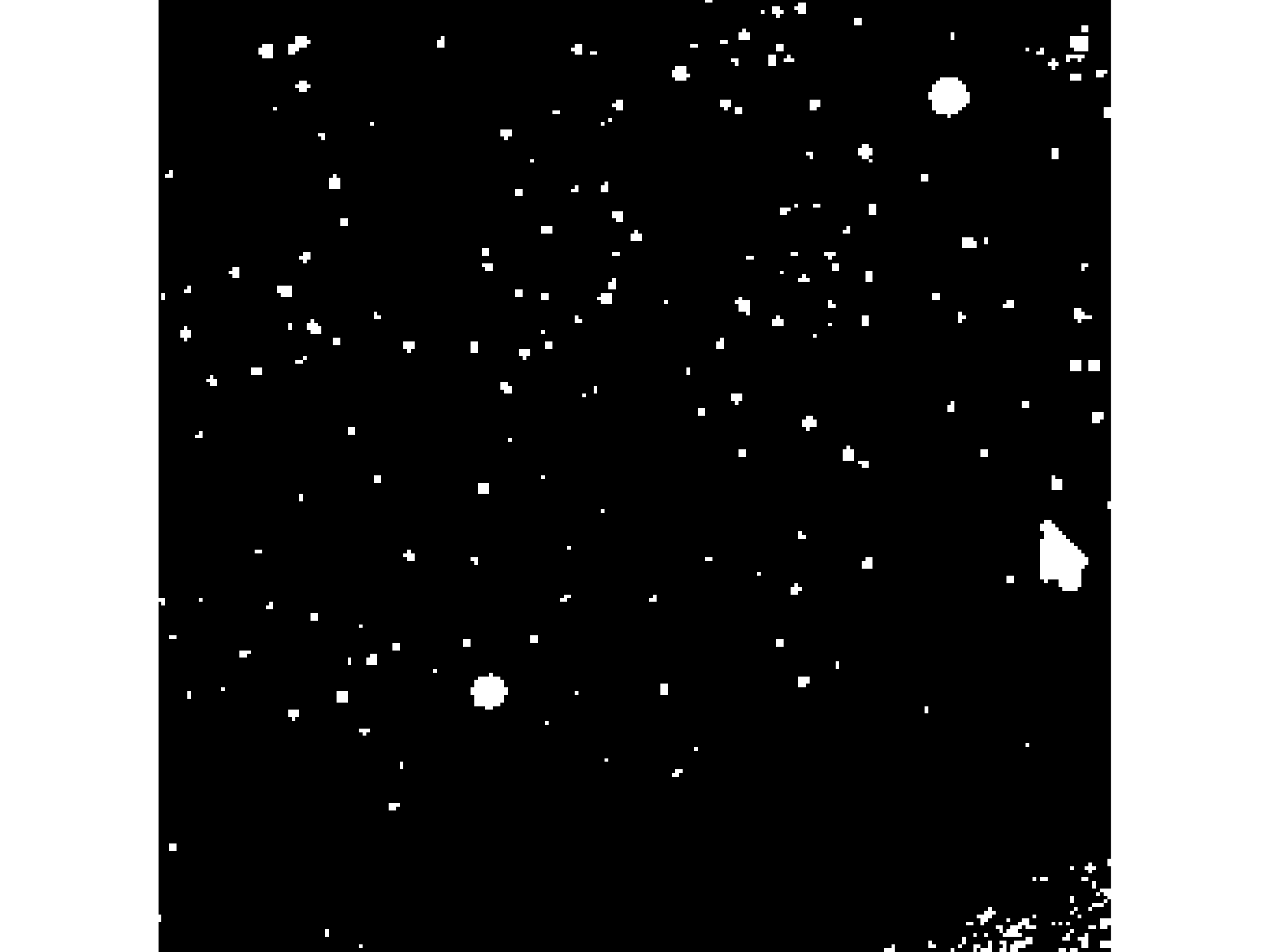}
                \vspace{-0.45cm}
            \hspace{-3cm} 
        \end{subfigure}%
        \begin{subfigure}[b]{0.18\textwidth}
			~ 
                \includegraphics[width=\textwidth]{2_map_basis10-eps-converted-to.pdf} 
                \vspace{-0.45cm}
            \hspace{-3cm} 
        \end{subfigure}%
        \begin{subfigure}[b]{0.18\textwidth}
                \includegraphics[width=\textwidth]{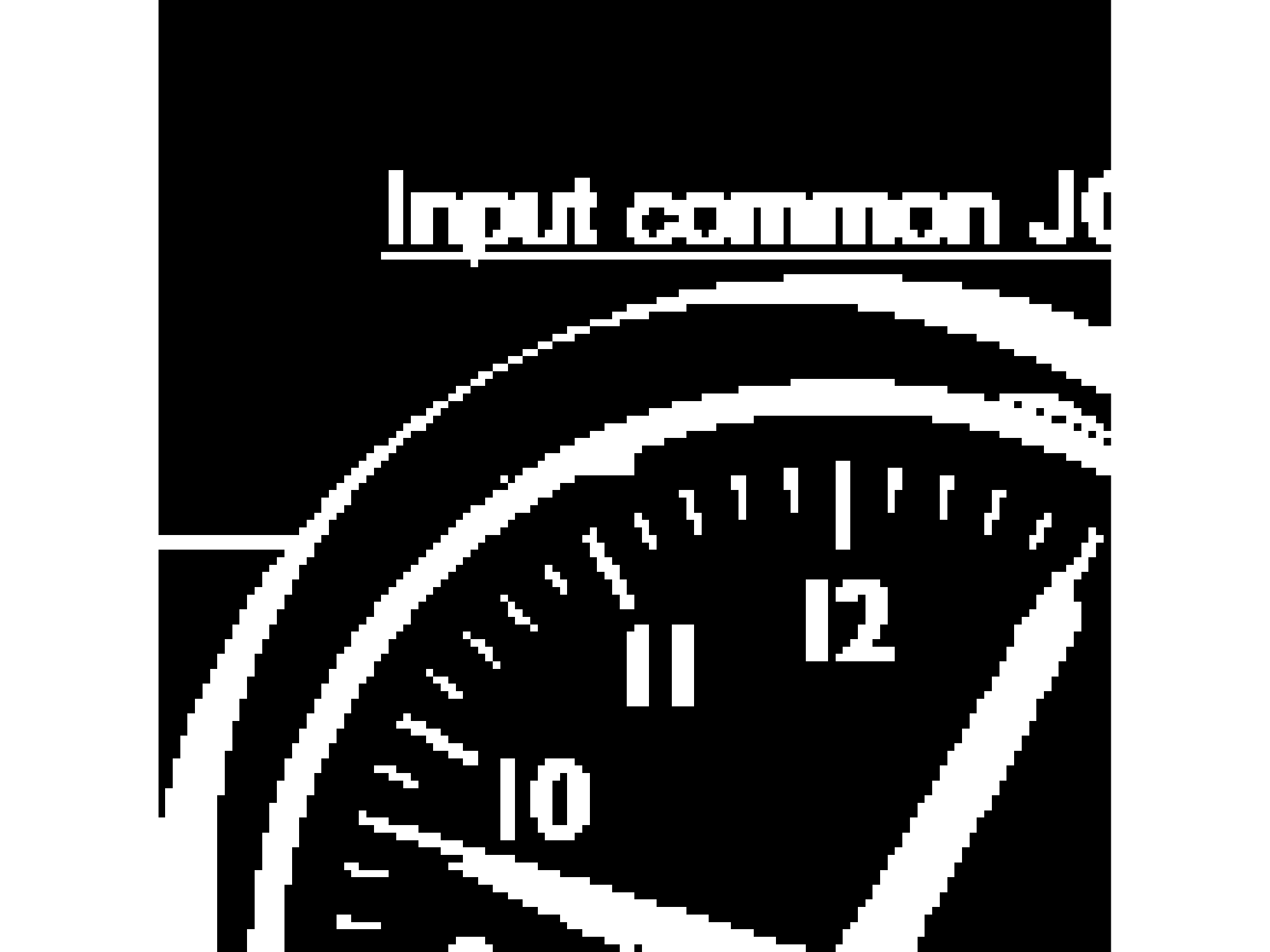}
                 \vspace{-0.45cm}
              \hspace{-4.8cm}
        \end{subfigure} \\[1ex]        
        \begin{subfigure}[b]{0.18\textwidth}
                \includegraphics[width=\textwidth]{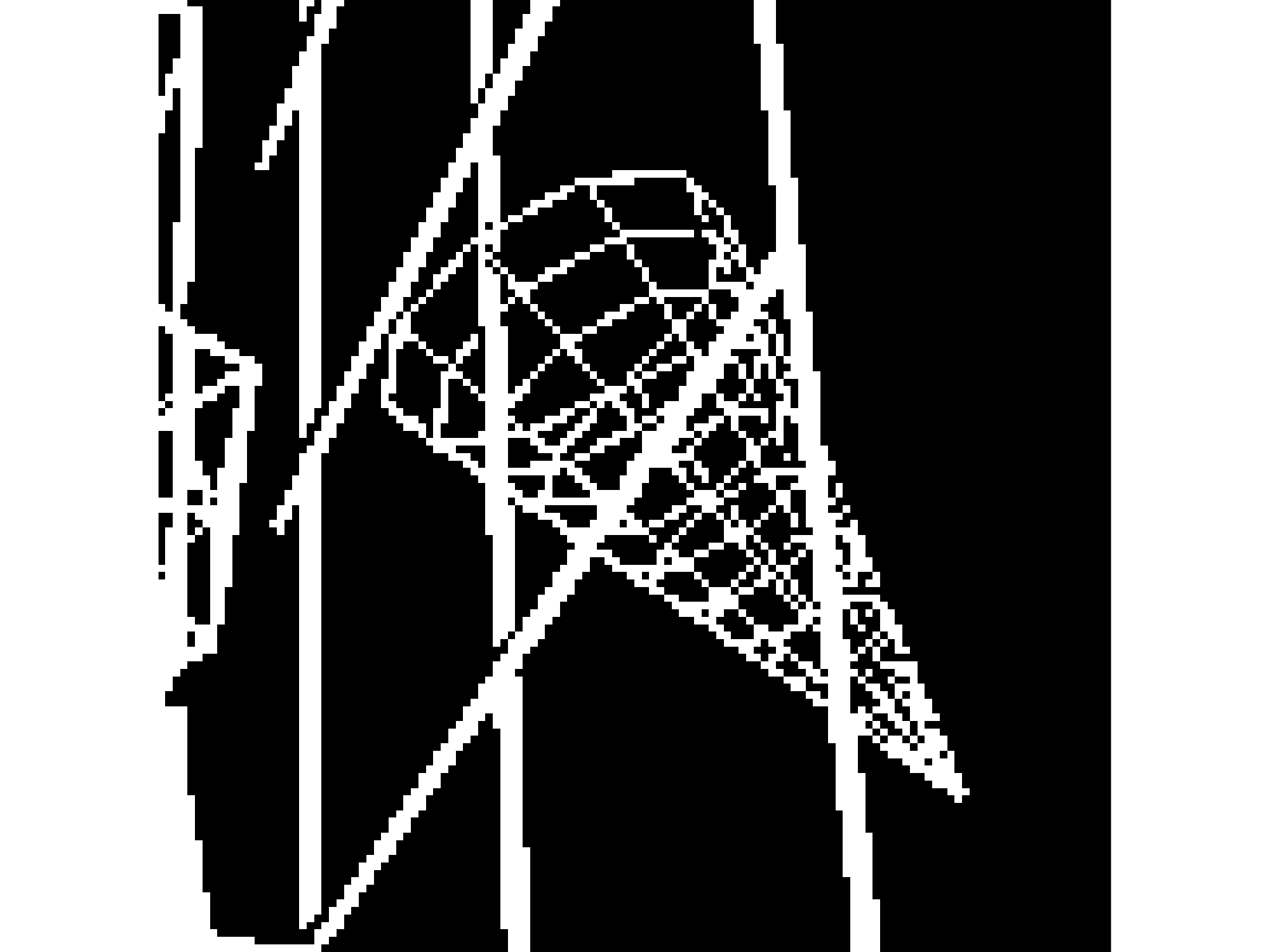}
                                \vspace{-0.5cm}
          \hspace{-2.5cm}    
        \end{subfigure}%
        ~ 
        \begin{subfigure}[b]{0.18\textwidth}
                \includegraphics[width=\textwidth]{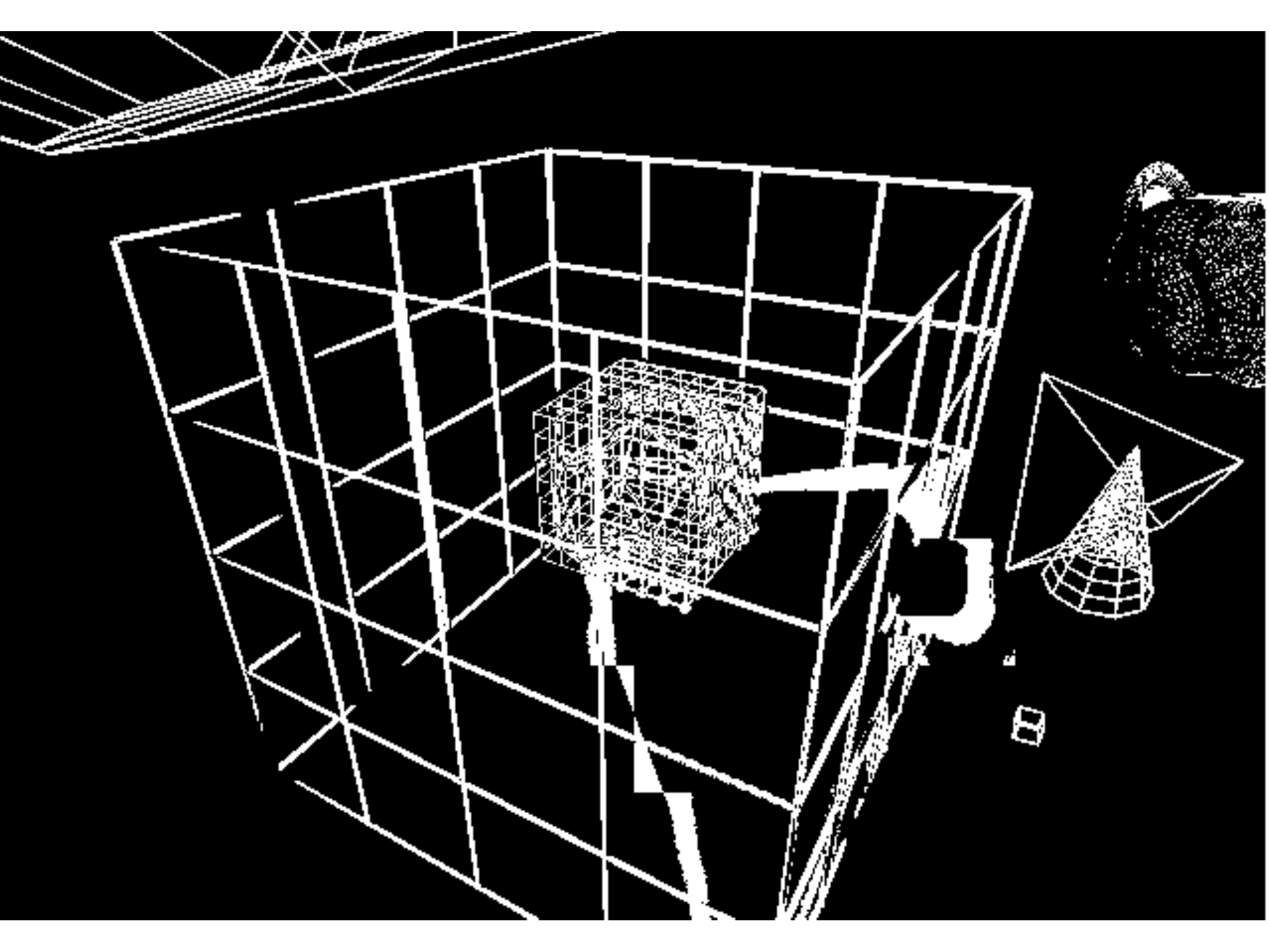}
                \vspace{-0.5cm}
            \hspace{-3cm} 
        \end{subfigure}%
        ~ 
        \begin{subfigure}[b]{0.18\textwidth}
                \includegraphics[width=\textwidth]{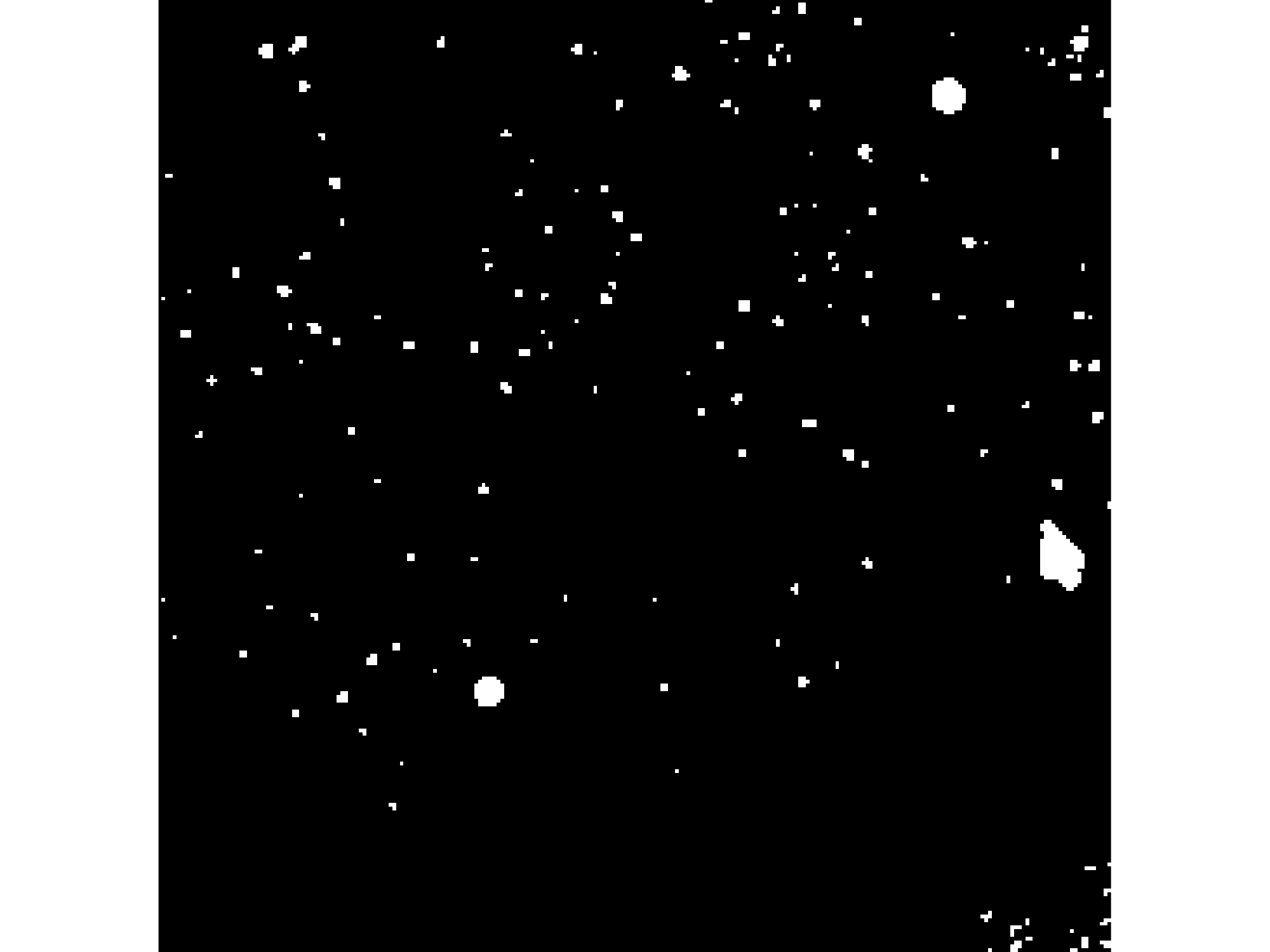}
                \vspace{-0.45cm}
            \hspace{-3cm} 
        \end{subfigure}%
        \begin{subfigure}[b]{0.18\textwidth}
			~ 
                \includegraphics[width=\textwidth]{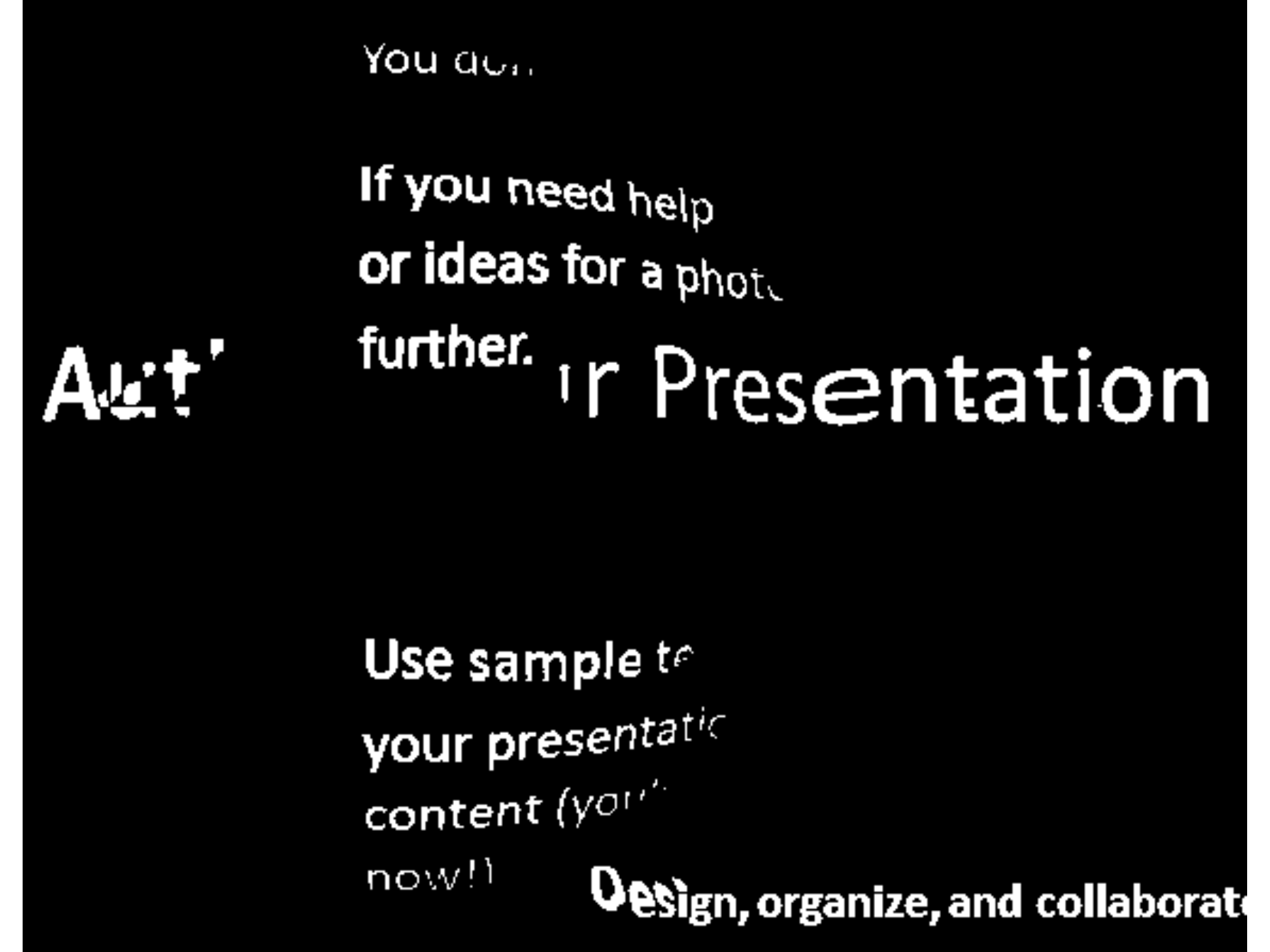} 
                \vspace{-0.45cm}
            \hspace{-3cm} 
        \end{subfigure}%
        \begin{subfigure}[b]{0.18\textwidth}
                \includegraphics[width=\textwidth]{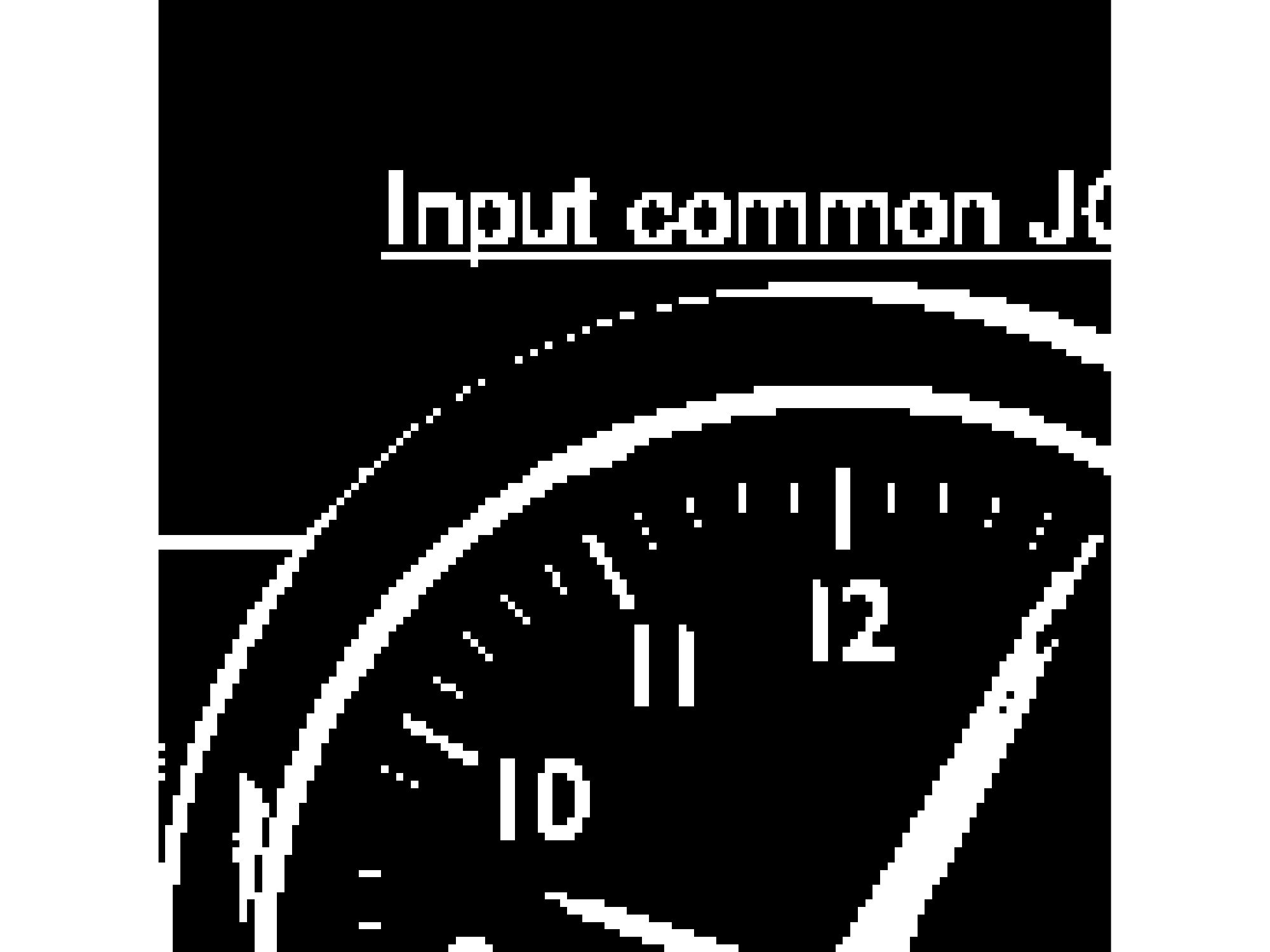}
                 \vspace{-0.45cm}
              \hspace{-4.8cm}
        \end{subfigure}
        \caption{Segmentation result for selected test images. The images in the first and second rows are the original and ground truth segmentation images. The images in the third, forth, fifth and the sixth rows are the foreground maps obtained by shape primitive extraction and coding, hierarchical clustering in DjVu, least square fitting, and least absolute deviation fitting approaches. The images in the seventh and eighth rows include the results by the proposed RANSAC and sparse decomposition algorithms respectively.}
\end{figure*}
It can be seen that in all cases the proposed algorithms give superior performance over DjVu and SPEC, and slightly better than our prior LAD approach in some images. 
Note that our dataset mainly consists of challenging images where the background and foreground have overlapping color ranges. For simpler cases where the background has a narrow color range that is quite different from the foreground, both DjVu and the proposed methods will work well. On the other hand,  SPEC  does not work well when the background is fairly homogeneous within a block and the foreground text/lines have varying colors. 

In terms of complexity, it took 20, 506 and 962 ms on average for a block of $64 \times 64$ to be segmented using RANSAC, LAD and sparse decomposition based segmentation algorithms (with the pre-processing steps) using MATLAB 2015 on a laptop with
Windows 10 and Core i5 CPU running at 2.6GHz.
Because the RANSAC method is much faster than SD and yielded very similar segmentation accuracy, the RANSAC method is preferred for practical applications.
The percentages of blocks processed by different steps of the proposed segmentation algorithm are listed in table IV (for the images in our dataset).
Note that the parameters used to identify the first three types of blocks can be chosen to achieve the desired trade-off between the segmentation accuracy and computation time. Had we chosen looser thresholds than those given in Table I, fewer blocks will need to go through the RANSAC or SD algorithm, which will reduce  the overall computation time.

\begin{table}[h]
\centering
  \caption{Percentage of blocks processed by different steps of the overall segmentation algorithm}
  \centering
\begin{tabular}{|m{4.5cm}|m{2.5cm}|}
\hline
Type of blocks  &   Percentage of Blocks\\
\hline
Pure background  & \ \ \ \ \ \ \ \ \ 15\%  \\
\hline
Smoothly varying background  & \ \ \ \ \ \ \ \ \ 14\% \\
\hline
Text/graphics on constant background & \ \ \ \ \ \ \ \ \ 11\% \\ 
\hline
 RANSAC/SD based segmentation & \ \ \ \ \ \ \ \ \  60\% \\
\hline
\end{tabular}
\label{TblComp}
\end{table}

\section{Conclusion}
This paper proposed two new algorithms for segmentation of background and foreground in screen content images. The background is defined as the smooth component of the image that can be well modeled by a set of low frequency DCT basis functions and the foreground refers to those pixels that cannot be modeled with this smooth representation. 
One of the proposed algorithms uses robust regression technique to fit a smooth function to an image block and detect the outliers. The outliers are considered as the foreground pixels. Here RANSAC algorithm is used to solve this problem.
The second algorithm uses sparse decomposition techniques to separate the smooth background from the sparse foreground layer.
Total variation of the foreground component is also added to the cost function to enforce the foreground pixels to be connected.
Instead of applying the proposed algorithms to every block, which are computationally demanding, we first check whether the block satisfies several conditions and can be segmented using simple methods. We further propose to apply the algorithm recursively using quad-tree decomposition, starting with larger block sizes. A block is split only if RANSAC or sparse decomposition cannot find sufficient inliers in this block. 
These algorithms are tested on several test images and compared with three other well-known algorithms for background/foreground separation and the proposed algorithms show significantly better performance for blocks where the background and foreground pixels have overlapping intensities.

\section*{Acknowledgment}
The authors would like to thank Ivan Selesnick and Patrick Combettes for their useful comments regarding sparse decomposition based segmentation.
We would like to thank Huawei Technologies for supporting this research, and the JCT-VC group for providing the HEVC test sequences.

\ifCLASSOPTIONcaptionsoff
  \newpage
\fi



\end{document}